\documentclass[tikz,twoside,11pt]{article}
\usepackage[nohyperref]{jmlr2e}

\usepackage{xcolor,pifont}
\usepackage{graphicx,subfigure,wrapfig}
\usepackage{booktabs,multirow,makecell,caption}
\usepackage{amsmath,amssymb,amsfonts,bm}
\usepackage{algorithm,algorithmic}
\usepackage{listings}
\usepackage{stmaryrd}
\usepackage{threeparttable}

\usepackage{color}
\definecolor{dkgreen}{rgb}{0,0.6,0}
\definecolor{gray}{rgb}{0.5,0.5,0.5}
\definecolor{mauve}{rgb}{0.58,0,0.82}

\definecolor{ao(english)}{rgb}{0.0, 0.5, 0.0}
\definecolor{ceruleanblue}{rgb}{0.16, 0.32, 0.75}
\definecolor{darkpowderblue}{rgb}{0.0, 0.2, 0.6}
\definecolor{internationalkleinblue}{rgb}{0.0, 0.18, 0.65}

\usepackage[colorlinks,linkcolor=darkpowderblue,anchorcolor=blue,citecolor=ao(english)]{hyperref}

\newcommand\err{\mathrm{err}}

\usepackage{dsfont}

\jmlrheading{1}{2000}{1-48}{4/00}{10/00}{Marina Meil\u{a} and Michael I. Jordan}

\ShortHeadings{Transferability in Deep Learning: A Survey}{Jiang et al.}
\firstpageno{1}

\begin{document}

\title{Transferability in Deep Learning: A Survey}

\author{\name Junguang Jiang
	\email jiangjunguang1123@outlook.com \\
	\addr School of Software, BNRist, Tsinghua University \\
	Beijing 100084, China
	\AND
	\name Yang Shu \thanks{Equal contribution}
	\email shu-y18@mails.tsinghua.edu.cn \\
	\addr School of Software, BNRist, Tsinghua University \\
	Beijing 100084, China
	\AND
	\name Jianmin Wang
	\email jimwang@tsinghua.edu.cn \\
	\addr School of Software, BNRist, Tsinghua University \\
	Beijing 100084, China
	\AND
	\name Mingsheng Long \thanks{Correspondence to: Mingsheng Long \textless mingsheng@tsinghua.edu.cn\textgreater.}
	\email mingsheng@tsinghua.edu.cn \\
	\addr School of Software, BNRist, Tsinghua University \\
	Beijing 100084, China
}

\editor{Leslie Pack Kaelbling}

\maketitle

\begin{abstract}%
	The success of deep learning algorithms generally depends on large-scale data, while humans appear to have inherent ability of knowledge transfer, by recognizing and applying relevant knowledge from previous learning experiences when encountering and solving unseen tasks. 
	Such an ability to acquire and reuse knowledge is known as \emph{transferability} in deep learning. It has formed the long-term quest towards making deep learning as \emph{data-efficient} as human learning, and has been motivating fruitful design of more powerful deep learning algorithms. 
	We present this survey to connect different isolated areas in deep learning with their relation to transferability, and to provide a unified and complete view to investigating transferability through the whole \emph{lifecycle} of deep learning. 
	The survey elaborates the fundamental goals and challenges in parallel with the core principles and methods, covering recent cornerstones in deep architectures, pre-training, task adaptation and domain adaptation. 
	This highlights unanswered questions on the appropriate objectives for learning transferable knowledge and for adapting the knowledge to new tasks and domains, avoiding catastrophic forgetting and negative transfer.
	Finally, we implement a benchmark and an open-source library, enabling a fair evaluation of deep learning methods in terms of transferability. 
\end{abstract}

\begin{keywords}
	Deep learning, transferability, pre-training, adaptation, library, benchmark
\end{keywords}

\newpage
\tableofcontents
\newpage

\section{Introduction}

Deep learning~\citep{cite:Nature15DeepLearning} is a class of machine learning algorithms that utilize multiple processing layers to learn representations of data with multiple levels of abstraction.
These multiple processing layers, also called deep neural networks (DNNs), are empowered with the ability to discover different explanatory factors of variation behind the intricate structured data~\citep{RepresentationLearningReview}.
With essential advances in network architectures, training strategies and computation devices, deep learning has made breakthroughs or even revolutions in various areas, such as computer vision~\citep{cite:NIPS12AlexNet,cite:CVPR16ResNet}, natural language processing~\citep{cite:GPT}, speech processing~\citep{cite:ICML16DeepSpeech2},
computational biology~\citep{cite:Nature20AlphaFold},  games~\citep{cite:Nature16AlphaGO,cite:Nature19AlphaStar} and so forth.
Despite its great success in these important areas, deep learning is still faced with the grand challenge of \textit{data efficiency}. Most mainstream deep learning methods require big datasets in the order of millions or even trillions to achieve good performance, yet collecting and annotating such huge amount of data for each new task or domain are expensive and even prohibitive. This data efficiency challenge heavily impedes the adoption of deep learning to a wider spectrum of application scenarios.

An effective solution to this challenge is to explore the \textit{transferability} in deep learning.
Transferability is a foundational ability of human learning: human beings can gain relevant knowledge from other related problems and apply it to handle new problems with extremely few samples \citep{cite:LearningtoLearnBook}.
In deep learning, transferability refers to the ability of deep neural networks to
extract transferable representations from some source tasks and then adapt the gained representations to improve learning in related target tasks \citep{cite:BengioUnsupervised}.
Recent advances in deep learning reveal that deep models trained via upstream tasks on large-scale data tend to yield good transferability to a variety of downstream tasks, such as visual object detection \citep{Faster}, natural language understanding \citep{cite:NAACL19BERT}, to name a few. Transferability has become the central property of deep learning for improving data efficiency. It is on par with generalizability, interpretability, and robustness for bridging the gap between machine learning and human learning.

\begin{figure}[h]
	\centering
	\includegraphics[width=0.8\textwidth]{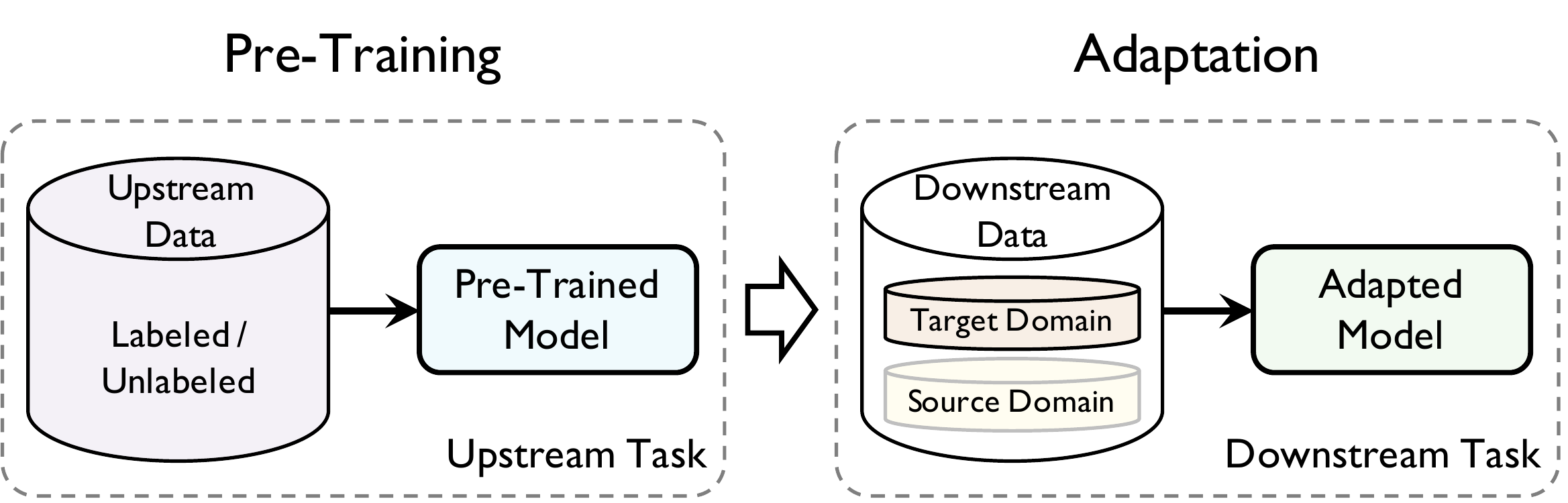}
	\caption{The two-stage lifecycle of most deep learning applications. In the first stage, the deep model is \emph{pre-trained} on an {upstream task} with large-scale data (labeled or unlabeled) for gaining transferable knowledge. In the second stage, the pre-trained model is \emph{adapted} to a {downstream task} in the target domain with labeled data; If the downstream task only has unlabeled data, then additional labeled data from another source domain of identical learning task but different data distribution will be used to improve performance.}
	\label{fig:lifecycle}
\end{figure}

Towards gaining and applying knowledge with good transferability, the lifecycle of many deep learning applications is divided into two stages: pre-training and  adaptation (Figure~\ref{fig:lifecycle}).
The goal of the \textit{pre-training} stage is to gain the transferable knowledge. The deep models are pre-trained on an upstream task with large-scale data (either labeled or unlabeled) to learn disentangled representations or reusable parameters that are transferable to a variety of downstream tasks.
The goal of the \textit{adaptation} stage is to reuse the transferable knowledge. The pre-trained models are adapted to a downstream task in the target domain with labeled data, and the previously learned knowledge enables better generalization with fewer labeled samples.
When the downstream task only has unlabeled data, additional labeled data from another source domain of identical learning task but different data distribution will be used to improve the data efficiency of the adapted model \citep{DANN}.

It is helpful to highlight the difference underlying the transferability in the two stages. The pre-training stage focuses mainly on the \emph{generic} transferability, i.e., obtaining a general transferable representation that can improve the performance of as many downstream tasks as possible. In contrast, the adaptation stage pays attention to the \emph{specific} transferability, i.e., how to exploit the transferable knowledge in pre-trained models for a specific kind of downstream tasks, or how to improve the transferability between related
domains of the same downstream task.
The generic transferability is attractive since it may benefit many downstream tasks without additional cost or special design.
Yet it may ignore the special structures of downstream tasks that are crucial for stronger transferability, thus the specific transferability is still necessary in many cases.
Recently, the gap between the pre-training stage and the adaptation stage is getting closer.
Several pre-training methods are designed to obtain fast model adaptation ability in the adaptation stage \citep{cite:ICML17MAML}, while some adaptation methods try to convert downstream tasks into pre-training tasks to make full use of the generic transferability of pre-trained models \citep{cite:NIPS20GPT3}.

Transferability lies at the core of the whole lifecycle of deep learning, yet different areas such as domain adaptation~\citep{Zhuang21} and continual learning~\citep{Delange21}, mainly explore transferability in a partial regime of the lifecycle. This is not enough to achieve a complete picture of transferability.
Thereby, we present this survey to connect different isolated areas in deep learning with their relation to transferability, and to provide a unified and complete view to investigate transferability through the whole lifecycle of deep learning.
Due to the broadness of the scope and the limitation of the space, we do not aim to cover all methods towards transferability. Instead, we elaborate on the core principles and methods and then give a brief review of the expanded literature. We further implement \texttt{TLlib}, a high-quality open library to provide a fair evaluation of typical methods.
We hope this survey can highlight the grand picture of transferability in deep learning, and provide a useful navigation to researchers  interested in improving the data efficiency of deep learning.

\begin{table}[!t]
	\centering
	\small
	\caption{Notations and descriptions used in the survey.}
	\label{table:notation}
	\begin{tabular}{ll}
		\hline
		\toprule
		$\mathcal{X}$           & Input space                                                        \\
		$\mathcal{Y}$           & Output space                                                       \\
		$\mathcal{D}$           & A fixed but unknown distribution over $\mathcal{X}$                \\
		$\mathcal{\widehat{D}}$ & Empirical distribution of a sample drawn i.i.d. from $\mathcal{D}$ \\
		$P(\cdot)$              & Probability of an event                                            \\
		$\mathbb{E}(\cdot)$     & Expectation of a random variable                                   \\
		$\mathcal{U}$           & Upstream data                                                      \\
		$\mathcal{S}$           & Source domain in downstream data                                   \\
		$\mathcal{T}$           & Target domain in downstream data                                   \\
		$\mathcal{H}$           & Hypothesis space                                                   \\
		$h$                     & A hypothesis in the hypothesis space $\mathcal{H}$                 \\
		$\psi$                  & Feature generator                                                  \\
		$\theta$                & Hypothesis parameter                                               \\
		$\mathbf{x}$            & Model input                                                        \\
		$\mathbf{y}$            & Model output                                                       \\
		$\mathbf{z}$            & Hidden activation of the feature generator                         \\
		$D$                     & A discriminator to distinguish different distributions             \\
		\bottomrule
	\end{tabular}
\end{table}

\subsection{Terminology}

Foremost, we give several definitions related to transferability, and the summary of notations and their descriptions used in this survey can be found in Table \ref{table:notation}.
Denote the input space as $\mathcal{X}$ and the output space as $\mathcal{Y}$, and assume that there exists an unknown labeling function $f:\mathcal{X}\mapsto \mathcal{Y}$.
Formally, a \textit{task} corresponds to learning an underlying labeling function $f$.
To learn a task, we first collect a set of samples $\widehat{\mathcal{D}}=\{\mathbf{x}_1,..., \mathbf{x}_n\}$, which are drawn {independently and identically distributed (i.i.d.)} from some fixed but unknown distribution $\mathcal{D}$.
Formally, a \textit{domain} is a marginal probability distribution $P(\mathbf{X})$ defined on a certain input space $\mathcal{X}$.
Consider a set of hypotheses $\mathcal{H}$ and a specific loss function $\ell: \mathcal{Y}\times \mathcal{Y}\mapsto \mathbb{R}_+$, the objective of the learner is to select a hypothesis $h\in \mathcal{H}$ that yields the lowest generalization error, $\min_{h\in\mathcal{H}} \mathbb{E}_{\mathbf{x} \sim \mathcal{D}} \ell(h(\mathbf{x}), f(\mathbf{x}))$.

\begin{definition}[Transferability]
	\label{def:transferability}
	Given a source domain $\mathcal{S}$ with learning task $t_{\mathcal{S}}$ and a target domain $\mathcal{T}$ with learning task $t_{\mathcal{T}}$,
	transferability is the ability of gaining transferable knowledge from $t_{\mathcal{S}}$ on $\mathcal{S}$ and reusing the knowledge to decrease the generalization error of $t_{\mathcal{T}}$ on $\mathcal{T}$, under the distribution shift $\mathcal{S} \neq \mathcal{T}$ or the task discrepancy $t_{\mathcal{S}} \neq t_{\mathcal{T}}$.
\end{definition}

In the deep learning lifecycle (Figure~\ref{fig:lifecycle}), the {pre-training} stage aims to \emph{gain} transferable knowledge via learning on upstream task with large-scale data, while the {adaptation} stage aims to \emph{reuse} the pre-trained knowledge to improve the data efficiency in downstream tasks. The upstream and downstream are different in both learning tasks and data distributions. To conform with the literature, in the pre-training stage, we will replace the notions of source domain/task with the widely-used upstream data/task, denoted as $\mathcal{U}$ and $t_{\mathcal{U}}$ respectively.

\begin{figure}[!t]
	\centering
	\includegraphics[width=1.0\textwidth]{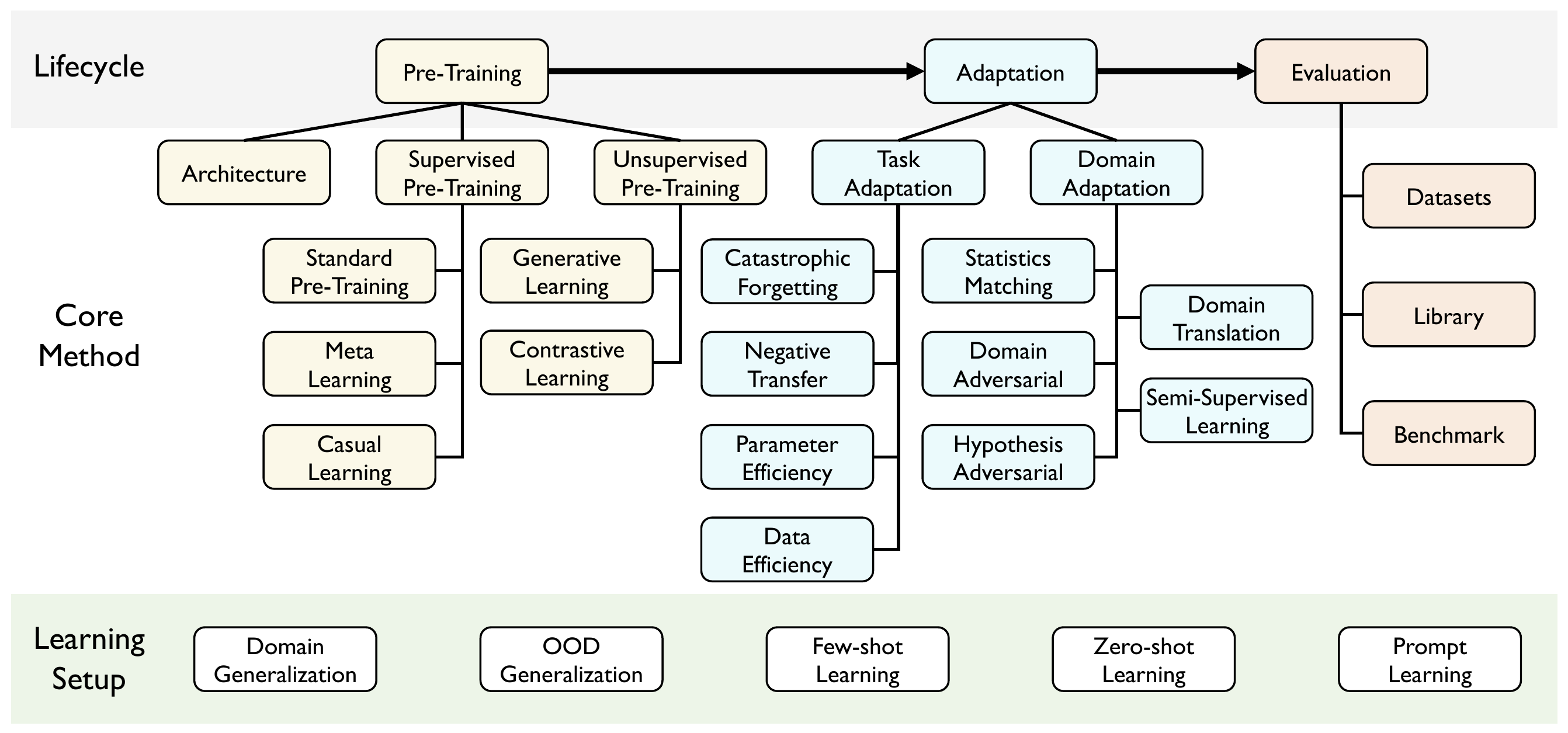}
	\caption{Overview of this survey. The survey is organized around the lifecycle (pre-training, adaptation, and evaluation) of deep learning applications and focuses on the core problems and methods towards transferability. Besides, we briefly review related learning setups.}
	\vspace{-10pt}
	\label{fig:overall_survey}
\end{figure}

\subsection{Overview}

The survey is organized around how to acquire and utilize the transferability in deep learning throughout its whole lifecycle, including \textit{pre-training}, \textit{adaptation}, and \textit{evaluation} (Figure~\ref{fig:overall_survey}).
\begin{itemize}
	\item \textbf{Pre-Training.} We first briefly discuss some important model \textit{architectures} that make pre-trained representations transferable. Then we elaborate on \textit{supervised pre-training} and \textit{unsupervised pre-training}, which are distinguished by the availability of labeled or unlabeled data for pre-training.
	      In \textit{supervised pre-training}, we cover both standard practices commonly used in the industry and research advances in academia to acquire transferability on the labeled data.
	      In \textit{unsupervised pre-training}, we cover the latest designs of proper pre-training tasks on unlabeled data to gain transferability.
	\item \textbf{Adaptation.} We mainly elaborate on \textit{task adaptation} and \textit{domain adaptation}, which are divided by whether there exists another related source domain in addition to the pre-trained model for boosting the downstream task performance.
	      In \textit{task adaptation}, we	first pinpoint several open problems caused by the discrepancy between upstream tasks and downstream tasks, then illustrate how	different task adaptation paradigms \citep{cite:NIPS14HowTransferable, cite:NIPS20GPT3} close the task discrepancy to better utilize the transferability.
	      In \textit{domain adaptation}, we first pinpoint the most influential theories for closing the distribution shift \citep{DATheroy07, DATheroy10}, then elaborate how to derive solid learning algorithms \citep{DAN, DANN} from these theories to enhance the transferability of deep models across domains.
	\item \textbf{Evaluation.} We mainly investigate the
	      transferability gained and reused by different pre-training and adaptation methods on several large-scale datasets released recently in the literature. Note that we omit some small-scale and relatively obsolete datasets to make our benchmark concise and easy to report. To facilitate fair evaluation and full reproduction of existing algorithms, we open source \texttt{TLlib}, a high-quality library along with this survey at \url{https://github.com/thuml/Transfer-Learning-Library}.
\end{itemize}

Pre-training and adaptation lie at the core methods towards transferability. In parallel with them, there are some fields that are also closely related to the transferability in deep learning, such as \textit{domain generalization}~\citep{DomainBed}, \textit{out-of-distribution (OOD) generalization} \citep{cite:Turing21DeepLearningForAI}, \textit{few-shot learning}~\citep{cite:ICLR19ACloserLook}, etc. Recent evaluation shows that these learning setups can largely benefit from the advancement in pre-training and adaptation and we will give a brief review to them in the related sections.

\section{Pre-Training}
\label{sec:pretraining}

\begin{figure}[!b]
	\centering
	\includegraphics[width=1.0\textwidth]{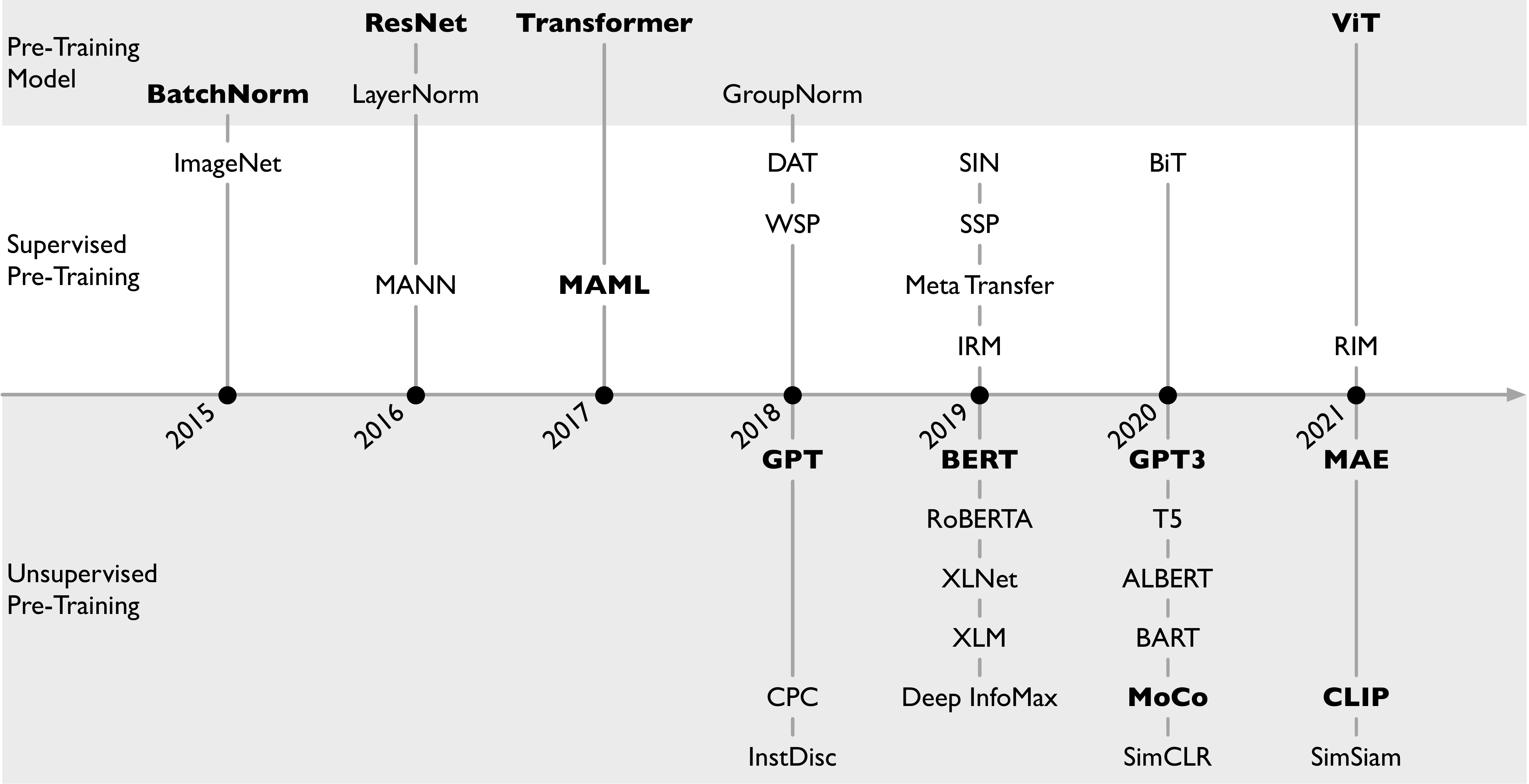}
	\caption{Cornerstones of pre-training methods for \emph{gaining} knowledge of transferability.}
	\label{fig:pretrain_overview}
\end{figure}

Despite yielding unprecedented performances on various machine learning tasks, the deep learning methods require large amounts of labeled data to generalize well. This \emph{data hungry} nature limits their application to a wide variety of domains and tasks, especially to scenarios short of data and annotations. Pre-training, which obtains transferable representations or models from upstream tasks with large-scale data to boost the performance on downstream tasks, is one of the most common and practical solutions to the problem of data scarcity.
In this section, we will first review some important model architectures that have a great impact on the transferability of pre-trained representations in Section \ref{sec:pre_trained_models}. Then we elaborate on how to \emph{gain knowledge} of improved transferability via supervised pre-training on large-scale labeled data in Section \ref{sec:supervised_pre_training}
and via unsupervised pre-training on much larger unlabeled data in Section \ref{sec:unsupervised_pre_training}.
Figure \ref{fig:pretrain_overview} overviews the recent cornerstones of pre-training methods.

\subsection{Pre-Training Model}
\label{sec:pre_trained_models}

Pre-training has a big interplay with the model architecture. On the one hand, pre-training techniques, such as greedy layerwise unsupervised pre-training \citep{cite:NIPS07GreedyLayerWiseTraining}, have eased the training of many deep architectures. On the other hand, as neural networks evolve from shallow to deep, they have a larger capacity to capture knowledge by pre-training from large-scale data, which increases their transferability to downstream tasks.

Model architecture has a great influence on the transferability of knowledge obtained via pre-training. \cite{cite:CVPR19DoBetterTransfer} find that the performance of the pre-trained models on the downstream tasks is highly correlated with the accuracy on the pre-training tasks, which suggests that improving the performance on the pre-training task serves as a direct way for improving transferability.
The depth of the architecture, or more precisely, the \emph{capacity}  of the model, is deemed the most critical factor to its transferability.
However, training very deep neural networks have remained a grand difficulty for decades.
\cite{cite:CVPR16ResNet} observe a degradation of training accuracy by increasing the network depth, which implies that deeper models are more difficult to optimize.
Instead of fitting a desired mapping $h(\mathbf{x})$ by a few stacked layers, they proposed Residual Network (ResNet) to explicitly fit a residual mapping $\delta(\mathbf{x}):=h(\mathbf{x})-\mathbf{x}$ and then recast the original mapping into $\delta(\mathbf{x})+\mathbf{x}$.
As a result, ResNet improves feature and gradient flows and enables end-to-end training of hundreds of and even thousands of layers, allowing the capacity of pre-trained models to scale up easily.
\cite{BatchNorm} hypothesize that the optimization difficulty also comes from the internal covariate shift caused by layerwise transformation. To stabilize training very deep models, they proposed Batch Normalization (BatchNorm) \citep{BatchNorm}, which performs normalization for each training mini-batch within the architecture. This design is extensively used by ResNet.
\cite{cite:ECCV20BigTransfer} find that BatchNorm is suboptimal for transfer due to the requirement of distribution-dependent moving averaged statistics. They proposed Big Transfer (BiT) to replace BatchNorm by GroupNorm \citep{GroupNorm}, which generates pre-trained models of strong performance on downstream tasks.

\begin{figure}[!b]
	\vspace{-10pt}
	\centering
	\includegraphics[width=0.6\textwidth]{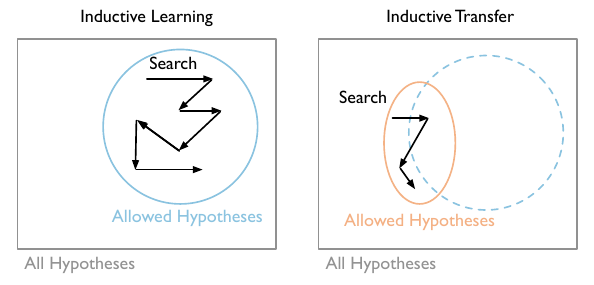}
	\caption{Designed inductive bias (left) and learned inductive bias from pre-training (right).}
	\label{fig:intrinsic}
	\vspace{-10pt}
\end{figure}

The pre-training paradigm also reshapes the design of model architectures.
In classic supervised learning, models usually have strong \emph{inductive bias} such as the local connectivity assumption in Convolutional Neural Network (CNN) and Recurrent Neural Network (RNN). A strong inductive bias makes pre-training of deep models more data-efficient and generalize better when training data is scarce, yet on the other hand, it also limits the expressiveness and transferability of the deep models when there is large-scale data for pre-training.
Thus, Transformer~\citep{cite:NIPS17Transformer} removes the local connectivity assumption and models the global dependencies between every two tokens.
The connection weights are dynamically computed by the self-attention mechanism and then the feature aggregation in Transformer depends on these attentions calculated from the input sequence, while the token positions in the sequence are encoded by positional embedding.
Transformers are powerful for sequence modeling in natural language processing, and Vision Transformer (ViT) \citep{cite:ICLR21VIT} extends them to computer vision. ViT splits an image into fixed-size patches, linearly embeds each of them, adds positional embeddings, and feeds the resulting sequence of vectors to a standard Transformer encoder.
In summary, Transformer makes least assumptions on the structural information of data, which makes Transformer an \emph{expressive} architecture for storing the transferable knowledge extracted by pre-training on large amounts of training data~\citep{cite:NAACL19BERT, cite:GPT}.

In some sense, pre-training provides a \emph{learned} inductive bias for the downstream tasks \citep{transfer_learning}.
Many downstream tasks only have hundreds or thousands of labeled samples, yet the pre-trained Transformers with hundreds of millions of parameters can generalize well after fine-tuning on such small data.
To explain this phenomenon, \cite{aghajanyan2020intrinsic} empirically show that pre-training minimizes the intrinsic dimension \citep{IntrinsicDimension}, which measures the number of parameters required to closely approximate the optimization problem. Further, an intrinsic-dimension generalization bound is given, indicating that the pre-trained parameters implicitly affect the \textit{inductive bias} of models and a larger pre-trained model might correspond to a  smaller allowed hypothesis space during fine-tuning (see Figure \ref{fig:intrinsic}).
The success of Transformer reveals that as the amount of pre-training data increases, the learned inductive bias is able to outperform the manually designed inductive bias in terms of transferability.

\subsection{Supervised Pre-Training}
\label{sec:supervised_pre_training}

\begin{figure}[h]
	\centering
	\includegraphics[width=0.55\textwidth]{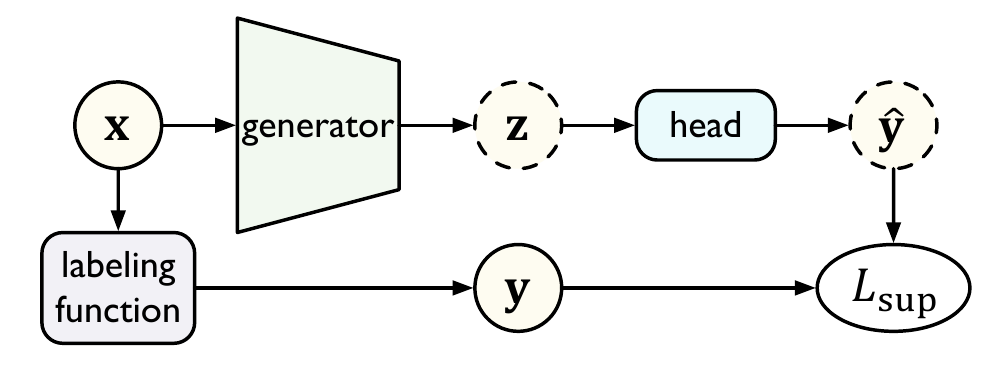}
	\caption{Standard supervised pre-training. The model is composed of a feature generator and a task-specific head. The goal is to obtain a feature generator capturing transferable knowledge from large-scale labeled data. After pre-training, the feature generator is adapted to downstream tasks, while the task-specific head is usually discarded.
	}
	\label{fig:supervised}
\end{figure}

Supervised pre-training aims to obtain models on large-scale labeled data and then transfers these models to boost downstream tasks (see Figure \ref{fig:supervised}).
Supervised pre-training is commonly employed in computer vision, where image classification on ImageNet \citep{deng_imagenet:_2009,ImageNet} is often used as the pre-training task.
The pre-trained models can be transferred to downstream tasks by reusing the representations from the feature generator \citep{cite:Arxiv13Overfeat}. \cite{cite:ICML14Decaf} find that the generic visual representations pre-trained on ImageNet outperforms many conventional feature descriptors on various object recognition tasks. \cite{cite:NIPS14HowTransferable} find that transferring the pre-trained models by fine-tuning the whole models yields better generalization performance on new tasks.

Among the factors that influence the transferability of pre-trained models, the quantity and quality of the pre-training data might be the most important.
BiT~\citep{cite:ECCV20BigTransfer} emphasizes that training on larger datasets is vital for better transferability. Yet the data labeling is labor-exhaustive and time-consuming, which limits the possible size of the annotation data.
To break this limitation, \cite{cite:ECCV18ExploringWeakly} explore Weakly Supervised Pre-training (WSP) on IG-1B-Targeted, a dataset of  billions of images with social media hashtags.
\cite{cite:Arxiv19BillionSemiSupervised} further explore web-scale Semi-Supervised Pre-training (SSP) on YFCC100M, a dataset of billions of unlabeled images along with a relatively smaller set of task-specific labeled data. These methods improve clearly against the counterpart trained with only clean labeled data and achieve stronger transfer performance. On the other hand, Domain Adaptive Transfer (DAT) \citep{cite:Arxiv18DomainAdaptiveTransfer} studies the influence of data quality and finds that using more data does not necessarily lead to better transferability, especially when the dataset is extremely large. Thus, an importance weighting strategy is proposed to carefully choose the pre-training data that are most relevant to the target task.~\cite{cite:CVPR18LargeScaleFinegrained} also find that pre-training on more similar upstream data improves transferability to fine-grained downstream tasks. They propose to estimate domain similarity via the Earth Mover's Distance to choose proper pre-training data. \cite{cite:ICLR19ImageNetBias} find that models trained supervisedly on ImageNet are biased towards textures in images, and propose to pre-train with a Stylized ImageNet (SIN), which fixes the texture bias and encourages the models to learn shape-based representations of better transferability.

While standard supervised pre-training is powerful when there are enough labeled data, it still has drawbacks that may limit the transferability of the model.
For instance, standard supervised pre-trained models are vulnerable to adversarial examples~\citep{cite:ICLR15AdversarialExample}, and \cite{cite:NIPS20AdversariallyRoubstTransfer} enhance the adversarial robustness of the pre-trained models to achieve better transferability.
In addition, there are alternative pre-training methods for improving the transferability of deep models.
Section \ref{sec:meta_learning} will elaborate on meta-learning, which aims to obtain pre-trained models that adapt to downstream tasks with less training time and less training data.
Section \ref{sec:causal_learning} will review causal learning, which aims to obtain distributionally robust and generalizable pre-trained models.

\subsubsection{Meta-Learning}
\label{sec:meta_learning}

Standard supervised pre-training gains transferable representations to boost the learning of new tasks.
However, it still requires to fine-tune the pre-trained models with hundreds or thousands of labeled data and with many gradient updates when adapting to the new task.
In contrast, people have the ability to quickly adapt to different related new tasks with few labeled data.
Meta-learning, also known as learning to learn \citep{cite:1987EvolutionaryPrinciples}, aims to pursue such kind of \emph{efficient} transferability in the pre-training stage.

\begin{figure}[!t]
	\centering
	\subfigure[Learning Setup]{
		\includegraphics[height=0.27\textwidth]{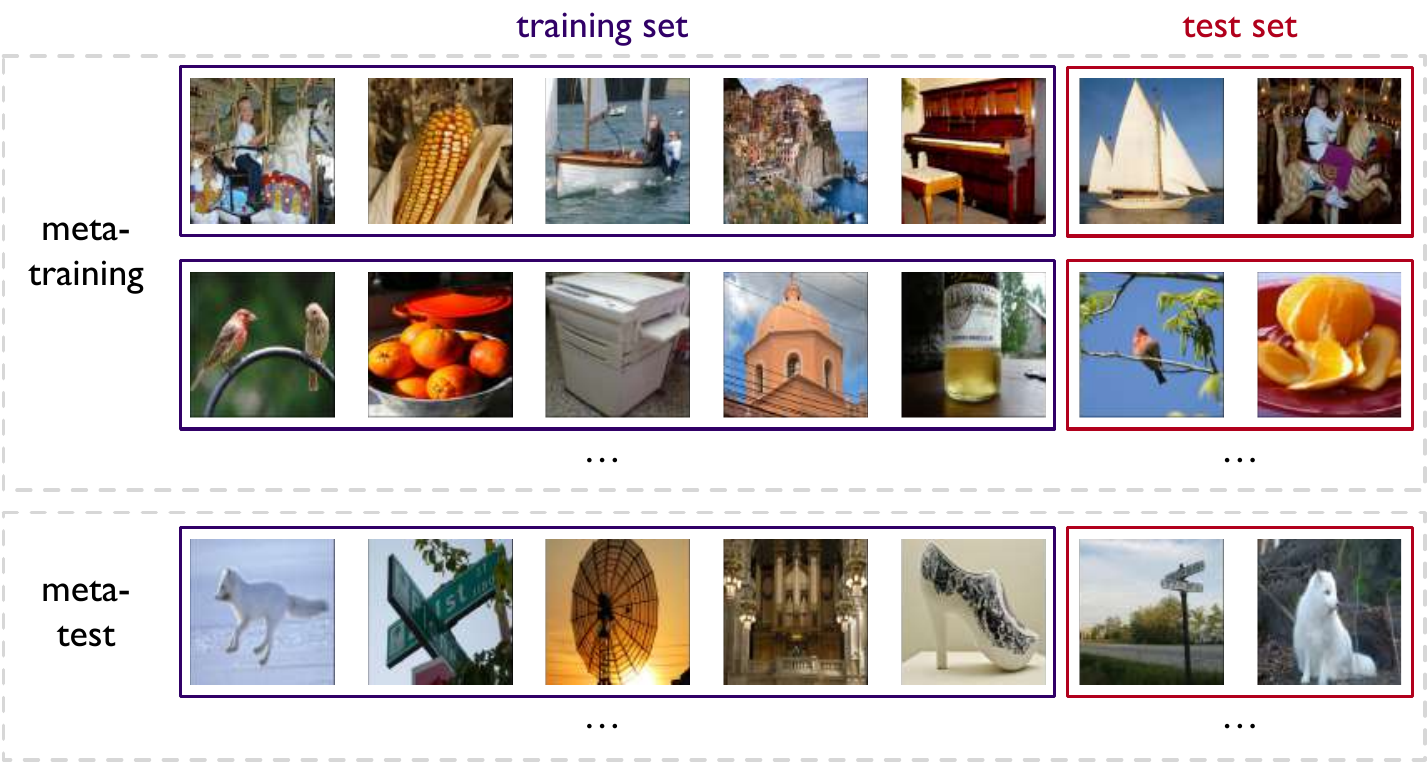}
		\label{fig:meta_learning}
	}
	\hfil
	\subfigure[Architecture]{
		\includegraphics[height=0.27\textwidth]{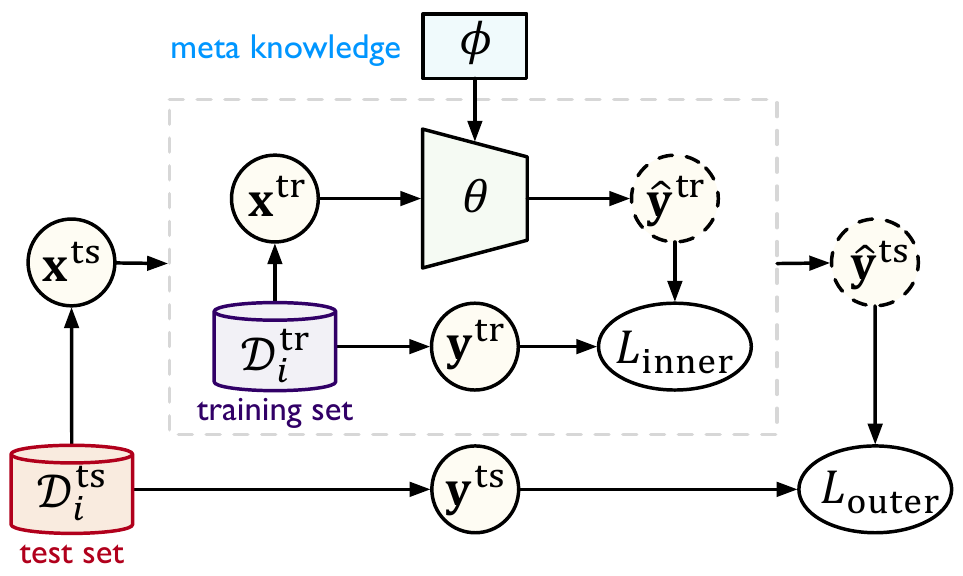}
		\label{fig:meta_training}
	}
	\caption{Learning setup and architecture for meta-learning.
	{(a)} Meta-learning consists of two phases, \textit{meta-training} and \textit{meta-testing}.  Meta-training gains meta knowledge $\phi$ from training tasks to help the model $\theta$ adapt quickly to a new task in meta-testing, where each task consists of a training set and a test set.
	{(b)}
	In the inner level optimization, the model $\theta$ is updated with the training set $\mathcal{D}^{\text{tr}}_{i}$ using meta knowledge $\phi$. In the outer level optimization, the updated model is evaluated on the test set $\mathcal{D}^{\text{ts}}_{i}$ to find better meta knowledge $\phi$.
	}
	\label{fig:meta_learning_overview}
\end{figure}

The core idea of meta-learning is to equip the model with some \textit{meta knowledge} $\phi$ that captures intrinsic properties of different learning tasks, which is called \textit{meta-training}.
When facing a new task, the learned meta knowledge could help the target model $\theta$ adapt to the task faster, which is called \textit{meta-testing}.
Meta-learning is based on a simple machine learning principle that test and training conditions should be matched.
As shown in Figure~\ref{fig:meta_learning}, to simulate the fast adaptation condition during meta-testing,
the meta-training data is constructed into a collection of $n$ learning tasks, and each task $i\in[n]$ contains a training set $\mathcal{D}^{\text{tr}}_{i}$ for adaptation to this task and a test set $\mathcal{D}^{\text{ts}}_{i}$ for evaluation\footnote{$\mathcal{D}^{\text{ts}}$ is a \emph{surrogate} test set used during meta-training to simulate different tasks and improve the model. It is different from the true test set in the general setting in machine learning.}.
As shown in Figure~\ref{fig:meta_training}, the learning objective of meta-training is a bi-level optimization problem,
\begin{equation}
	\phi^{*} = {\arg}\mathop{\max}_{\phi} \sum_{i=1}^{n} \log P(\theta_{i}(\phi)|\mathcal{D}^{\text{ts}}_{i}),  \quad
	\text{where} \ \theta_{i}(\phi) = {\arg}\mathop{\max}_{\theta}\log P(\theta| \mathcal{D}^{\text{tr}}_{i}, \phi). \\
\end{equation}
Here the inner level optimization updates the model $\theta$ with the training set $\mathcal{D}^{\text{tr}}_{i}$ using meta knowledge $\phi$, and the outer level optimization evaluates the updated model with the test set $\mathcal{D}^{\text{ts}}_{i}$ to find better meta knowledge of stronger transferability. The key to enhancing the transferability of meta-learning methods is to design a proper form of meta knowledge.

\textbf{Memory-Based Meta-Learning} considers \emph{memory mechanisms} as the meta knowledge. A controller writes knowledge extracted from training data $\mathcal{D}_i^{\text{tr}}$ into the memory, and reads from the memory to adapt the base learner $\theta$ to make predictions on test data $\mathcal{D}_i^{\text{ts}}$. The parameter of the controller is updated to find transferable knowledge.
Memory-Augmented Neural Network (MANN) \citep{cite:ICML16MetaLearningMemory} stores bound sample representation-class label information in the external memory, which can then be retrieved as features for making predictions when a sample from the same class is presented.
Meta Network~\citep{cite:ICML17MetaNetworks} designs another memory mechanism
where a base learner provides information about the status of the current task while the meta learner interacts with the external memory to generate parameters for the base learner to quickly learn the new task.
Memory-based meta-learning methods improve transferability in various downstream tasks, such as few-shot classification and reinforcement learning. However, they require a careful design of the black-box  architecture to incorporate the memory mechanism, and it is unclearer what is stored and retrieved in the memory and why it helps adapt the model.

\textbf{Optimization-Based Meta-Learning} considers a good \emph{initialization} of the model as the meta knowledge.
The motivation of Model-Agnostic Meta-Learning (MAML)~\citep{cite:ICML17MAML} is to explicitly seek for an initialization that is most transferable for fine-tuning, i.e., only a small amount of gradient steps and a few labeled data are needed for the model to generalize to a new task.
To learn such an initialization, for each sampled task $i \in [n]$, the model $\phi$ is first updated on its training data $\mathcal{D}_i^{\text{tr}}$ using one gradient step of size $\alpha$,
\begin{equation}
	\theta_i = \phi - \alpha \nabla_\phi L(\phi, \mathcal{D}_i^{\text{tr}}).
\end{equation}
which mimics the situation of fine-tuning the model from the starting point of $\phi$.
As meta knowledge, $\phi$ should have good transferablity, such that for all tasks $i\in[n]$, the fine-tuned parameters  $\theta_i$ could perform well on the test set $\mathcal{D}_i^{\text{ts}}$,
\begin{equation}
	\min_\phi \sum_{i=1}^nL(\theta_{i}(\phi), \mathcal{D}_i^{\text{ts}}) = \sum_{i=1}^n L(\phi - \alpha \nabla_\phi L(\phi, \mathcal{D}_i^{\text{tr}}), \mathcal{D}_i^{\text{ts}}).
\end{equation}
The meta knowledge of MAML is high-dimensional, hindering MAML from deeper models.
To tackle it, Meta Transfer~\citep{cite:CVPR19MeteTransferLearning} uses standard pre-training for initialization and performs meta-training with light-weight neuron operations (e.g.~scaling and shifting over tasks), which reduces the training tasks needed to acquire the meta knowledge.
~\cite{cite:ICLR20ANIL} find that feature reuse of the backbone is the predominant reason for efficient learning on downstream tasks with MAML. They thus propose the Almost No Inner Loop algorithm, which performs inner loop updates and task adaptation only on the task-specific head layer. Another limitation of MAML is that the fixed meta knowledge is globally shared by all tasks. To break this, Latent Embedding Optimization~\citep{cite:ICLR19LEO} performs gradient-based meta-learning in a low-dimensional latent space, and learns data-dependent latent embedding as meta knowledge to generate target model parameters.
\cite{cite:ICML19HierarchicallyMetaLearning} perform Hierarchically Structured Meta-Learning over hierarchical tasks based on clustering structures and learns to tailor transferable meta knowledge to different tasks.

While meta-learning methods enable fast model adaptation across tasks, they are weak in transferring to data from different domains, and some sophisticated methods even perform worse than standard pre-training baselines~\citep{cite:ICLR19ACloserLook}. Thus, Omni-Training
~\citep{cite:Arxiv21OmniTraining} incorporates both standard pre-training and meta-training in a framework with a tri-flow architecture to equip the pre-trained model with both domain transferability across different distributions and task transferability for fast adaptation across related tasks.

\subsubsection{Causal Learning}
\label{sec:causal_learning}
It remains difficult for supervised pre-training to obtain a transferable representation that generalizes well to an out-of-distribution (OOD) domain~\citep{cite:Turing21DeepLearningForAI}. In contrast, humans have the ability to adapt to different domains or new environments. Causal learning aims to pursue such kind of \emph{extrapolated} transferability in the pre-training stage.

\begin{figure}[!t]
	\centering
	\includegraphics[width=0.5\textwidth]{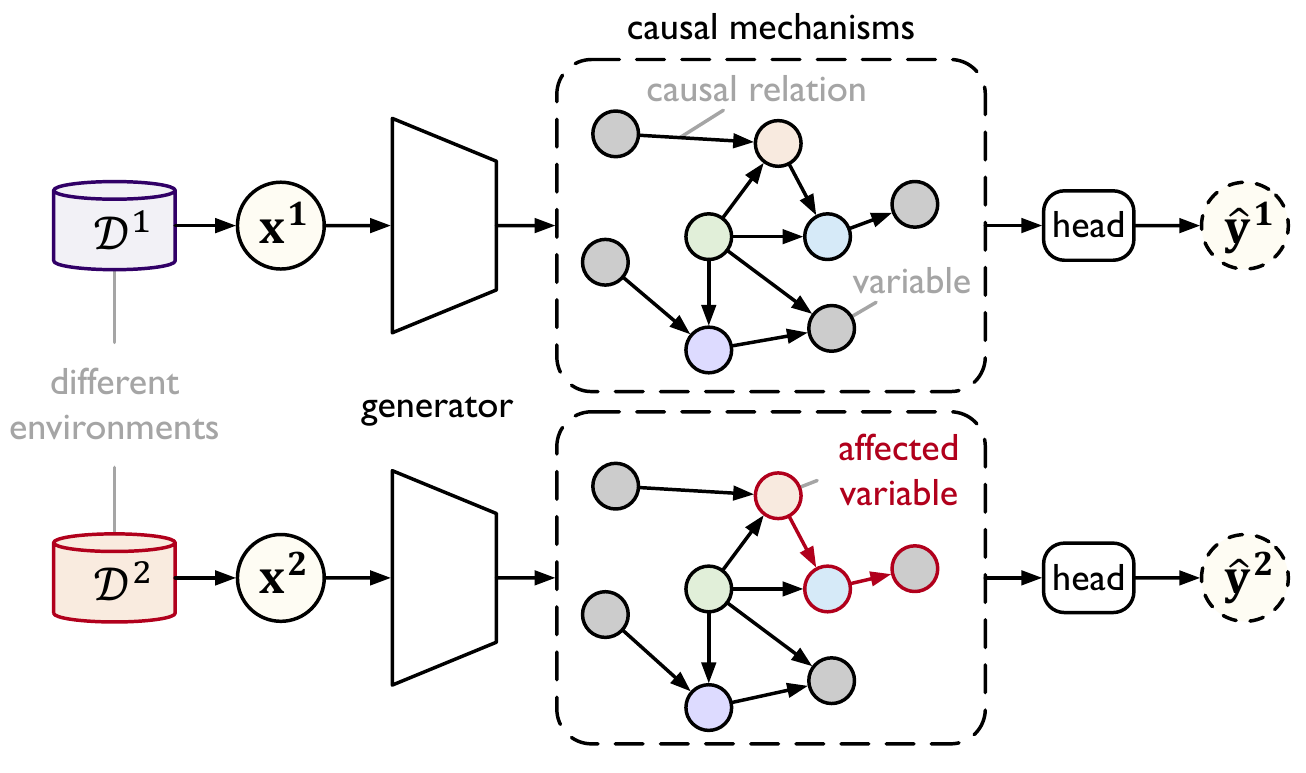}
	\caption{Causal mechanisms consider a set of observations or variables as the vertices of a directed acyclic graph, where each edge corresponds to a mechanism of direct causation.
		Causal learning seeks a model with variables governed by certain causal mechanisms, and if the environment or distribution changes, only part of the causal mechanisms will be affected.}
	\label{fig:causal_learning}
	\vspace{-10pt}
\end{figure}

The core idea of causal learning is to equip the model with some \textit{causal mechanisms} that capture independent and disentangled aspects of the complex real-world distributions. When the distribution changes, only one or several causal mechanisms change, with others remaining invariant, which could result in better \textit{out-of-distribution} (OOD) generalization.
The causal mechanisms are described by Structural Causal Models. As shown in Figure \ref{fig:causal_learning}, causal mechanisms consider a set of variables as the vertices of a directed acyclic graph, and each edge represents a mechanism of direct causation in that the parents directly affect the assignment of the child. This induces a canonical factorization of the joint distribution of these variables into the disentangled distribution of them conditioned on their parents. The independent causal mechanism principle states that given its mechanism, the conditional distribution of each variable does not inform or influence the other mechanisms~\citep{cite:ICML12OnCausalandAntiCausal, cite:Book17ElementsofCausalInference}. This implies that small distribution changes should only affect the causal mechanisms along with the disentangled factorization in a sparse and local way \citep{cite:ProceedingIEEE21Causal}, thereby enabling transferability towards different distributions. The key problem of causal learning is to obtain the variables governed by independent causal mechanisms. One way is to explicitly introduce independence with the \emph{modular} models. Another common practice is to leverage the \emph{invariance} assumption that causal relationships remain invariant across distributions.

\textbf{Modular Model.}
Recurrent Independent Mechanism (RIM) \citep{cite:ICLR21RIM} takes a modular model composed of several modules of different functions, where each module is a recurrent cell such as LSTM or GRU~\citep{cite:Arxiv14GRU} and represents a causal mechanism.
To obtain independence in distinct modules, RIM introduces attention between the hidden states of each module and the current inputs. For specific inputs, only the most relevant modules with larger attention are activated and updated, which forms competition between different modules and encourages their independence. RIM is shown to capture independent causal mechanisms and generalize well over different temporal patterns.

\textbf{Invariant Learning.}
The invariance assumption indicates that the conditional probability of the target output given its direct cause should be invariant across all environments or distributions.
Invariant Causal Prediction (ICP) \citep{cite:JRSS16ICP} uncovers independent causal mechanisms by performing a statistical test to find the subset of the variables satisfying the invariance assumption.
Invariant Risk Minimization (IRM) \citep{cite:Arxiv19IRM} extends this idea to representation learning and learns a good representation such that the conditional probability of the target output given the representation should be invariant across training environments.
Formally, given a data representation $\psi:\mathcal{X}\rightarrow\mathcal{Z}$ and training environments $\mathcal{E}^\text{tr}$, the conditional probability between the representation and the output is invariant if there is a classifier $h:\mathcal{Z}\rightarrow\mathcal{Y}$ simultaneously optimal for all the environments.
This can be formalized as the following constrained optimization problem,
\begin{equation}
	\mathop{\min}_{\psi:\mathcal{X}\rightarrow\mathcal{Z},{h}:\mathcal{Z}\rightarrow\mathcal{Y}} \  \sum_{e\in\mathcal{E}^\text{tr}} \epsilon^{e}(h\circ\psi), \quad \text{subject to  }  h\in \mathop{\arg\min}_{\bar{h}:\mathcal{Z}\rightarrow\mathcal{Y}}\epsilon^{e}(\bar{h}\circ\psi), \ \text{for all} \  e\in\mathcal{E}^\text{tr},
\end{equation}
where $\epsilon^{e}(h\circ\psi)$ refers to the expected error of the predictor $h\circ\psi$ on the environment $e$.
The transferability across environments relies on how the invariance across training environments implies invariance across all environments. Thus, the diversity of training environments is important for gaining transferability. IRM can be extended to complex situations where the causal relations are defined on some latent variables that need to be extracted from data.

\subsection{Unsupervised Pre-Training}
\label{sec:unsupervised_pre_training}

Being a canonical successful approach, supervised pre-training still requires a large amount of labeled data which are expensive to annotate and only available in certain fields. This hinders pre-training on huge-scale data and limits its transferability to particular tasks.
To break this shackle, unsupervised learning~\citep{cite:BengioUnsupervised}, typically in the form of self-supervised learning, is used for pre-training on very large unlabeled data to acquire generally transferable knowledge.
To improve the transferability on downstream tasks, it is crucial to design a proper self-supervised task for pre-training. According to the type of task, we can divide common unsupervised pre-training methods into generative learning and contrastive learning, which will be discussed in Sections \ref{sec:generative_pretraining} and \ref{sec:contrastive_pretraining} respectively.

\subsubsection{Generative Learning}
\label{sec:generative_pretraining}
\begin{figure}[htbp]
	\centering
	\includegraphics[width=0.55\textwidth]{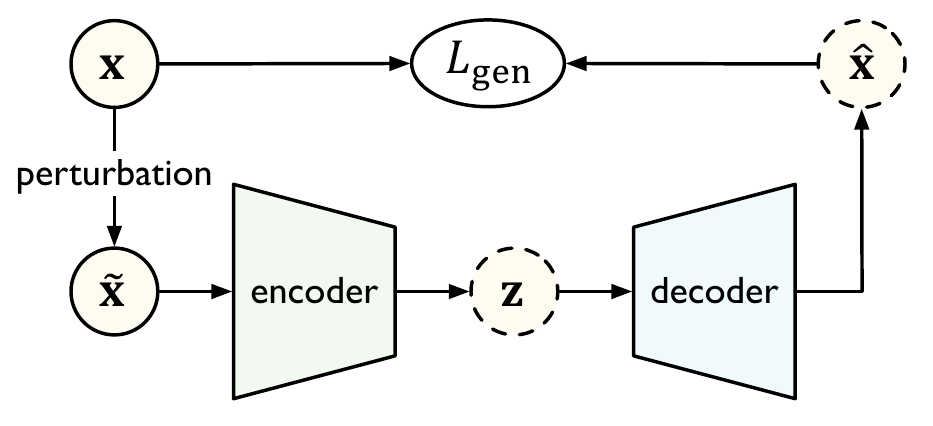}
	\vspace{-10pt}
	\caption{
		Generative pre-training tries to reconstruct the original input $\mathbf{x}$ from a perturbed input $\tilde{\mathbf{x}}$. The generative learning task shall encourage the learned representation $\mathbf{z}$ to capture the intrinsic and transferable explanatory factors from the data.
	}
	\label{fig:pretrain_generative}
\end{figure}

Generative learning is underpinned by the idea of learning to generate data distribution $P(\mathbf{X})$ for unsupervised pre-training.
It aims to learn the intrinsic representation in data and has been commonly used for pre-training deep neural networks~\citep{cite:NIPS07GreedyLayerWiseTraining}.
As shown in Figure \ref{fig:pretrain_generative},
we employ an encoder $f_\theta$ that maps the perturbed input $\tilde{\mathbf{x}}$ into a latent representation $\mathbf{z}=f_{\theta}(\tilde{\mathbf{x}})$ and a decoder $g_{\theta}$ that maps the representation back to derive a reconstructed version of the input $\widehat{\mathbf{x}}=g_{\theta}(\mathbf{z})$. The model is then optimized by minimizing the reconstruction error $L_\text{gen} (\widehat{\mathbf{x}}, \mathbf{x})$.
Most generative pre-training methods are based on two models: \emph{Autoregressive} Model, which generates future inputs given only past inputs,
and \emph{Autoencoding} Model, which generates full inputs given partial inputs.

\begin{figure}[!b]
	\centering
	\includegraphics[width=1.0\textwidth]{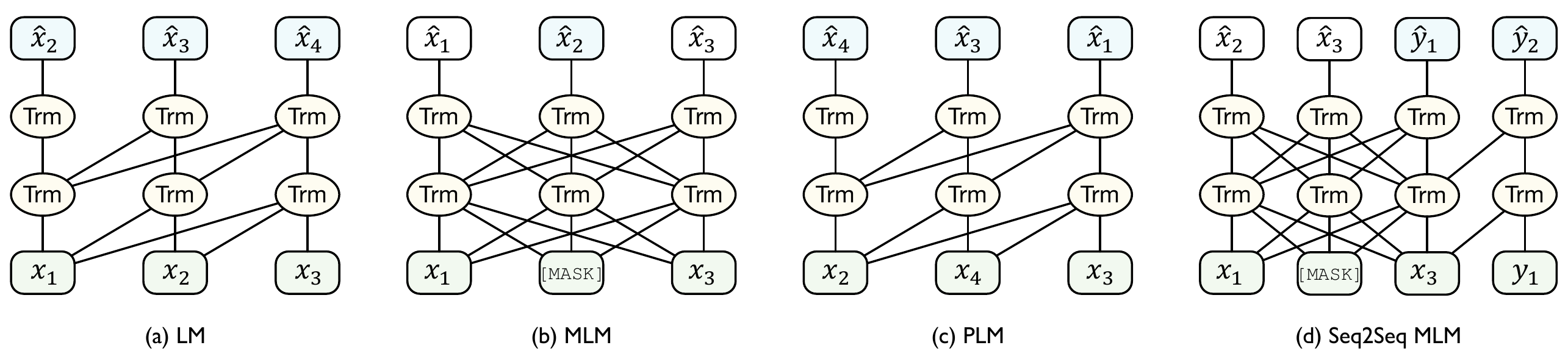}
	\vspace{-10pt}
	\caption{Attention visibility in Transformer (Trm) for language models. (a) LM maximizes the probabilities of all words conditioned on their previous words. (b) MLM maximizes the probabilities of random masked words conditioned on all unmasked words. (c) PLM permutes the original sequence and then performs autoregression. (d) Seq2Seq MLM encodes the input masked sequence $x$ and then decodes the output masked tokens $y$ sequentially.}
	\label{fig:LM}
\end{figure}

\textbf{Autoregressive Model} approximates the distribution of a sequence by predicting each entry conditioned on its previous context, which is called Language Modeling (LM) task in NLP.
As shown in Figure~\ref{fig:LM}, given a text sequence $\mathbf{x}_{1:T}=[x_1, x_2, ..., x_T]$, the learning objective of LM is to maximize the conditional probability of each entry $x_t$,
\begin{equation}
	\max_\theta \sum_{t=1}^T \log P_\theta(x_t|x_{t-k},\cdots,x_{t-1}),
\end{equation}
where $k$ is the size of the context window  and $\theta$ is the parameter of the neural network. Generative Pre-Training (GPT)~\citep{cite:GPT} explores unsupervised pre-training of Transformer with LM on the BooksCorpus~\citep{cite:ICCV15BookCorpus} dataset with over $7000$ unpublished books. This equips the model with great transferability to various NLP tasks, such as question answering, commonsense reasoning, and so on.
The advantage of LM is that it models the context dependency while the drawback is that it only encodes contextual information from one direction, yet contextual representations encoded in both directions may be more suitable to many downstream tasks, such as natural language inference.

\textbf{Autoencoding Model} approximates the data distribution by generating original data from encoded representations. \cite{cite:ICML08DenoisingAE} hypothesize that a good representation should also be robust to partial corruption of the input. Thus Denoising Autoencoder~\citep{cite:ICML08DenoisingAE} is trained to reconstruct the original input $\mathbf{x}$ with the corrupted input $\tilde{\mathbf{x}}$.
Inspired from Denoising Autoencoder, BERT \citep{cite:NAACL19BERT} adopts the Masked Language Modeling (MLM) task as a pre-training task to overcome the drawback of the unidirectional LM.
As shown in Figure~\ref{fig:LM}, MLM first randomly masks out some tokens $m(\mathbf{x})$ from the input sentences $\mathbf{x}$ with a special \texttt{[MASK]} token and then trains the models to predict the masked tokens by the rest of the tokens $\mathbf{x}_{\setminus m(\mathbf{x})}$,
\begin{equation}
	\max_\theta \sum_{x\in m(\mathbf{x})} \log P_\theta (x|\mathbf{x}_{\setminus m(\mathbf{x})}).
\end{equation}
Masked pre-training has also been used in many other areas. For instance, Masked Autoencoders (MAE) \citep{cite:Arxiv21MAE}  pre-trains vision transformers on large-scale unlabeled image datasets using the image generation task.
The difficulty is that the signals are highly redundant in images, thus it is hard for generative tasks, such as filling a few missing pixels, to capture high-level knowledge from data. To tackle this issue, MAE randomly masks a very large portion of patches, forcing the model to go beyond low-level understanding and reconstruct the whole image based on a small subset of visible patches, which improves its transferability to semantic-level tasks.
For another instance, to pre-train Graph Neural Network (GNN) \citep{cite:ICLR18GNN} for transferable representations, Attribute Masking \citep{graph_pretrain} conceals node or edge attributes and asks GNNs to predict those attributes based on neighboring structures, which can capture the regularities of attributes distribution over different graph structures, such as the chemistry rules in molecular graphs, and improve transferability on the downstream node or edge classification tasks.

\textbf{Combining Autoregressive and Autoencoding Models.}
In MLM, some special tokens, such as \texttt{[MASK]}, are only used in pre-training while absent in the downstream tasks, leading to the mismatch between the pre-training phase and the fine-tuning phase.
To mitigate this discrepancy, Permuted Language Modeling (PLM)~\citep{cite:NIPS19XLNet}
randomly samples a permutation of the sequence and then performs autoregression on the permuted sequence to predict the last few tokens.
To explore the limits of transferability of knowledge gained in different generative pre-training methods, T5~\citep{cite:JMLR20T5} unifies all text-based language tasks into the text-to-text format and then adopts a Sequence-to-Sequence MLM (Seq2Seq MLM), where the encoder processes a masked sequence and the decoder sequentially generates the masked tokens in an autoregression manner.

The design of unsupervised pre-training tasks has a great influence on the transferability to the downstream tasks, thus many efforts have been made to optimize the pre-training tasks and exploit better training objectives.
RoBERTa~\citep{cite:Arxiv19Roberta} explores the under-training issue of BERT and highlights that training with more data, longer sequences, and dynamically changed masking patterns
helps the model transfer better.
Besides, MLM randomly masks out some independent words, which are the smallest semantic units in English but may not have complete semantics in other languages, such as Chinese.
Thus, ERNIE (Baidu) \citep{sun2019ernie} introduces entity-level and phrase-level masking, where multiple words that represent the same semantic meaning are masked. This achieves good transferability on Chinese NLP tasks.
To improve transferability to tasks where span selection is important, such as question answering and coreference resolution,
SpanBERT \citep{joshi2019spanbert} masks a random variable length of span in the text and trains the span boundary representations to predict the entire content of the masked span.
BART~\citep{lewis_bart:_2020} introduces more perturbation functions such as sentence permutation, document rotation, token deletion, and text infilling for more transferable pre-trained models.

The generative pre-training on large-scale data greatly improves the transferability of models and even enables \emph{few-shot} task transfer.
By scaling up the model size to $175$B and pre-training on the corpus over $500$GB, GPT-3 \citep{cite:NIPS20GPT3} obtains impressive transferability. Using only task demonstrations and a few examples, GPT-3 achieves better performance than prior state-of-the-art fine-tuning approaches on some tasks.
The success of GPT-3 comes from the fact that the web-scale corpus contains a vast amount of natural language sentences, which potentially demonstrate different tasks without explicit task symbols. A high-capacity language model trained on such data would perform unsupervised multi-task learning and absorb transferable knowledge to handle downstream tasks. The generative pre-training on large-scale data also improves the transferability across domains.
Multilingual BERT \citep{multilingual_bert} is pre-trained with MLM on Wikipedia texts from $104$ languages and then achieves great cross-lingual transferability in the downstream tasks, where each language can be considered as a domain. Further, XLM \citep{cite:Arxiv19XLM} introduces the translation language modeling task, which extends MLM to parallel bilingual sentence pairs, encouraging more transferable representations across language.

\subsubsection{Contrastive Learning}
\label{sec:contrastive_pretraining}
\begin{figure}[htbp]
	\centering
	\includegraphics[width=0.55\textwidth]{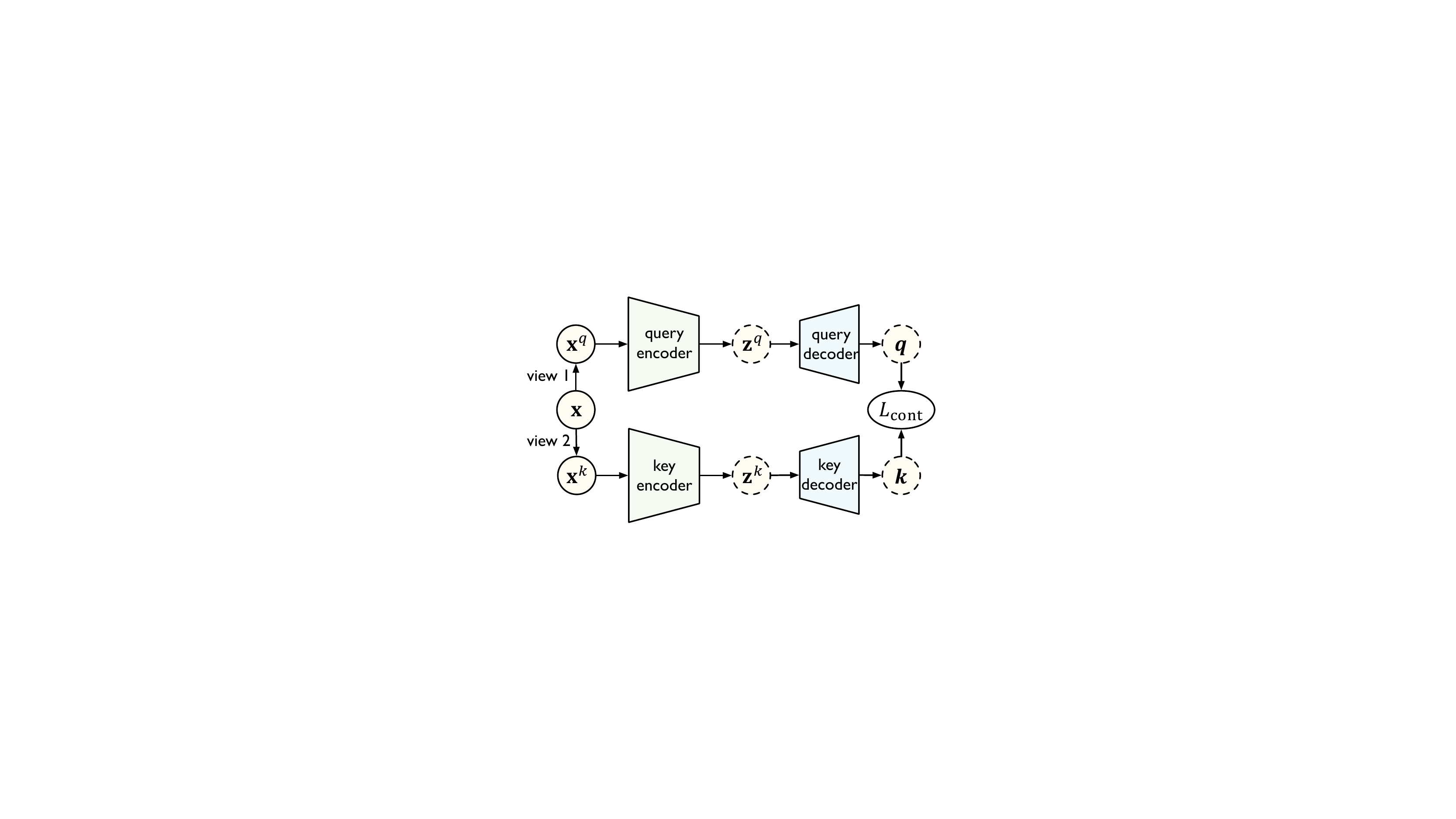}
	\vspace{-10pt}
	\caption{
		Contrastive pre-training aims to minimize the similarity between the query $\mathbf{q}$ and the key $\mathbf{k}$ that are generated from different views of the same data input $\mathbf{x}$.
	}
	\vspace{-10pt}
	\label{fig:pretrain_contrastive}
\end{figure}

Contrastive learning utilizes the idea of learning to compare for unsupervised pre-training.
As shown in Figure \ref{fig:pretrain_contrastive}, two different views, query $\mathbf{x}^q$ and key $\mathbf{x}^k$, are constructed from the original data $\mathbf{x}$.
Encoders will map different views into latent representations and decoders will further map the representation to the metric space. The model is learned by minimizing the distance between query and key of the same instance.
We will review three typical contrastive learning methods widely used in pre-training: Mutual Information Maximization (uses the global context and the local features as different views), Relative Position Prediction (uses different local components as different views), and Instance Discrimination (uses data augmentations to generate different views of the same instance). Different ways of generating and comparing different views encourage these methods to respectively capture the \textit{global-local} relation, \textit{local-local} relation and \textit{global-global} relation of the training data.

\paragraph{Mutual Information Maximization.}
Deep InfoMax \citep{cite:ICLR19DeepInfoMax} aims to acquire transferable representations from the relation between the high-level global context and the low-level local features.
Given input $\mathbf{x}$, Deep InfoMax learns an encoder $\psi$ to maximize the mutual information between its input and output of the same instance. The mutual information can be estimated and bounded by training a discriminator to distinguish between their joint distribution and the product of their marginals. Using Noise-Contrastive Estimation (NCE), the training objective of Deep InfoMax becomes,
\begin{equation}
	\max_\psi \mathbb{E}_{\mathbf{x}\sim\mathcal{U}}\left[ D(\mathbf{x},\psi(\mathbf{x})) - \mathbb{E}_{\mathbf{x}' \sim \widetilde{\mathcal{U}}} \Big( \log \sum_{\mathbf{x}'}   e^{D(\mathbf{x}',\psi(\mathbf{x}))} \Big) \right],
\end{equation}
where $\mathbf{x}$ is the input sampled from the training distribution $\mathcal{U}$ of upstream task, $\mathbf{x}'$ is another input sampled from $\widetilde{\mathcal{U}}=\mathcal{U}$, and $D$ is the discriminator to distinguish between the joint distribution and the product of marginals.
A parallel work, Contrastive Predictive Coding (CPC)~\citep{cite:Arxiv18CPC}, also maximizes the mutual information between pairs of global representation and local representation. Given a sequence input, CPC processes it with an encoder and summarizes the results into a context by an autoregression model. Then it maximizes the mutual information between the summarized context and the hidden representation of the future observation in the sequence, which guides the learned representations to capture information for predicting future samples.

Mutual information maximization has been used to obtain pre-trained models on many data formats, such as Deep InfoMax on image data and CPC on sequence data.
On graph data, Deep Graph Infomax \citep{cite:ICLR19GraphInfoMax} maximizes the mutual information between a node's local representations and the k-hop neighborhoods' context representations.
On multimodal data, Contrastive Language-Image Pre-training (CLIP) \citep{cite:Arxiv21CLIP}
maximizes the mutual information between the image and the corresponding text in a multimodal embedding space. After training with a large-scale dataset of image-text pairs from the Internet, it enables the \emph{zero-shot} transfer of the model to downstream tasks, competitive with the prior task-specific supervised models.

\paragraph{Relative Position Prediction.}
Next Sentence Prediction (NSP) \citep{cite:NAACL19BERT}, which is first introduced in BERT, acquires transferable representations from the relation between \emph{local parts}.
Specifically, NSP uses a binary classifier to predict whether two sentences are coherent from the training corpus, aiming to enhance the transferability to tasks with multiple sentences, such as question answering and natural language inference.
However, subsequent work questions the necessity of NSP tasks \citep{cite:NIPS19XLNet, roberta} and \cite{cite:ICLR20Albert} conjecture that NSP only forces the model to learn topic prediction, rather than more difficult coherence prediction.
Since inter-sentence coherence is important to many downstream tasks, ALBERT~\citep{cite:ICLR20Albert} introduces a sentence-order prediction task, where two consecutive segments from the same document are taken as positive examples, and the same segments with order swapped are taken as negative examples.
Similar ideas are also explored in vision, where the pre-training task is to predict relative positions of two patches from an image \citep{doersch2015unsupervised}.

\paragraph{Instance Discrimination.}
InstDisc~\citep{wu2018unsupervised} aims to learn transferable representations from the relation between \emph{instances}.
Given $n$ instances, an encoder $\psi$ is trained to distinguish each instance from others, i.e., minimize the distance between the query $\mathbf{q}$  and key $\mathbf{k}_+$ from the same instance (also called positive samples) and maximize the distance between that of different instances (also called negative samples),
\begin{equation}
	\min_\psi  -\log \frac{\text{exp}(\mathbf{q} \cdot \mathbf{k}_+ / \tau)} {\sum_{j=0}^{K} \text{exp}(\mathbf{q} \cdot \mathbf{k}_j / \tau)},
\end{equation}
where $\tau$ is a temperature hyper-parameter and the sum is over one positive and $K$ negative samples.
Note that the computation of the features of all samples and the non-parametric softmax is costly especially when the number of training instances $n$ is extremely large. To tackle this issue, negative sampling is used to approximate the softmax, i.e., $K<n$.

\begin{figure}[!t]
	\centering
	\includegraphics[width=1.0\textwidth]{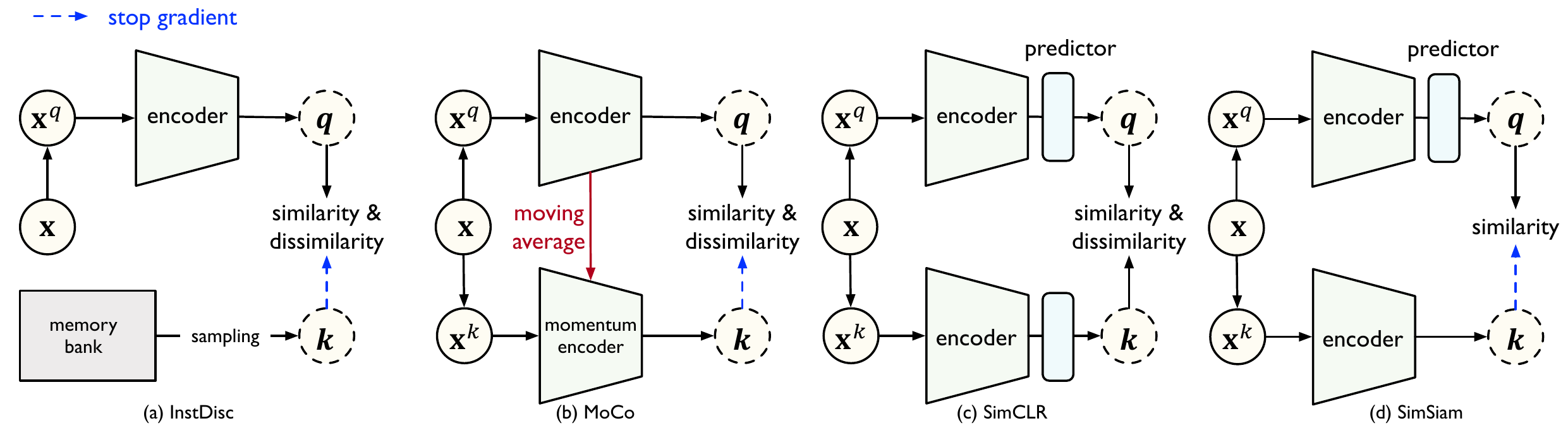}
	\vspace{-20pt}
	\caption{
		Comparison of different contrastive learning mechanisms.
		(a) InstDisc samples the keys from a memory bank.
		(b) MoCo encodes the new keys on the fly by a momentum encoder and maintains a queue of keys.
		(c) SimCLR encodes the keys and queries in the same batch with the same encoder and adds a nonlinear predictor to improve the representation.
		(d) SimSiam applies an MLP predictor on one side and applies a stop-gradient operation on the other side, and maximizes the similarity in two views without using negative pairs.
	}
	\vspace{-10pt}
	\label{fig:Instance_Discrimination}
\end{figure}

The discriminability of representations to contrast one instance from another instance is closely related to the transferability on downstream tasks. Thus, many efforts have been made to increase the number and improve the quality of keys.
As shown in Figure \ref{fig:Instance_Discrimination}, InstDisc \citep{wu2018unsupervised} uses a memory bank to store the latest updated representations for each key, which increases the number of negative samples, yet may result in less consistent representations. Momentum Contrast (MoCo) \citep{cite:CVPR20MoCo} maintains a dynamic queue of encoded features to enlarge the size of negative samples and encodes the keys with a momentum-updated encoder, which increases encoding consistency between different samples in the queue and improves the quality of keys.
The way how the positive samples and negative samples are constructed is also important for transferability. Contrastive Multiview Coding (CMC)~\citep{cite:ECCV20CMC} takes multiple views, rather than multiple augmentations, of the same instance as positive samples and achieves better transferability.
SimCLR~\citep{cite::ICML20SimClr} emphasizes that data augmentations play a crucial role in implicitly defining different pretext tasks, and the composition of stronger augmentations leads to better transferability even without the need for a memory bank or a queue.
The introduction of negative samples is to avoid trivial solutions that all outputs collapse to a constant. However, BYOL \citep{NEURIPS2020_f3ada80d} finds that when maximizing the similarity between two augmentations of one image, negative sample pairs are not necessary.
Further, SimSiam~\citep{cite:CVPR21SiameseRepresentation} finds that momentum encoder is also not necessary while a stop-gradient operation applied on one side is enough for learning transferable representations.

Compared with supervised pre-training, contrastive pre-training leads to competitive performance on downstream classification tasks and even better performance on various other downstream tasks, such as object detection and semantic segmentation.
To explain the stronger transferability of contrastive pre-training, ~\cite{cite:ICLR21InstanceTransfer} observe that standard supervised pre-training usually transfers high-level semantic knowledge, while contrastive pre-training usually transfers low-level and mid-level representations. When the target tasks are different from the supervised pre-trained tasks, the supervised pre-training methods have the risk of over-fitting the semantic discriminative parts of objects defined by the class labels, which
hurts the transferability. On the contrary, the contrastive pre-training tasks lead to more holistic modeling of the objects, which relaxes the task misalignment issue and achieves better transferability for widespread downstream tasks.

\subsection{Remarks}

While standard supervised pre-training is well established, its transferability also depends on the relationship between the pre-training task and the target task, and no pre-training task can dominate all downstream tasks.
\cite{cite:ICCV19RethinkingPretraining} show that compared with the random initialization, supervised pre-training on ImageNet only speeds up the convergence of object detection on the COCO dataset, but does not lead to better final accuracy. \cite{cite:NIPS19Transfusion} observe similar phenomena in medical imaging, where training lightweight models from scratch perform comparably with transferring from ImageNet pre-trained models.
\cite{abnar2021exploring} explore the limits of large-scale supervised pre-training and find that as the pre-training accuracy increases by scaling up data, model size and
training time, the performance of downstream tasks gradually saturates and there are even some extreme scenarios where performance on pre-training and downstream tasks are at odds with each other.
These controversial results encourage us to \emph{rethink} the common practice of supervised pre-training and design new supervised pre-training strategies for specific fields, especially when large gaps exist between the pre-training and target tasks.

\begin{table}[htbp]
	\scriptsize
	\centering
	\caption{Comparison between different pre-training methods.}
	\label{table:compare_pretraining}
	\begin{threeparttable}
		\begin{tabular}{lcccc}
			\toprule
			Method                & Modality Scalability$^1$   & Task Scalability$^2$       & Data Efficiency$^3$        & Labeling Cost$^4$          \\
			\midrule
			Standard Pre-Training & $\bigstar\bigstar\bigstar$ & $\bigstar\bigstar$         & $\bigstar\bigstar\bigstar$ & $\bigstar$                 \\
			\midrule
			Meta-Learning         & $\bigstar\bigstar\bigstar$ & $\bigstar$                 & $\bigstar$                 & $\bigstar$                 \\
			\midrule
			Causal Learning       & $\bigstar\bigstar$         & $\bigstar$                 & $\bigstar$                 & $\bigstar$                 \\
			\midrule
			Generative Learning   & $\bigstar\bigstar$         & $\bigstar\bigstar\bigstar$ & $\bigstar\bigstar\bigstar$ & $\bigstar\bigstar\bigstar$ \\
			\midrule
			Contrastive Learning  & $\bigstar$                 & $\bigstar\bigstar\bigstar$ & $\bigstar\bigstar\bigstar$ & $\bigstar\bigstar\bigstar$ \\
			\bottomrule
		\end{tabular}
		\begin{tablenotes}
			\scriptsize
			\item[1] Modality Scalability: whether models can be pre-trained on various modalities, such as text, graph.
			\item[2] Task Scalability: whether pre-trained models can be easily transferred to different downstream tasks.
			\item[3] Data Efficiency: whether stronger transferability can be yielded from large-scale pre-training.
			\item[4] Labeling Cost: whether relies on manual data labeling.
		\end{tablenotes}
	\end{threeparttable}
\end{table}

Table \ref{table:compare_pretraining} compares pre-training methods from four perspectives: modality scalability, task scalability, data efficiency, and labeling cost.
Though meta-learning enables fast adaptation to new tasks, it mainly considers related tasks such as reinforcement learning under environments with small changing factors, while standard pre-training can transfer to broader task gaps such as from image classification to object detection.
Besides, the existing meta-learning and causal learning methods are empirically verified only on small datasets, and it remains unclear whether they can acquire stronger transferability via pre-training on large-scale data.
Despite the promising performance without manual labeling, unsupervised pre-trained models require a large number of gradient steps for fine-tuning to downstream tasks.
Also, strong data augmentations are required by contrastive learning to gain transferability, but they are not easy to design in other modalities, such as text and graphs.
Finally, the design of unsupervised pre-training tasks remains heuristic, lacking solid analysis on how the task shift is bridged and what enables the transferability of these models.

Acquiring transferability only through the pre-training stage may limit our horizon.
As the shift in tasks and domains naturally exists between the pre-training and adaptation stages, many pre-training methods are tailored to adaptation.
Unsupervised pre-training aims to improve the transferability to downstream tasks by exploring different kinds of self-supervised tasks to reduce the task discrepancy between pre-training and adaptation, or by enlarging the size and diversity of the upstream data to reduce the upstream-downstream discrepancy.
The distribution shift commonly tackled by domain adaptation~\citep{DANN} also influences the transferability of the pre-trained model.
For instance, the data distribution in a specific domain, such as biological and scientific literature, is quite different from that in the general pre-training domain and may degrade transferability, thus BioBert \citep{lee2020biobert} and SciBERT  \citep{Beltagy2019SciBERT} perform pre-training on domain-specific data to improve the transferability on domain-specific tasks.

\section{Adaptation}
\label{sec:adaptation}

While pre-training on large-scale datasets can gain transferable knowledge in deep models, performing task adaptation with the target data is still necessary for most applications, as the target task is usually different from the pre-training task.
When the labeled data for the target task is not enough, domain adaptation from a related source domain with labeled data to boost the performance on the target domain is also necessary in many applications.
We will review task adaptation and domain adaptation in Sections \ref{sec:task_adaptation} and \ref{sec:domain_adaptation} respectively.

\subsection{Task Adaptation}
\label{sec:task_adaptation}

In task adaptation, there exist a pre-trained model $h_{\theta^0}$ and a target domain $\widehat{\mathcal{T}}=\{\mathbf{x}_i, \textbf{y}_i \}_{i=1}^{m}$ of $m$ labeled samples.
The goal is to find a hypothesis $h_{\theta}:\mathcal{X} \mapsto \mathcal{Y}$ in the space $\mathcal{H}$ using the pre-trained model and target data to achieve a low generalization risk $\epsilon_{\mathcal{T}}(h_{\theta})$.
In general, there are two simple ways to \emph{adapt} a pre-trained model to the downstream tasks: feature transfer and fine-tuning. Feature transfer freezes the weights of the pre-trained models and trains a linear classifier on top of that.
In contrast, fine-tuning uses the pre-trained models to initialize the target model parameters and update these parameters during training.
Feature transfer is fast in training and efficient in parameter storage, yet fine-tuning yields better performance \citep{cite:NIPS14HowTransferable}, and has become a common practice for task adaptation in both vision and NLP \citep{cite:CVPR14RichFeature, cite:NAACL19BERT}.

\begin{figure}[!b]
	\centering
	\includegraphics[width=1\textwidth]{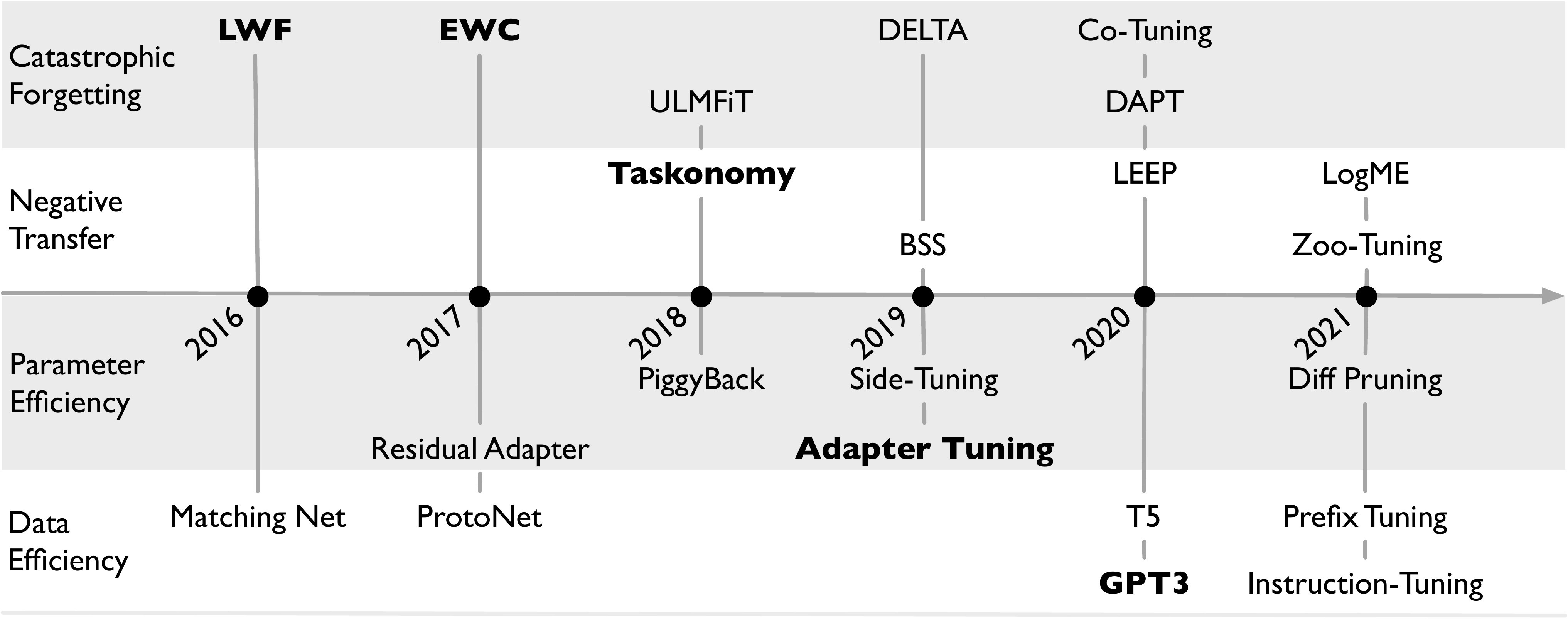}
	\vspace{-10pt}
	\caption{Cornerstones of task adaptation methods for \emph{applying} transferable knowledge.}
	\label{fig:TA_overview}
\end{figure}

\textit{Vanilla fine-tuning}, which tunes the pre-trained models by empirical risk minimization on the target data, has been widely used in various downstream tasks and scenarios.
However, vanilla fine-tuning still suffers from several issues, including \textit{catastrophic forgetting} and \textit{negative transfer}.
We will introduce how to alleviate these issues in Sections \ref{sec:CatastrophicForgetting} and \ref{sec:NegativeTransfer}.
Besides, as the parameters of deep models keep increasing and some of them reach billions or trillions, parameter efficiency and data efficiency have become increasingly important in task adaptation. We will give an introduction on how to explore the transferability in the pre-trained models to solve these problems in Sections \ref{sec:parameter_efficiency} and \ref{sec:data_efficiency}.
Overall, Figure \ref{fig:TA_overview} shows the progress made by the task adaptation algorithms to solve different problems.

\subsubsection{Catastrophic Forgetting}
\label{sec:CatastrophicForgetting}

Catastrophic forgetting, which was first studied in \textit{lifelong learning}, refers to the tendency of neural networks to lose knowledge acquired from previous tasks when learning new tasks \citep{EWC}. In the fine-tuning scenario where labeled data is usually scarce, it will lead to the overfitting of models on the target data. This phenomenon is also called \textit{representational collapse}, i.e., the degradation of generalizable representations during the fine-tuning stages \citep{aghajanyan2020intrinsic}. The most simple way to avoid catastrophic forgetting might be selecting a small learning rate and adopting an early-stopping strategy, which avoids updating the parameters too much. However, this strategy may lead the model to falling into the local minimal, especially when there is a large gap between the pre-training parameters and the optimal parameters for the downstream task.

\cite{cite:NIPS14HowTransferable} find that the transferability of different layers is not the same --- the first layers learn general features, the middle layers learn semantic features and the last layers learn task-specific features. Thus, to make the model retain the knowledge acquired in the pre-training task and fit the target task well at the same time, different layers should not be treated the same. Specifically, the first layers should retain more pre-trained knowledge while the last layers should adapt more to the downstream tasks.
Inspired by this finding, DAN \citep{DAN} sets the learning rate of the task-specific head to be $10$ times larger than that of the lower layers, which is simple yet effective when the labeled data is scarce or the target domain is close with the pre-training domain.
ULMFiT \citep{ULMFiT} gradually unfreezes the model starting from the last layers to the first layers, which effectively retains general knowledge in the first layers.
To automatically determine which layers should be fine-tuned or frozen for each sample,
Spottune \citep{cite:CVPR19SpotTune} proposes a policy network that is deployed to output the routing decision based on the input of each sample and is jointly trained with the main model during fine-tuning.

\begin{figure}[htbp]
	\centering
	\includegraphics[width=0.8\textwidth]{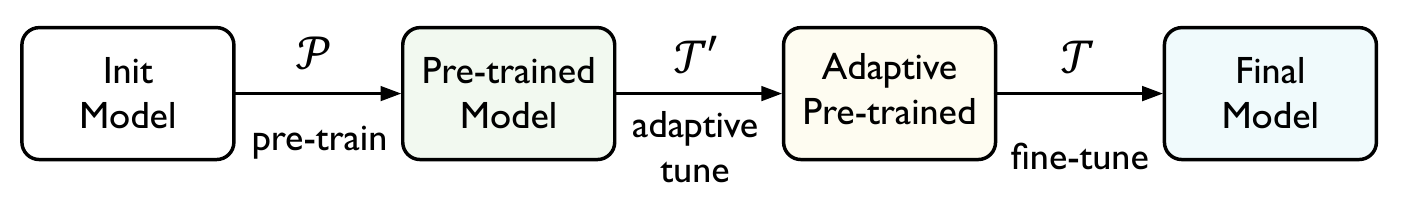}
	\caption{Domain Adaptive Tuning often consists of two consecutive steps: first, adaptive-tune on an auxiliary domain $\mathcal{T}'$ that is related to the target domain using the pre-training task; second, fine-tune on the target domain $\mathcal{T}$ using the target learning task. }
	\label{fig:adaptive_tuning}
\end{figure}

\textbf{Domain Adaptive Tuning}
reveals that an important source of catastrophic forgetting is the dataset shift between the pre-training and the target domain.
To bridge such shift, ULMFiT \citep{ULMFiT} and DAPT \citep{dontstoppretraining2020} first tune the pre-trained model on data related to the target domain, or simply data of the target task, with the pre-training task. Then they fine-tune the adaptive-tuned model on the target task (Figure \ref{fig:adaptive_tuning}).
Usually, the pre-training task is unsupervised, thus further pre-training with in-domain data can provide rich information about the target data distribution for better task adaptation with no additional labeling costs.
The two stages, domain adaptive tuning and regular fine-tuning, in the above methods can also be done jointly via multi-task learning. SiATL \citep{chronopoulou-etal-2019-embarrassingly} adds an auxiliary language model loss to the task-specific optimization function, which alleviates catastrophic forgetting and learns task-specific features at the same time.

\begin{figure}[htbp]
	\centering
	\includegraphics[width=1.0\textwidth]{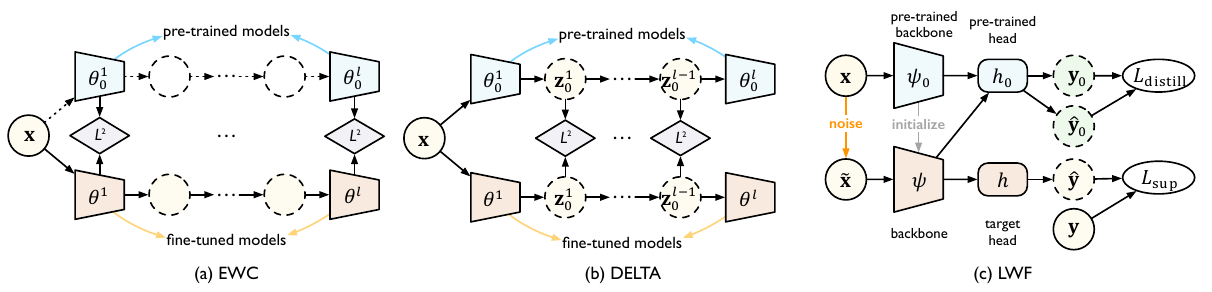}
	\caption{
		Regularization methods for task adaptation that avoids catastrophic forgetting.
		Blue: pre-trained parameters; Red: fine-tuned parameters. (a) EWC regularizes the \textit{parameters} of the new models $\theta$ and that of the pre-trained models $\theta_0$ with weighted $L_2$-penalty. (b) DELTA regularizes the \textit{feature maps} of the new models $\mathbf{z}$ and that of the pre-trained models $\mathbf{z}_0$. (c) LWF enforces the \textit{output} of the old tasks $\widehat{\mathbf{y}}_0$ close to the initial response $\mathbf{y}_0$.
	}
	\label{fig:EWC}
\end{figure}

\textbf{Regularization Tuning} is another way to prevent the models from deviating far away from the pretrained ones. The optimization objective with a general regularization is
\begin{equation}
	\min_\theta \sum_{i=1}^{m} L(h_{\theta}(\mathbf{x}_i), \mathbf{y}_i) + \lambda \cdot \Omega({\theta}),
\end{equation}
where $L$ is the loss function, $\Omega$ is a general form of  regularization, and $\lambda$ is the trade-off between them. A typical regularization in supervised learning is $L_2$ penalty, $\Omega(\theta) = \frac{1}{2}||\theta||_2^2$,
which drives the weights $\theta$ to zero to control the model complexity.
Different from typical supervised learning, in fine-tuning, there exists a pre-trained model $h_{\theta^0}$ setting a reference that can be used to define the hypothesis space (Figure~\ref{fig:intrinsic}).
Thus, Elastic Weight Consolidation (EWC) \citep{EWC} constrains the
distance between  the weights of the pre-trained and fine-tuned networks (Figure \ref{fig:EWC}) to overcome catastrophic forgetting,
\begin{equation}
	\Omega(\theta) = \sum_{j} \frac{1}{2} F_j\left\| \theta_j - \theta^0_j \right\|_2^2,
\end{equation}
where $F$ is the estimated Fisher information matrix.
EWC is based on the assumption that the networks with similar weights should produce similar outputs. However, due to the complex structures of deep networks, similar parameters do not necessarily produce the same output, and the same output may also come from completely different model parameters.
Thus, DELTA \citep{cite:ICLR19Delta} constraints the behavior, i.e., the feature maps of the model by selecting the discriminative features with a supervised attention mechanism and regularizing the distance of these features between pre-trained and fine-tuned networks.
Learning Without Forgetting (LWF) \citep{LWF} constrains the output prediction of the model by encouraging the model's response for old tasks to keep the same throughout the fine-tuning process (Figure \ref{fig:EWC}). Regularization on the output often performs better than regularization on the parameters or the features, yet the latter two have better scalability and versatility to more complex downstream tasks.

An explanation for the effect of regularization is that it makes the hypothesis smoother.
Therefore, TRADES \citep{robust_accuracy_trade_off} and SMART \citep{SMART} directly enforce the smoothness of the hypothesis by encouraging the output of the model to not change much when injecting a small perturbation to the input,
\begin{equation}
	\Omega(\theta) = \sum_{i=1}^{m}\max_{|| {\widetilde{\mathbf{x}}_i}  - \mathbf{x}_i||_p \leq \epsilon} L_s (h_{\theta}({\widetilde{\mathbf{x}}_i}), h_{\theta}(\mathbf{x}_i)),
\end{equation}
where $\epsilon > 0$ is a small positive,  $L_s$ is the distance between two predictions, such as the symmetrized KL-divergence in classification and the squared loss in regression.

Apart from the training objective, an alternative regularization approach is through the parameter updating strategy. Stochastic Normalization \citep{cite:NIPS20StocNorm} randomly replaces the target statistics in the batch-normalization layer~\citep{BatchNorm} with their pre-trained statistics, which serves as an implicit regularization by avoiding over-depending on the target statistics.
Mixout \citep{lee2020mixout} randomly replaces part of the model parameters with their pre-trained weights during fine-tuning to mitigate catastrophic forgetting.
Child-Tuning \citep{xu2021raise} selects a subset of parameters (child network) by some criterion and only updates them during fine-tuning. In some senses, the above methods decrease the hypothesis space to preserve the transferability in pre-trained models.

\subsubsection{Negative Transfer}
\label{sec:NegativeTransfer}

Although the paradigm of pre-training and fine-tuning has been used in various downstream tasks, it does not necessarily produce a positive effect, which is known as \textit{negative transfer} \citep{Rosenstein2005ToTO}.
\cite{CharacterizingNegativeTransfer} propose to quantitatively measure the degree of negative transfer across different domains and we extend this idea to the paradigm of pre-training and fine-tuning.

\begin{definition}[Negative Transfer Gap]
	Let $h_\theta (\mathcal{U}, \mathcal{T})$ be a hypothesis obtained by adapting the model pre-trained from the upstream data $\mathcal{U}$ to the target data $\mathcal{T}$, and  $h_\theta (\emptyset, \mathcal{T})$ be a hypothesis obtained by training from scratch on $\mathcal{T}$, then negative transfer gap is defined as
	\begin{equation}
		NTG = \epsilon_{\mathcal{T}} (h_\theta (\mathcal{U}, \mathcal{T})) - \epsilon_{\mathcal{T}} (h_\theta (\emptyset, \mathcal{T})),
	\end{equation}
	and negative transfer occurs if $NTG$ is positive and vice versa.
\end{definition}

\begin{figure}[h]
	\centering
	\includegraphics[width=0.45\textwidth]{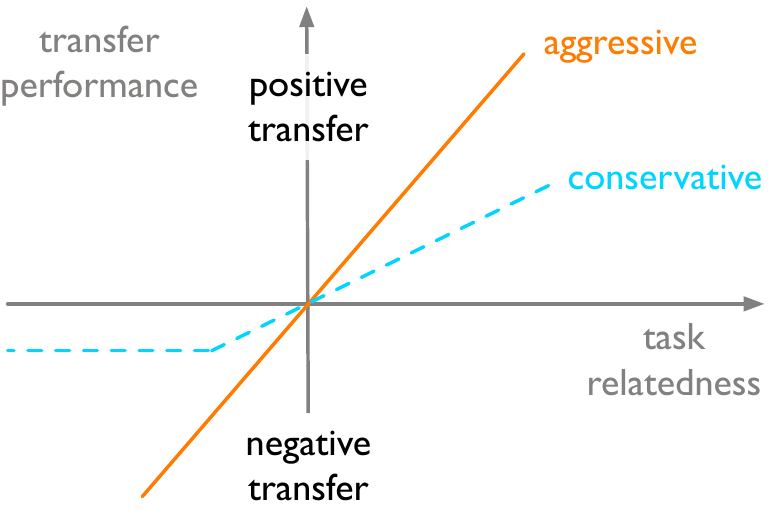}
	\vspace{-5pt}
	\caption{Dilemma to promote positive transfer and avoid negative transfer: Aggressive strategies that promote larger positive transfer will suffer from severer negative transfer; Conservative strategies can decrease negative transfer, yet lead to smaller positive transfer.}
	\label{fig:negative_transfer}
	\vspace{-10pt}
\end{figure}

First, negative transfer will happen when the relatedness between the upstream task and downstream task is not strong, e.g.~Next Sentence Prediction pre-training task will hurt token-level classification tasks \citep{cite:Arxiv19Roberta}.
Negative transfer will also happen when there is a large shift between the pre-training domain and the target domain, e.g.~for legal documents classification, pre-training only on legal documents is better than pre-training on more diverse datasets \citep{zhengguha2021}.
Second, negative transfer depends on the size of the labeled target dataset \citep{CharacterizingNegativeTransfer}. For example, \cite{cite:ICCV19RethinkingPretraining} empirically show that on large-scale object detection datasets (e.g.~COCO), ImageNet pre-training is not beneficial when training for enough iterations.
Third, negative transfer depends on the task adaptation algorithms. An ideal adaptation algorithm should promote positive transfer between related tasks while avoiding negative transfer between unrelated tasks.
In practice, however, these two goals are often contradictory and result in the \emph{dilemma}: approaches that promote larger positive transfer will suffer from severer negative transfer (Figure \ref{fig:negative_transfer}).

\paragraph{Enhancing Safe Transfer.} One way to avoid negative transfer is to recognize and reject harmful knowledge in the pre-trained model. \cite{cite:NIPS19BSS} observe that with sufficient training data, the spectral components with small singular values vanish during fine-tuning, indicating that small singular values correspond to detrimental pre-trained knowledge and may cause negative transfer. Thus BSS penalizes smaller singular values to suppress untransferable spectral components for safe transfer.
\cite{cite:ICML19LearningtoTransfer} meta-learns the weights determining which pairs of layers should be matched and to what extent the knowledge should be transferred, which rejects irrelevant information during transfer.
Zoo-tuning \citep{cite:ICML21ZooTuning} enables adaptive transfer from a zoo of models, which adaptively aggregates multiple pre-trained models to derive the target model using data-dependent gating mechanisms to highlight transferable parameters.
Another way to mitigate negative transfer of the pre-trained model is to fully explore the target data. Self-Tuning \citep{cite:ICML21SelfTuning} proposes a pseudo group contrastive mechanism to explore the intrinsic structure of the target data in the process of fine-tuning with standard supervised objective.

\paragraph{Choosing Pre-trained Models.}
With the fast development of deep learning, a large zoo of pre-trained models are available, thus a simpler way to avoid negative transfer is to select a model that is pre-trained on the upstream data/task relevant to the downstream data/task.
The most common practice to choose pre-trained models is based on rich past experience or through heavy experiments.
To facilitate faster selection, Taskonomy \citep{taskonomy} proposes a fully computational approach for explicitly modeling the relationship between $26$ different visual tasks.
Another more efficient strategy to select pre-trained models is to predict the transferability of pre-trained models.
LEEP \citep{cite:ICML20LEEP} constructs an empirical predictor by estimating the joint distribution over pre-trained labels and the target labels, and then uses the log expectation of the empirical predictor to measure the transferability.
LogME \citep{cite:ICML21LogME} proposes to predict the fine-tuning performance from the compatibility of features $\{\mathbf{z}_i=\psi (\mathbf{x}_i)\}_{i=1}^{m}$ and labels $\{\mathbf{y}_i\}_{i=1}^{m}$.
Still, these methods may underestimate strong but non-linear features.
\cite{cite:Arxiv21MAE} show that features from contrastive pre-training, such as MoCo v3 \citep{chen2021mocov3}, have higher linear probing accuracy while worse fully fine-tuning results than generative pre-training, such as MAE \citep{cite:Arxiv21MAE}, indicating that the linear separability of the pre-trained features is not the sole metric for evaluating transferability.

\begin{figure}[!t]
	\centering
	\includegraphics[width=0.9\textwidth]{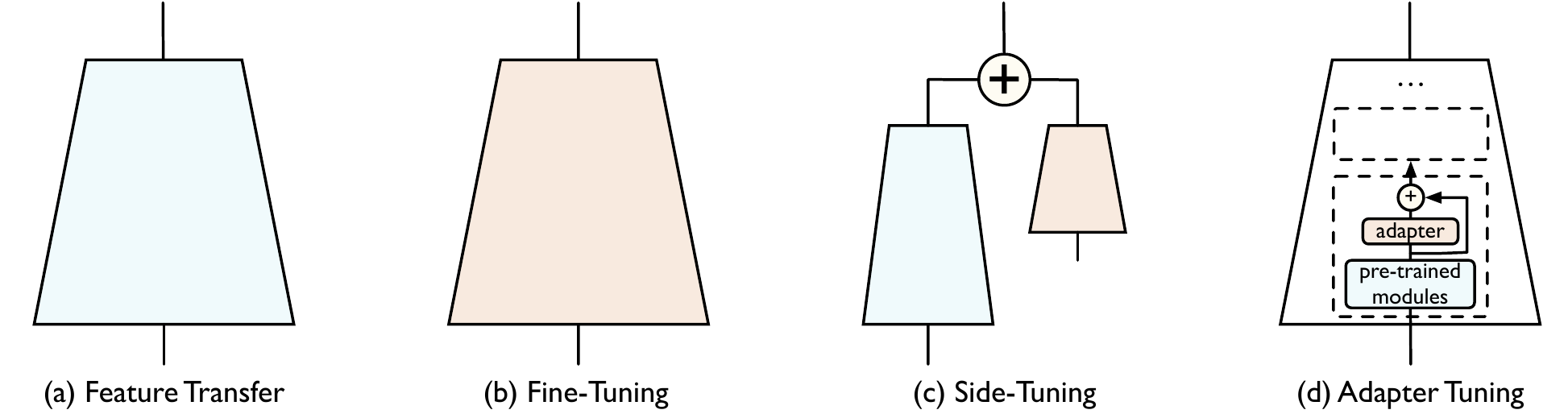}
	\vspace{-5pt}
	\caption{
		Comparison on how different adaptation methods freeze (blue) and tune (red) pre-trained parameters.
		(a) Feature transfer freezes all the pre-trained parameters.
		(b) Fine-tuning re-trains all the pre-trained parameters.
		(c) Side-tuning trains a lightweight conditioned side network that is fused with the fixed pre-trained network using summation.
		(d) Adapter-Tuning inserts adapter modules for tuning into each frozen pre-trained layer.
	}
	\label{fig:adapter_tuning}
\end{figure}

\subsubsection{Parameter Efficiency}
\label{sec:parameter_efficiency}

Fine-tuning large pre-trained models yields strong performances on many
downstream tasks \citep{cite:GPT, cite:NAACL19BERT}, yet it is not parameter efficient since it generates a full set of model parameters for each downstream task, which will cause unacceptable storage cost as the number of tasks increases. The simplest solution is \textit{Multi-task Learning} \citep{Caruana97multitasklearning}, i.e., fine-tuning a single model to solve multiple target tasks, which might be mutually beneficial to each other \citep{MaskRCNN, liu2019mt-dnn}. Yet when different target tasks are weakly related, multi-task learning will underperform fine-tuning for each task separately. Also, multi-task learning requires simultaneous access to all target tasks, which is not feasible in online scenarios where the target tasks arrive in sequence. Hereafter, we will introduce new tuning paradigms proposed to improve parameter efficiency.

\paragraph{Residual Tuning.}
Inspired by the fact that approximation to a difference is easier than the original function \citep{cite:CVPR16ResNet},
Side-Tuning \citep{sidetuning2019} adds a small side network $h_{\text{side}}$ to adapt the frozen pre-trained model $h_{\text{pretrained}}$ for the target task and obtain a combined model $h(x) = \alpha h_{\text{pretrained}}(x) + (1-\alpha) h_{\text{side}}(x)$, where $\alpha$ is a weight that changes during training.
When there is a big gap between the pre-trained model and the downstream task,
it may be difficult to learn the residuals of the entire model. Thus, Adapter Tuning \citep{houlsby2019parameter} inserts residual adapter modules into each frozen layer.
Residual Adapter was first introduced for learning multiple visual domains \citep{Rebuffi17} and consists of a skip connection, such that it is set as a near-identity function and will not impair the whole model when the training starts.
By choosing a much smaller amount of parameters for the adapters, Adapter Tuning can extend pre-trained models to new tasks without increasing much storage cost.
\cite{houlsby2019parameter} find that Adapter Tuning with only $3.6\%$ tunable parameters can match the performance of the fully fine-tuned BERT on the GLUE benchmark~\citep{GLUE}, revealing the great potential of this method.

\paragraph{Parameter Difference Tuning.}
While residual adapter tuning changes the model activations by adding new modules, parameter difference tuning extends the pre-trained models through a task-specific parameter difference vector,
\begin{equation}
	\theta_{\text{task}} = \theta_{\text{pretrained}} \oplus \delta_{\text{task}},
\end{equation}
where $\oplus$ is the element-wise addition function, $\theta_{\text{pretrained}}$ is the fixed pre-trained parameters and $\delta_{\text{task}}$ is the tuned task-specific difference vector. Instead of storing a copy of $\theta_{\text{task}}$ for every task, difference tuning only needs to store a single copy of $\theta_{\text{pretrained}}$ and a copy of $\delta _{\text{task}}$ for every task.
As long as the size of $\delta_{\text{task}}$ can be reduced, we can achieve parameter efficient models.
To this end, Diff Pruning \citep{guo-etal-2021-parameter} utilizes $L_0$-norm penalty \citep{louizos2018learning} to encourage sparsity of the difference vector $\delta_{\text{task}}$. \cite{aghajanyan2020intrinsic} adopt FastFood transform $M$ \citep{IntrinsicDimension} to convert $\delta_{\text{task}}$ into a low-dimensional vector $\delta_{\text{low}}$, i.e., $\delta_{\text{task}}=\delta_{\text{low}} M$.
The element-wise addition can also be replaced by element-wise multiplication. For instance,
Piggyback \citep{Piggyback} multiplies  real-valued mask weights to the pre-trained parameters, i.e., $\theta_{\text{task}} = \theta_{\text{pretrained}} \odot \delta_{\text{task}}$ during training. After training, the mask weights $\delta_{\text{task}}$ are passed through a thresholding function to obtain binary-valued masks, further reducing the parameter storage of $\delta_{\text{task}}$ at inference.

The essential difference between the above two tuning methods lies in their different assumptions about the root of transferability.
Residual tuning assumes that transferability is encoded in the \emph{behaviors} of each module, i.e., the features output by each module.
When adapting to the downstream tasks, we only need to add some task-specific behaviors by stacking the pre-trained modules with the residual adapter modules.
In contrast, parameter difference tuning assumes that transferability lies in the pre-trained \emph{parameters}. Most of the pre-trained parameters can be reused, and only a small part of them need to be adapted to the downstream tasks, thus we only need to store the increment.
Another thing to mention is that when limiting the size of the residual adapters or the complexity of the difference vector, these methods naturally overcome the catastrophic forgetting issue in Section \ref{sec:CatastrophicForgetting}.

\subsubsection{Data Efficiency}
\label{sec:data_efficiency}

Currently, when fine-tuning large pre-trained models,  hundreds or even thousands of labeled samples are still required to achieve strong performance on a specific downstream task,
which limits the application of the ``pre-training and fine-tuning'' paradigm to wider range of tasks where labeled data are expensive to collect.
In contrast, people can adapt to a new task with extremely few labeled samples, which is known as  \textit{few-shot learning}, or even with no labeled samples, which is known as \textit{zero-shot learning}.
Considering the lifecycle of deep learning, we can tackle this problem in three ways. The first is to improve the cross-task transferability of the pre-trained models, such as by increasing the model capacity or the pre-training dataset size, which is mentioned in Section \ref{sec:pretraining}.
The second is to transfer from another labeled source domain where labeled data is cheaper to collect, which will be discussed in Section \ref{sec:domain_adaptation}.
The last is to reformulate the target task to close its gap with the pre-trained models, which is the focus of this part.

\paragraph{Metric Learning.}
Fine-tuning in low data regimes will easily cause over-fitting as updating the model of large-scale parameters using few labeled samples is ill-posed. In contrast, many \emph{non-parametric} methods, such as nearest neighbors, can deal with low-sample regimes without suffering from catastrophic forgetting. To combine the advantages of parametric and non-parametric methods, Matching Net \citep{cite:NIPS16MatchingNet} uses an attention mechanism over the learned representations to predict the classes for the query samples, which can be interpreted as weighted nearest neighbors.
Since labeled data is severely limited, ProtoNet \citep{PrototypicalNetworks} adds a stronger inductive bias that there exists a single prototype representation for each class, where each prototype is the mean of the features of the labeled samples in each class, and classification boils down to finding the nearest prototype.
Since no gradient update is performed on the feature representation, choosing a proper distance metric that has good transferability across tasks plays an important role.
A common choice is the cosine distance, which explicitly reduces the intra-class variations and improves the cross-task transferability.
\cite{cite:ICLR19ACloserLook} find that by replacing the linear classifier with a cosine-distance based classifier, the naive feature transfer method without fine-tuning serves as a strong baseline in few-shot learning.

\paragraph{Prompt Learning.}
Prompting, firstly proposed in GPT-3 \citep{cite:NIPS20GPT3}, is another way to reformulate the downstream task to make it similar to the solved pre-training task.
In fine-tuning, models will take input $\mathbf{x}$ and predict an output $\mathbf{y}$ as $P(\mathbf{y}|\mathbf{x})$. In prompting, the original input $\mathbf{x}$ is modified by a prompt template into a new string $\tilde{\mathbf{x}}$ that has unfilled slots, then the pre-trained language model will fill $\tilde{\mathbf{x}}$ to obatin a final string $\widehat{\mathbf{x}}$ and derive the output $\mathbf{y}$ from $\widehat{\mathbf{x}}$ \citep{ppp_survey}. Table \ref{table:prompt_example} provides an example of prompting methods.

\begin{table}[htbp]
	\centering
	\scriptsize
	\caption{An example of prompting method from \cite{ppp_survey}.}
	\label{table:prompt_example}
	\begin{tabular}{@{}lll@{}}
		\toprule
		\textbf{Name}   & \textbf{Notation}               & \textbf{Example}                                  \\ \midrule
		Input           & $\mathbf{x}$                    & I love this film.                                 \\
		Output          & $\mathbf{y}$                    & \texttt{positive}                                 \\
		Prompt Template & ${f_\text{prompt}(\mathbf{x})}$ & {[}X{]} Overall, it was a {[}Z{]} film.           \\
		Prompt          & $\tilde{\mathbf{x}}$            & I love this film. Overall, it was a {[}Z{]} film. \\
		Filled Prompt   & $\mathbf{\widehat{x}}$          & I love this movie. Overall, it was a good movie.  \\ \bottomrule
	\end{tabular}
\end{table}

\begin{figure}[!bp]
	\centering
	\includegraphics[width=1\textwidth]{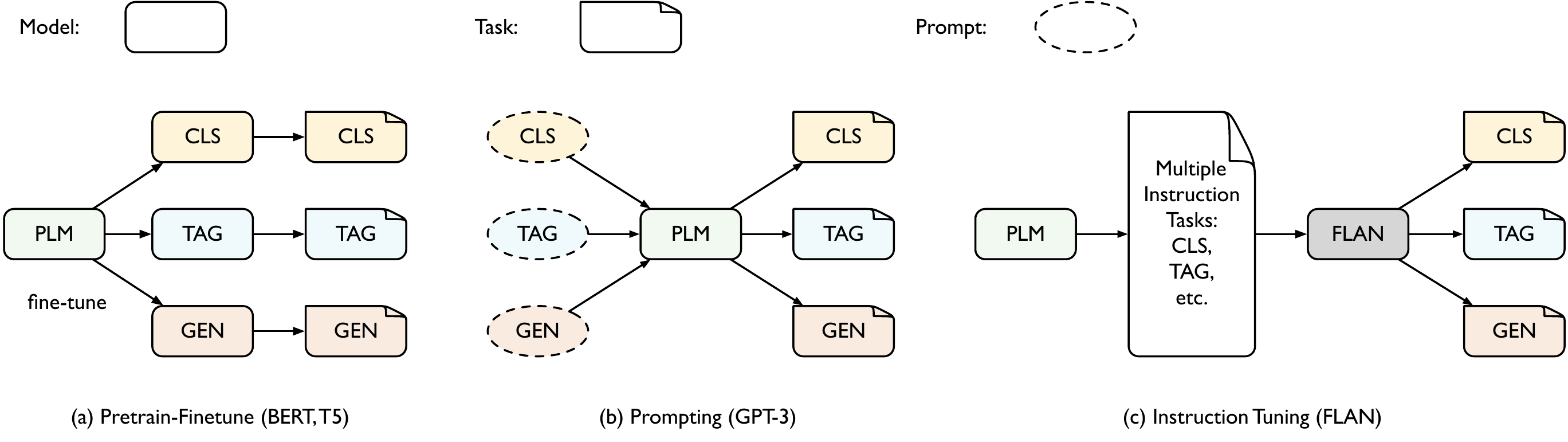}
	\caption{
	{(a)} Fine-tuning generates a new set of parameters for each downstream task. {(b)} Prompting fixes the pre-trained parameters and finds task-specific prompts to solve each downstream task.	{(c)} Instruction Tuning tunes the pre-trained models on instruction-format dataset and uses the obtained model to do inference with multiple downstream tasks.
	}
	\label{fig:prompt}
\end{figure}

The advantage of prompting is that it enables few-shot or even zero-shot task adaptation. The strong cross-task transferability stems from the implicit multiple tasks such as question-answering that language models are forced to learn on the large-scale pre-training corpus. However, this transferability requires a large model capacity to deal with potential implicit tasks and is also very sensitive to the choice of prompts.  Thus, the disadvantage is that it introduces the necessity for prompt engineering, i.e., finding the best prompts to solve each downstream task, which is work-heavy and time-consuming especially on large datasets.

Combining prompting and fine-tuning may tackle this problem (Figure~\ref{fig:prompt}).  PET-TC \citep{schick2020exploiting} tunes the parameters of pre-trained language models in prompt learning while Prefix-Tuning \citep{li2021prefixtuning} adds additional prompt-related parameters and tunes these parameters.
Instruction Tuning \citep{wei2021finetuned} explicitly fine-tunes the pre-trained models on a mixture of datasets expressed through natural language instructions (similar to the filled prompt) and obtain Fine-tuned LAnguage Model (FLAN), which largely increases the models' transferability to unseen tasks.
In summary, prompt learning has provided a revolutionary way on how to utilize the transferability of pre-trained models.

\subsubsection{Remarks}

Table \ref{table:TA_comparison} gives a comparison of different task adaptation methods. Fine-tuning (including vanilla fine-tuning, domain adaptive tuning, and regularization tuning) has better performance when there are enough labeled data in the downstream tasks.
In contrast, prompt learning requires much fewer labeled data to achieve decent performance, yet its applications are still limited to NLP and it is still non-trivial to extend it to vision or other areas.
Fine-tuning is the most parameter-inefficient since it generates a full set of model parameters for each downstream task, while residual tuning, difference tuning, and prompt learning are all parameter efficient.
Also, these latter methods naturally mitigate the catastrophic forgetting problem, but \emph{negative transfer} is still a hard problem to be resolved.

\begin{table}[htbp]
	\addtolength{\tabcolsep}{2pt}
	\centering
	\scriptsize
	\caption{Comparison between different task adaptation methods. }
	\label{table:TA_comparison}
	\begin{threeparttable}
		\begin{tabular}{lccccc}
			\toprule
			& \begin{tabular}[c]{@{}c@{}}Adaptation\\  Performance\tnote{1}\end{tabular} &
			\begin{tabular}[c]{@{}c@{}}Data\\  Efficiency\tnote{2}\end{tabular} &
			\begin{tabular}[c]{@{}c@{}}Parameter \\  Efficiency\tnote{3}\end{tabular} &
			\begin{tabular}[c]{@{}c@{}}Modality \\ Scalability\tnote{4}\end{tabular} &
			\begin{tabular}[c]{@{}c@{}}Task \\ Scalability\tnote{5}\end{tabular}  \\
			\midrule
			Feature Transfer            & $\bigstar$                 & $\bigstar\bigstar$         & $\bigstar\bigstar\bigstar$ & $\bigstar\bigstar\bigstar$ & $\bigstar\bigstar\bigstar$ \\
			\midrule
			Vanilla Fine-tuning         & $\bigstar\bigstar\bigstar$ & $\bigstar$                 & $\bigstar$                 & $\bigstar\bigstar\bigstar$ & $\bigstar\bigstar\bigstar$ \\
			\midrule
			Domain Adaptive Tuning      & $\bigstar\bigstar\bigstar$ & $\bigstar\bigstar$         & $\bigstar$                 & $\bigstar\bigstar$         & $\bigstar\bigstar\bigstar$ \\
			\midrule
			Regularization Tuning       & $\bigstar\bigstar\bigstar$ & $\bigstar\bigstar$         & $\bigstar$                 & $\bigstar\bigstar\bigstar$ & $\bigstar$                 \\
			\midrule
			Residual Tuning             & $\bigstar\bigstar$         & $\bigstar\bigstar$         & $\bigstar\bigstar$         & $\bigstar\bigstar$         & $\bigstar\bigstar$         \\
			\midrule
			Parameter Difference Tuning & $\bigstar\bigstar$         & $\bigstar\bigstar$         & $\bigstar\bigstar$         & $\bigstar\bigstar\bigstar$ & $\bigstar\bigstar\bigstar$ \\
			\midrule
			Metric Learning
			                            & $\bigstar$                 & $\bigstar\bigstar\bigstar$ & $\bigstar\bigstar\bigstar$ & $\bigstar\bigstar\bigstar$ & $\bigstar$                 \\
			\midrule
			Prompt Learning
			                            & $\bigstar\bigstar$         & $\bigstar\bigstar\bigstar$ & $\bigstar\bigstar\bigstar$ & $\bigstar$                 & $\bigstar$                 \\
			\bottomrule
		\end{tabular}
		\begin{tablenotes}
			\scriptsize
			\item[1] Adaptation Performance: performance when there are large-scale labeled data in downstream tasks.
			\item[2] Data Efficiency: performance when there are only small-scale labeled data in downstream tasks.
			\item[3] Parameter Efficiency: whether can control parameters when the number of downstream tasks increases.
			\item[4] Modality Scalability: whether can adapt pre-trained models to various modalities, such as text, graph.
			\item[5] Task Scalability: whether can adapt pre-trained models to different downstream tasks, such as detection.
		\end{tablenotes}
	\end{threeparttable}
\end{table}

The motivation of many task adaptation methods can be understood from the perspective of \emph{transferability}.
For instance, domain adaptive tuning aims to bridge the domain discrepancy between the pre-training task and the downstream task by further obtaining a pre-trained model on the target data distribution.
Prompt learning aims to bridge the task discrepancy between the pre-training task and the downstream task by reformulating all the tasks to the same format.
In this scenario, when all tasks can be expressed in the same form, the difference between the pre-training task and the downstream task is only the shift in data distributions, i.e., task adaptation becomes the domain adaptation problem.

\subsection{Domain Adaptation}
\label{sec:domain_adaptation}
\label{sec:DA_theory}

The pre-training and fine-tuning paradigm has greatly improved the state-of-the-arts for diverse machine learning problems and applications, and the pre-trained deep networks can be easily adapted to the tasks at hand
even with a small amount of labeled data.
However, in many practical scenarios, there is no labeled training data and thus there is the demand to transfer a deep network from a source domain where labeled training data is available to a target domain where only unlabeled data exists \citep{MarginalizedDenoisingAutoencoders, DAForSentimentClassification}.
In this situation, the deep models still suffer from performance
degradations due to \emph{distribution shift} \citep{cite:Book09DSS}. Thus, domain adaptation is proposed to reduce the distribution shift between training and testing domains \citep{transfer_survey}.

Many methods have been proposed for domain adaptation in the shallow regime, either by re-weighting or selecting samples from the source domain \citep{NIPS2007_be83ab3e} or seeking an explicit feature space transformation from the source distribution into the target distribution \citep{cite:DAtransform1}.
As seminal methods, \cite{NIPS2006_a2186aa7,TCA, JDA} explicitly match the distributions in the kernel-reproducing Hilbert space, while \cite{GFK} map the principal axes associated with each of the distributions.
This survey will focus on deep domain adaptation, where adaptation modules are embedded in deep architectures to match data distributions across domains.

In unsupervised domain adaptation (UDA), there is a source domain $\widehat{\mathcal{S}}=\{(\mathbf{x}_i^s, \mathbf{y}_i^s)\}_{i=1}^{n}$ of $n$ labeled samples and a target domain $\widehat{\mathcal{T}}=\{\mathbf{x}_i^t \}_{i=1}^{m}$ of $m$ unlabeled samples.
The goal of a learning algorithm is to find a hypothesis $h:\mathcal{X} \mapsto \mathcal{Y}$ in the hypothesis space $\mathcal{H}$ with a low target risk $\epsilon_{\mathcal{T}}(h)=\mathbb{E}_{(\mathbf{x}^t, \mathbf{y}^t)\sim \mathcal{T}} [\ell(h(\mathbf{x}^t), \mathbf{y}^t)]$ with no access to the labels of $\mathcal{T}$, where $\ell: \mathcal{Y} \times \mathcal{Y} \rightarrow \mathbb{R}_+$ is a loss function.
Several seminal theories have been proposed to tackle this problem and the
the main idea of them is to bound the target risk $\epsilon_{\mathcal{T}}(h)$  by the source risk
$\epsilon_{\mathcal{S}}(h)$
and a distribution distance.
In this survey, we will focus on the theory of $\mathcal H\Delta \mathcal H$-Divergence \citep{DATheroy07, DATheroy10, GeneralizedDATheory} and Disparity Discrepancy \citep{MDD} and illustrate how to derive different algorithms from these theories.
First, using triangle inequalities, we can relate the target risk to the source risk as follows.

\begin{theorem}[Bound with Disparity]
	Assume that the loss function $\ell$ is symmetric and obeys the triangle inequality.
	Define the \emph{disparity} between any two hypotheses $h$ and $h'$ on distribution $\mathcal{D}$ as
	\begin{equation}
		\epsilon_\mathcal{D}(h, h') = \mathbb{E}_{(\mathbf{x}, \mathbf{y})\sim \mathcal{D}} [\ell(h(\mathbf{x}), h'(\mathbf{x}))].
	\end{equation}
	Then the target risk $\epsilon_{\mathcal{T}}(h)$ can be bounded by
	\begin{equation}
		\begin{aligned}
			{\epsilon _{\mathcal{T}}}\left( h \right) & \leqslant {\epsilon _{\mathcal{S}}}\left( h \right) +
			\left[ {{\epsilon _{\mathcal{S}}}\left( {{h^ * }} \right) + {\epsilon _{\mathcal{T}}}\left( {{h^ * }} \right)} \right]
			+ \left| {{\epsilon _{\mathcal{S}}}\left( {h,{h^ * }} \right) - {\epsilon _{\mathcal{T}}}\left( {h,{h^ * }} \right)} \right|,\\
		\end{aligned}
	\end{equation}
	where $h^{*} = {\arg\min}_{h\in\mathcal{H}} \left[ {{\epsilon _{\mathcal{S}}}\left( {{h }} \right) + {\epsilon _{\mathcal{T}}}\left( {{h}} \right)} \right]$ is the \textit{ideal joint hypothesis}, $\epsilon_{ideal} ={{\epsilon _{\mathcal{S}}}\left( {{h^ * }} \right) + {\epsilon _{\mathcal{T}}}\left( {{h^ * }} \right)}$ is
	the \textit{ideal joint error}, $\left| {{\epsilon _{\mathcal{S}}}\left( {h,{h^ * }} \right) - {\epsilon _{\mathcal{T}}}\left( {h,{h^ * }} \right)} \right|$ is the \emph{disparity difference} between $\mathcal{S}$ and $\mathcal{T}$.
\end{theorem}

It is a common \emph{assumption} in domain adaptation that the ideal joint error ${\epsilon_{ideal}}$ shall be sufficiently small, otherwise domain adaptation will be infeasible, the \emph{impossibility} theorem \citep{pmlr-v9-david10a}. The goal is reduced to bound the disparity difference. However, since the ideal hypothesis $h^*$ is unknown due to the unavailability of labeled target data, the disparity difference cannot be estimated directly. To this end, $\mathcal H\Delta \mathcal H$-Divergence \citep{DATheroy07, DATheroy10} is proposed to measure the upper bound of the disparity difference.

\begin{definition}[$\mathcal H\Delta \mathcal H$-Divergence]
	Define $\mathcal H\Delta \mathcal H\triangleq  \{h|h=h_1 \otimes h_2, h_1, h_2 \in \mathcal{H} \}$ as the \emph{symmetric difference hypothesis space} of $\mathcal{H}$, where $\otimes$ stands for the XOR operator. Then the \emph{$\mathcal H\Delta \mathcal H$-Divergence} between $\mathcal{S}$ and $\mathcal{T}$ is
	\begin{equation}
		\label{Equ:H-divergence}
		\begin{aligned}
			d_{\mathcal H\Delta \mathcal H}(\mathcal{S},\mathcal{T}) & \triangleq \sup_{h,h'\in\mathcal H}\left| {{\epsilon _{\mathcal{S}}}\left( {h,{h^ \prime }} \right) - {\epsilon _{\mathcal{T}}}\left( {h,{h^ \prime }} \right)} \right|
			. \notag
		\end{aligned}
	\end{equation}
	For binary classification problem with the $01$-loss, $\ell(y, y')=\mathds{1}(y\neq y')$, we have
	\begin{equation}
		\begin{aligned}
			d_{\mathcal H\Delta \mathcal H}(\mathcal{S},\mathcal{T}) &
			= {\sup _{\delta  \in \mathcal H\Delta\mathcal H }}\left| {{\mathbb{E}_{{\mathcal{S}}}}\left[ {\delta  {\left({\mathbf{x}}\right) }  \ne 0} \right] - {\mathbb{E}_{{\mathcal{T}}}}\left[ {\delta \left( {{\mathbf{x}}} \right) \ne 0} \right]} \right|. \notag
		\end{aligned}
	\end{equation}
\end{definition}

The main advantage of the $\mathcal H\Delta \mathcal H$-Divergence is that it can be estimated from \emph{finite} unlabeled samples of source and target domains.
However, it is generally hard to compute and optimize. Thus, it is approximated by training a domain discriminator $D$ that separates the source and target samples \citep{DATheroy07, DANN}.
Assume that the family of the discriminators is rich enough, such as the multilayer perceptrons (MLP) that is universal approximator to any functions, to contain $\mathcal H\Delta \mathcal H$, i.e., $\mathcal H\Delta \mathcal H \subset \mathcal{H}_D$. The $\mathcal H\Delta \mathcal H$-Divergence can be further bounded by ${\sup_{D \in {\mathcal{H}_D}}}\left| {{\mathbb{E}_{{\mathcal{S}}}}\left[ {D  {\left({\mathbf{x}}\right) }  = 1} \right] + {\mathbb{E}_{{\mathcal{T}}}}\left[ {D \left( {{\mathbf{x}}} \right) = 0} \right]} \right|$,
which gives rise to the \emph{domain adversarial} methods in Section \ref{sec:DAAdversarialTraining}.
The $\mathcal H\Delta \mathcal H$-Divergence can also be estimated in a nonparametric way by replacing $\mathcal H\Delta \mathcal H$ with a proper function space $\mathcal{F}$, which induces the \emph{statistics matching} methods in Section \ref{sec:DAMomentMatching}.

The following theorem is the earliest theory in domain adaptation, which establishes the generalization bound based on the $\mathcal H\Delta \mathcal H$-Divergence for binary classification problems.

\begin{theorem}[\cite{DATheroy10}]
	Let $\mathcal{H}$ be a binary hypothesis space of VC dimension $d$.
	If  $\widehat{\mathcal{S}}$ and $\widehat{\mathcal{T}}$ are samples of size $m$ each, then for any $\delta \in (0,1)$, with probability at least $1-\delta$, for every $h\in \mathcal{H}$,
	\begin{equation}
		\epsilon_{\mathcal{T}} (h) \le \epsilon_{{\mathcal{S}}} (h) +  d_{\mathcal H\Delta \mathcal H}({\mathcal{\widehat{\mathcal{S}}}}, {\mathcal{\widehat T}}) +
		\epsilon_{ideal} +4\sqrt{\frac{2d \log (2m)+\log(\frac{2}{\delta})}{m}}.
	\end{equation}
\end{theorem}

This bound sheds key insights into algorithm designs. However, it has the limit of being based on the particular $01$-loss. Thus, \cite{GeneralizedDATheory} extend the domain adaptation theory to a general class of loss functions satisfying the symmetry and subadditivity.

\begin{theorem}[\cite{GeneralizedDATheory}]
	\label{theorem:GeneralizedDATheory}
	Assume that the loss function $\ell$ is symmetric and obeys the triangle inequality, and define $h_{\mathcal{S}}^*=\arg\min_{h\in \mathcal{H}}\epsilon _{\mathcal{S}}(h)$ and $h_{\mathcal{T}}^*=\arg\min_{h\in \mathcal{H}}\epsilon _{\mathcal{T}}(h)$ as the ideal hypotheses for the source and target domains, then for every $h \in \mathcal{H}$,
	\begin{equation}
		\epsilon_{\mathcal{T}} (h) \le \epsilon_{{\mathcal{S}}} (h, h_{\mathcal{S}}^*) +  d_{\mathcal H\Delta \mathcal H}({\mathcal{S}}, {\mathcal{T}}) +
		\epsilon,
	\end{equation}
	where $\epsilon_{{\mathcal{S}}} (h, h_{\mathcal{S}}^*)$ stands for the source risk and  $\epsilon=\epsilon_{{\mathcal{T}}} (h_{\mathcal{T}}^*) + \epsilon_{{\mathcal{S}}} (h_{\mathcal{T}}^*, h_{\mathcal{S}}^*)$ for the capacity to adapt.
	Further, let $\ell$ be bounded, $\forall (y, y')\in \mathcal{Y}^2, \ell(y, y')\leq M$ for some $M>0$, and defined as $ \ell(y, y')=|y-y'|^q$ for some $q$.
	If  $\widehat{\mathcal{S}}$ and $\widehat{\mathcal{T}}$ are samples of size $n$  and $m$ each,
	with probability at least $1-\delta$, we have
	\begin{equation}
		d_{\mathcal H\Delta \mathcal H}({\mathcal{S}}, {\mathcal{T}}) \le  d_{\mathcal H\Delta \mathcal H}(\widehat{\mathcal{S}}, \widehat{\mathcal{T}}) +  4q(\mathfrak R_{n, \mathcal{S}}(\mathcal{H}) + \mathfrak R_{m, \mathcal{T}}(\mathcal{H})) + 3M\Bigg( \sqrt{\frac{\log\frac{4}{\delta}}{2n}}+ \sqrt{\frac{\log\frac{4}{\delta}}{2m}}\Bigg),
	\end{equation}
	where $\mathfrak R_{n, \mathcal{D}}$ is the expected Rademacher Complexity \citep{RademacherComplexity} with respect to distribution $\mathcal{D}$ and sample size $n$.
\end{theorem}

Note that the $\mathcal H\Delta \mathcal H$-Divergence bounds are still \emph{loose} since the supremum is taken over both $h' \in \mathcal{H}$ and $h \in \mathcal{H}$. Observing that $h$ is known as the source classifier, the Disparity Discrepancy \citep{MDD} provides a tighter bound by computing directly on $\mathcal{H}$.

\begin{definition}[Disparity Discrepancy]
	Given a binary hypothesis space $\mathcal{H}$ and a \emph{specific hypothesis} \(h\!\in \!\mathcal H\), the \emph{Disparity Discrepancy} induced by $h' \in \mathcal{H}$ is defined by
	\begin{equation}
		d_{h,\mathcal H}(\mathcal{S},\mathcal{T}) = \sup_{h'\in \mathcal H}
		\left(\mathbb{E}_{ \mathcal{T}}\mathds{1} [h'\ne h] - \mathbb{E}_{ \mathcal{S}} \mathds{1}[h'\ne h]\right)
	\end{equation}
\end{definition}

Since the supremum is only take over $h' \in \mathcal{H}$, estimating and minimizing the disparity discrepancy jointly through a minimax game can be done much more easily. The disparity discrepancy can well measure the distribution shift and yields a tighter generalization bound.

\begin{theorem}[\cite{MDD}]
	\label{theorem:DD}
	Let  $\widehat{\mathcal{S}}$ and $\widehat{\mathcal{T}}$ be samples of size $n$  and $m$ each.
	For any $\delta>0$ and every binary classifier $h\in \mathcal{H}$, with probability at least $1-3\delta$, we have
	\begin{equation}
		\begin{aligned}
			\epsilon_\mathcal{T}(h) & \le \epsilon_{\widehat{\mathcal{S}}}(h) + d_{h,\mathcal{H}}(\widehat{\mathcal{S}},\widehat{\mathcal{T}}) +\epsilon_{ideal} +2\mathfrak R_{n, \mathcal{S}}(\mathcal{H})
			\\
			                        & +2\mathfrak R_{n, \mathcal{S}}(\mathcal{H} \Delta \mathcal{H} ) + 2\sqrt{\frac{\log\frac{2}{\delta}}{2n}} + 2\mathfrak R_{m, \mathcal{T}}(\mathcal{H} \Delta \mathcal{H} ) + \sqrt{\frac{\log\frac{2}{\delta}}{2m}}.
		\end{aligned}
	\end{equation}
\end{theorem}

The disparity discrepancy can be further extended to the \emph{multiclass} classification problem with hypothesis space $\mathcal{F}$ of scoring
functions $f: \mathcal{X} \times \mathcal{Y} \rightarrow \mathbb{R}$ and margin loss, which is going beyond existing bounds and closer to the choices for real tasks \citep{MDD}.

\begin{definition}[Margin Disparity Discrepancy]\label{def:MDD}
	Given a scoring hypothesis space $\mathcal{F}$, denote the margin of a real hypothesis $f$ at a labeled example
	$(x,y)$ as $\rho_f (x,y) \triangleq \frac{1}{2} (f(x,y) - \max_{y'\neq y} f(x,y'))$, the labeling function induced by $f$ as $h_f: x \mapsto \arg\max_{y\in\mathcal{Y}}f(x, y)$, and the margin loss as
	\begin{equation}
		\Phi_{\rho}(x)\triangleq
		\begin{cases}
			0           & \rho \leq {x}       \\
			1 - x/ \rho & 0 \leq {x} \le \rho \\
			1           & {x} \le 0           \\
		\end{cases},
	\end{equation}
	then the \emph{margin disparity} between $f$ and $f'$ on distribution $\mathcal{D}$ is
	\begin{equation}
		\epsilon_{\mathcal{D}}^{(\rho)}(f', f)=\mathbb{E}_{(x,y)\sim\mathcal{D}}[\Phi_{\rho}(\rho_{f'}(x,h_f(x))].
	\end{equation}
	Given a \emph{specific hypothesis} \(f\!\in \!\mathcal F\), the \emph{Margin Disparity Discrepancy} is defined by
	\begin{equation}
		d_{f,\mathcal F}^{(\rho)}(\mathcal{S},\mathcal{T}) = \sup_{f'\in \mathcal F}[
			\epsilon_{\mathcal{T}}^{(\rho)}(f', f) - \epsilon_{\mathcal{S}}^{(\rho)}(f', f)].
	\end{equation}
\end{definition}

Note that the margin disparity satisfies the nonnegativity and
subadditivity, but not the symmetry. Thus Theorem \ref{theorem:GeneralizedDATheory} cannot apply here and a new generalization bound is derived.

\begin{theorem}[\cite{MDD}]\label{theorem:MDD}
	Given the same settings with Definition \ref{def:MDD}, for any $\delta > 0$, with probability at least $1-3\delta$, the following \emph{margin bound} holds for all scoring functions $f\in\mathcal F$,
	\begin{equation}
		\begin{aligned}
			\epsilon_\mathcal{T}(f)\leq & \ \epsilon_{\widehat{\mathcal{S}}}^{(\rho)}(f)+d_{f,\mathcal F}^{(\rho)}(\mathcal{\widehat{\mathcal{S}}},\mathcal{\widehat T}) +\epsilon_{ideal}
			+\frac{2k^2}{\rho}{\mathfrak{R}_{n, \mathcal{S}}}(\Pi_1\mathcal F) \\
			+                           & \ \frac{k}{\rho}{\mathfrak{R}_{n, \mathcal{S}}}(\Pi_{\mathcal H}\mathcal F)+2\sqrt{\frac{\log \frac 2 {\delta}}{2n}}
			+\frac{k}{\rho}{\mathfrak{R}_{m, \mathcal{T}}}(\Pi_{\mathcal H}\mathcal F)+\sqrt{\frac{\log \frac{2}{\delta}}{2m}},
		\end{aligned}
	\end{equation}
	where $\Pi_{\mathcal{H}}\mathcal{F}\triangleq \{x \mapsto f(x, h(x)) | h\in \mathcal{H}, f \in \mathcal{F} \}$ is the scoring version of the symmetric hypothesis space $\mathcal H\Delta \mathcal H$, $\Pi_{1}\mathcal{F}\triangleq \{x \mapsto f(x,y) | y\in \mathcal{Y}, f \in \mathcal{F} \}$ and $\epsilon_{ideal}=\min_{f^* \in \mathcal{F}}\{\err_{\mathcal{S}}^{(\rho)}(f^*) + \err_{\mathcal{T}}^{(\rho)}(f^*)\}$ is the ideal joint error in terms of margin loss, $k$ is the number of classes.
\end{theorem}

This margin bound suggests that a proper margin $\rho$ could yield better generalization on the target domain. Theorems \ref{theorem:DD}  and \ref{theorem:MDD} together form the theoretical basis of the \emph{hypothesis adversarial} methods in Section \ref{sec:DADisparityDiscrepancy}.
Note that the supremum in both the $\mathcal H\Delta \mathcal H$-Divergence and Disparity Discrepancy will become meaningless, when the allowed hypothesis space $\mathcal H$ is too large, which is common in deep neural networks. Thus, pre-training the deep neural networks on large-scale upstream data to decrease the allowed hypotheses is still necessary for the domain adversarial methods and hypothesis adversarial methods.

\begin{figure*}[!bp]
	\centering
	\includegraphics[width=1.\textwidth]{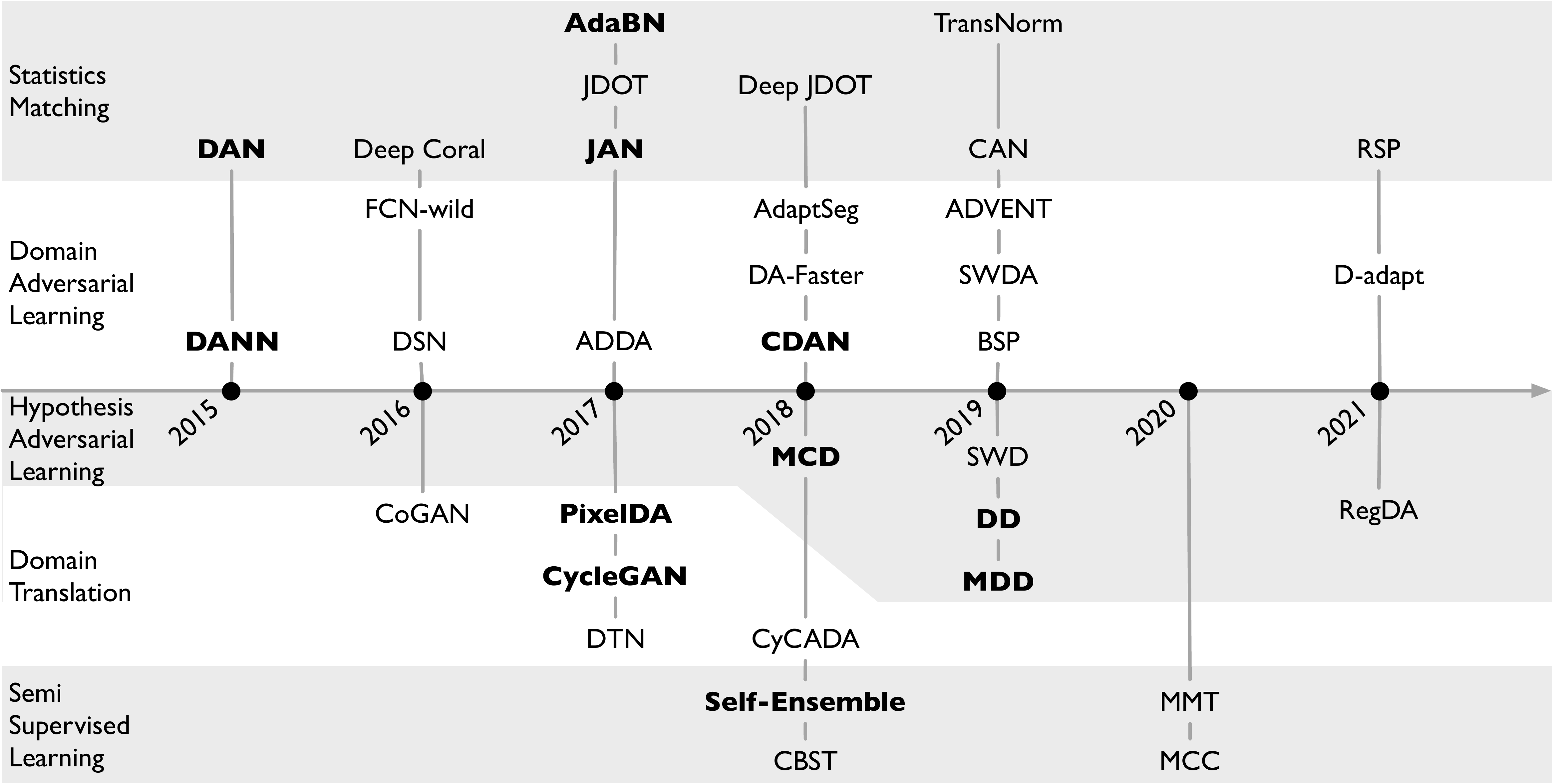}
	\caption{The cornerstones of domain adaptation methods in deep learning.}
	\label{fig:DA_family}
\end{figure*}

A final important note is that while there are no theoretical guarantees for some well-established methods, they have also achieved quite strong performance in practice, such as the \emph{domain translation} methods in Section \ref{sec:DATranslation} and the \emph{semi-supervised learning} methods in Section \ref{sec:semi_supervised}.
Figure \ref{fig:DA_family} highlights the cornerstones of domain adaptation methods in deep learning, which rely on the reuse of transferability gained in pre-trained deep models.

\subsubsection{Statistics Matching}
\label{sec:DAMomentMatching}

We have introduced several seminal theories on the generalization bounds for domain adaptation, which are all based on the \emph{hypothesis-induced} distribution distances. These distances are less intuitive because they rely on unknown hypotheses and cannot be computed before learning the hypotheses. In this section, we first introduce another family of metrics on the space of \emph{probability measures} well-studied in probability theory, which provide interpretable and complementary properties to the hypothesis-induced distribution distances and relate closely to a large set of domain adaptation algorithms \citep{DAN,JAN}.

\begin{definition}[Maximum Mean Discrepancy]\label{def:MMD}
	Given two probability distributions $\mathcal{S}$ and $\mathcal{T}$ on a measurable space $\mathbf{X}$, the \emph{integral probability metric} \citep{redko2020survey} is defined as
	$d_{\mathcal{F}}(\mathcal{S}, \mathcal{T}) \triangleq \sup_{f\in \mathcal F}
		\big|\mathbb{E}_{\mathbf{x}\sim\mathcal{S}} [f(\mathbf{x})] - \mathbb{E}_{\mathbf{x}\sim\mathcal{T}} [f(\mathbf{x})]\big| $,
	where $\mathcal{F}$ is a class of bounded functions on $\mathbf{X}$. \cite{HilbertSpaceOnProbability} further restrict $\mathcal{F}$ as the unit ball in Reproducing Kernel Hilbert Space (RKHS) $\mathcal{H}_k$ endowed with a \emph{characteristic} kernel $k$, $\mathcal{F} = \{f\in\mathcal{H}_k: ||f||_{\mathcal{H}_k} \leq 1 \}$, leading to the \emph{maximum mean discrepancy (MMD)} \citep{AKernelTwo-SampleTest},
	\begin{equation}
		d_{\emph{MMD}}^2(\mathcal{S}, \mathcal{T}) = \big\| \mathbb{E}_{\mathbf{x}\sim\mathcal{S}} [\phi(\mathbf{x})] - \mathbb{E}_{\mathbf{x}\sim\mathcal{T}} [\phi(\mathbf{x})] \big\|_{\mathcal{H}_k}^2,
	\end{equation}
	where $\phi(x)$ is a feature map associated with kernel $k$ such that $k(\mathbf{x}, \mathbf{x}')= \left \langle \phi(\mathbf{x}), \phi(\mathbf{x}') \right \rangle$. It can be proved from probability theory that $\mathcal{S} = \mathcal{T}$ if and only if $d_{\mathcal{F}}(\mathcal{S}, \mathcal{T}) = 0$ or $d_{\emph{MMD}}^2(\mathcal{S}, \mathcal{T}) = 0$.
\end{definition}

\begin{theorem}[\cite{redko2020survey}]\label{theorem:MMD}
	Given the same settings with Definition \ref{def:MMD}, let $\ell$ be a convex loss function with a parametric form $ \ell(y, y')=|y-y'|^q$ for some $q$. Then for any $\delta > 0$, with probability at least $1-\delta$, the following bound holds for all $h\in\mathcal F$,
	\begin{equation}
		\begin{aligned}
			\epsilon_\mathcal{T}(h) & \leq \epsilon_{\mathcal{S}}(h)+d_{\emph{MMD}}(\mathcal{\widehat{\mathcal{S}}},\mathcal{\widehat T}) +\epsilon_{ideal}                                                                                                                                                                  \\
			                        & + \frac{2}{n} \mathbb{E}_{\mathbf{x}\sim\mathcal{S}}[\sqrt{\mathrm{tr}(\mathbf{K}_\mathcal{S})}] + \frac{2}{m}\mathbb{E}_{\mathbf{x}\sim\mathcal{T}}[\sqrt{\mathrm{tr}(\mathbf{K}_\mathcal{T})}]  + \sqrt{\frac{\log \frac{2}{\delta}}{2n}} + \sqrt{\frac{\log \frac{2}{\delta}}{2m}},
		\end{aligned}
	\end{equation}
	where $\mathbf{K}_\mathcal{S}$ and $\mathbf{K}_\mathcal{T}$ are the kernel matrices computed on samples from $\mathcal{S}$ and $\mathcal{T}$, respectively.
\end{theorem}

This bound has several advantages compared to previous theories. First, it is \emph{hypothesis-free} and does not require estimating hypotheses to measure the distribution distance. Second, the complexity term does not depend on the Vapnik-Chervonenkis dimension. Third, the unbiased estimate of MMD can be computed in linear time. Fourth, minimizing MMD has a nice interpretation of \emph{statistics matching} in the probability space. These advantages make the bound particularly useful to underpin several seminal algorithms.

\begin{figure}[!t]
	\centering
	\includegraphics[width=0.9\textwidth]{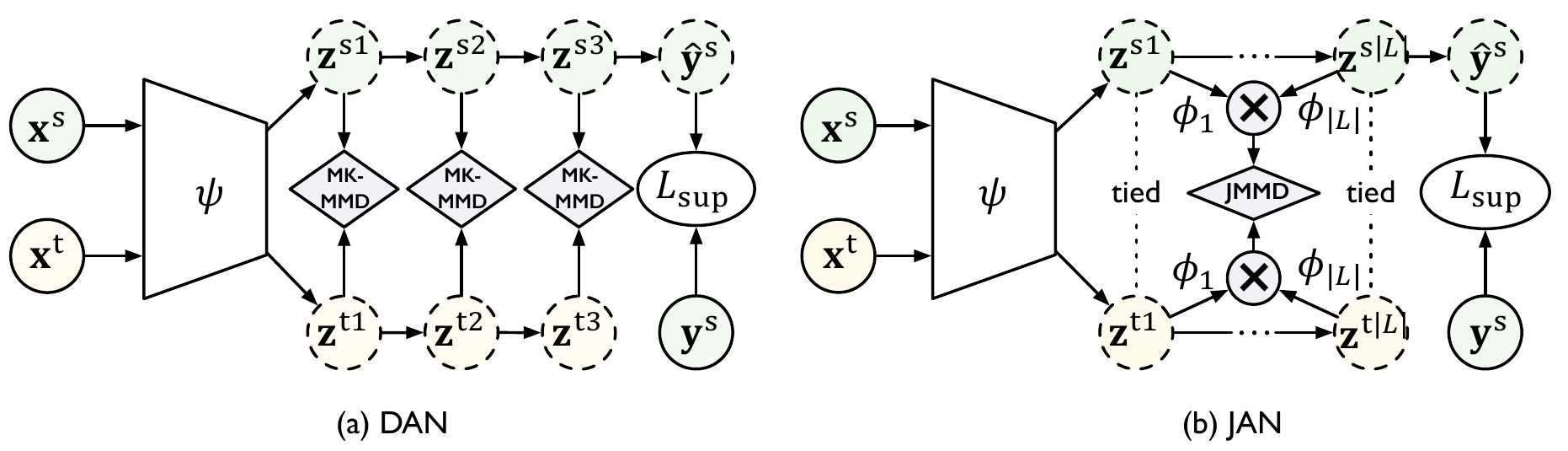}
	\caption{
		The cornerstone methods of statistics matching: (a) DAN adapts the \emph{marginal} distributions of activations in multiple task-specific layers with MK-MMD.
		(b) JAN adapts the \emph{joint} distributions of the feature activations and classification predictions with JMMD.
	}
	\label{fig:DAN}
\end{figure}

Deep Domain Confusion (DDC) \citep{DDC} applies MMD with a linear kernel to a single feature layer of the deep network, yet it has limited power for closing the domain gap since linear kernel is not characteristic and cannot ensure MMD to be a probability metric. Thereby, Deep Adaptation Network (DAN) \citep{DAN, DAN19} introduces the multiple-kernel variant of MMD (MK-MMD) \citep{MMD, AKernelTwo-SampleTest} to measure the domain relatedness, employing a convex combination of multiple characteristic kernels such as Gaussian kernel to make the function space $\mathcal{F}$ rich enough and enhance the distinguishing power of MK-MMD. Besides, as shown in Figure \ref{fig:DAN}, multiple domain-specific layers are adapted by MK-MMD, which enables learning transferable features for domain adaptation.

DAN mainly reduces the shift in the feature distribution and ignores that in the label distribution. Take AlexNet as example, the feature distribution shift mainly exists in layers $fc6$ and $fc7$ while the label distribution shift mainly exists in layer $fc8$. Joint Adaptation Network (JAN) \citep{JAN} proposes Joint Maximum Mean Discrepancy (JMMD) to measure the shift in joint distributions $P(\mathbf{X}^s, \mathbf{Y}^s)$ and $Q(\mathbf{X}^t, \mathbf{Y}^t)$. Denoting the activations of adapted layers $\mathcal{L}$ as $\{ (\mathbf{z}_i^{s1}, \dots , \mathbf{z}_i^{s|\mathcal{L}|}) \}_{i=1}^{n}$ and $\{ (\mathbf{z}_j^{t1}, \dots , \mathbf{z}_j^{t|\mathcal{L}|}) \}_{j=1}^{m}$, JMMD is defined as
\begin{equation}
	d_{\text{JMMD}}^2(\widehat{\mathcal{S}}, \widehat{\mathcal{T}}) = \Big\| \mathbb{E}_{i\in[n]} \otimes_{l \in \mathcal{L}} \phi^l(\mathbf{z}_i^{sl}) - \mathbb{E}_{j\in[m]} \otimes_{l \in \mathcal{L}} \phi^l(\mathbf{z}_j^{tl}) \Big\|_{\mathcal{H}_k}^2
\end{equation}
where $\phi^l$ is the feature map associated with kernel $k^l$ for layer $l$ and $\otimes$ is the outer product.

A characteristic kernel widely used in MMD is the Gaussian kernel, or $k(\mathbf{x}_1, \mathbf{x}_2)=\exp(-|| \mathbf{x}_1 - \mathbf{x}_1 ||^2 / {2\sigma^2})$.
After Taylor expansion, MMD can be considered as a weighted sum of distances between all orders of statistic moments. Thus, the statistic moments can be directly used to measure the distribution distance. For instance, deep CORAL \citep{coral} uses the second-order statistics (covariance) to measure distribution distance, which is frustratingly easy yet useful. Center moment discrepancy (CMD) \citep{CMD} further considers an explicit order-wise matching of higher-order moments.

One disadvantage of MMD is that it cannot take into account the geometry of the data distribution when estimating the discrepancy between two domains. Thus, Joint Distribution Optimal Transport (JDOT) \citep{jdot} is introduced into domain adaptation, and Deep JDOT \citep{deepjdot} further extends it to deep networks.
Another disadvantage is that minimizing MMD on the instance representation has the risk of changing the feature scale, while regression tasks are fragile to feature scaling. Thus, Representation Subspace Distance (RSD) \citep{DAR_ICML_21} closes the domain shift through orthogonal bases of the representation spaces, which are free from feature scaling.

Instead of explicitly matching the statistics moments of feature distributions, Adaptive Batch Normalization (AdaBN) \citep{AdaBN} implicitly minimizes domain discrepancy by aligning BatchNorm~\cite{BatchNorm} statistics. The hypothesis is that task-related knowledge is stored in the weight matrix while domain-related knowledge is represented by BatchNorm statistics. Thus AdaBN replaces the mean and variance of all BatchNorm layers with those estimated on the target domain at inference to reduce the domain shift. However, it is risky that AdaBN excludes the statistics on the target domain from training. Thus, Transferable Normalization (TransNorm) \citep{TransNorm} applies domain-specific mean and variance at both training and inference to capture sufficient statistics of both domains.

Finally, both MMD and JMMD may misalign samples from different classes due to a suboptimal modeling of the discriminative structure. To alleviate this problem, Contrastive Adaptation Network (CAN) \citep{CAN} alternatively estimates the labels of target samples through
clustering, and adapts the feature representations in a class-wise manner. Besides, CAN uses class-aware sampling for both domains to improve adaptation efficiency.

\begin{figure}[!t]
	\centering
	\includegraphics[width=1\textwidth]{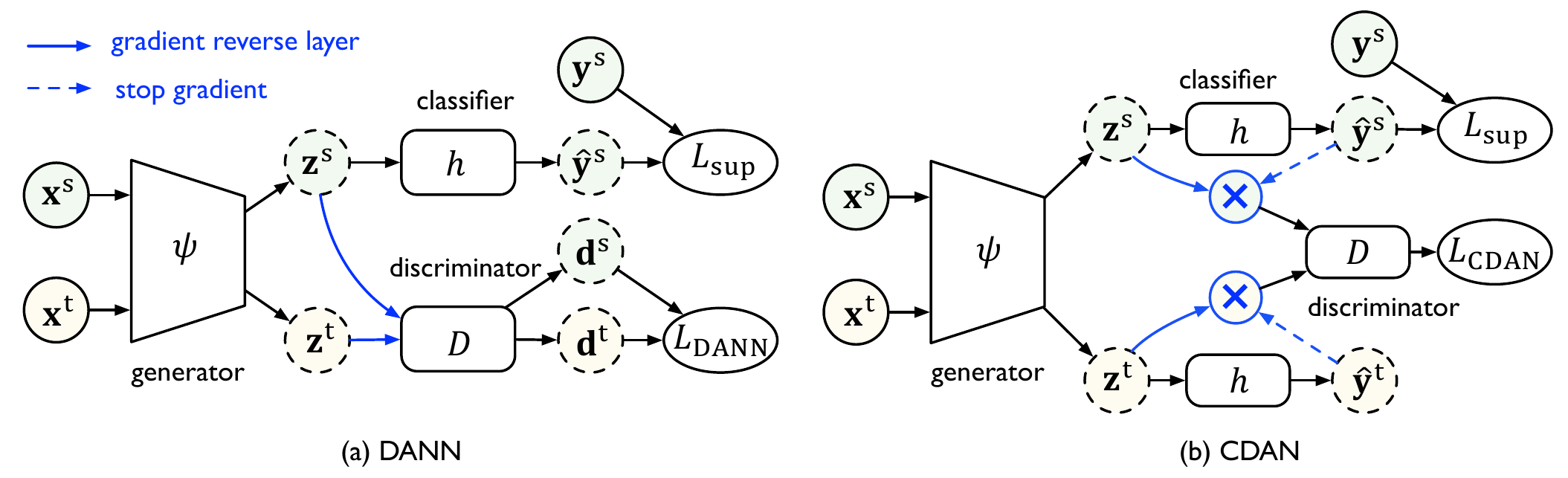}
	\caption{
		Both DANN and CDAN have a feature generator network $\psi$, a classifier $h$, and a domain discriminator $D$ connected to $\psi$ via a \textit{gradient reversal layer}.
		(a) In DANN, the discriminator $D$ is trained to distinguish between domains while the generator $\psi$ tries to make the feature distributions indistinguishable for the discriminator.
		(b) In CDAN, the discriminator $D$ is conditioned on the classifier prediction $\widehat{\mathbf{y}}$ via a multilinear map $\mathbf{z} \otimes \widehat{\mathbf{y}}$.
	}
	\label{fig:DANN}
\end{figure}

\subsubsection{Domain Adversarial Learning}\label{sec:DAAdversarialTraining}

\paragraph{Domain Adversarial Neural Network.}
An important milestone for modeling distributions is the Generative Adversarial Net (GAN) \citep{GAN}. Inspired by GAN, Domain Adversarial Neural Network (DANN) \citep{DANN, DANN_JMLR} integrates a two-player game into domain adaptation (Figure \ref{fig:DANN}). The first player is the domain discriminator $D$ trained to distinguish the source features from the target features and the second player is the feature generator $\psi$ trained simultaneously to confuse the domain discriminator.
As mentioned in Section \ref{sec:DA_theory}, the upper bound of $\mathcal H\Delta \mathcal H$-Divergence between feature distributions can be estimated by training the domain discriminator $D$,
\begin{equation}\label{equ:DANN}
	L_{\text{DANN}}(\psi) = \max_D \mathbb{E}_{\mathbf{x}^s\sim \widehat{\mathcal{S}}} \log [D(\mathbf{z}^s)] +  \mathbb{E}_{\mathbf{x}^t\sim \widehat{\mathcal{T}}} \log [1-D(\mathbf{z}^t)],
\end{equation}
where $\mathbf{z}=\psi(\mathbf{x})$ is the feature representation for $\mathbf{x}$.
The objective of the feature generator $\psi$ is to minimize the source error as well as the $\mathcal H\Delta \mathcal H$-Divergence bounded by Equation \ref{equ:DANN},
\begin{equation}
	\min_{\psi, h} \mathbb{E}_{(\mathbf{x}^s, \mathbf{y}^s) \sim \widehat{\mathcal{S}}} L_{\text{CE}}(h(\mathbf{z}^s), \mathbf{y}^s) + \lambda L_{\text{DANN}}(\psi),
\end{equation}
where $L_{\text{CE}}$ is the cross-entropy loss, $\lambda$ is a hyper-parameter that trades off source error and domain adversary. Minimizing the cross-entropy loss will lead to discriminative representations while decreasing the domain adversarial loss will result in transferable representations.

DANN aligns the marginal feature distributions through adversarial training.
However, this may be insufficient when the feature-label joint distributions change between domains.
Besides, the feature distribution is usually multimodal in multi-class classification, thus even if the discriminator is fully confused, there is no guarantee that the two feature distributions are similar  \citep{EquilibriumOfGAN}.
To tackle these two issues,
Conditional Domain Adversarial Network (CDAN) \citep{CDAN} conditions features $\mathbf{z}$ on classifier predictions $\widehat{\mathbf{y}}=h(\mathbf{z})$ and introduces multilinear map $\mathbf{z} \otimes \widehat{\mathbf{y}}$ instead of $\mathbf{z}$ as the input to domain discriminator $D$:
\begin{equation}\label{equ:CDAN}
	L_{\text{CDAN}}(\psi) = \max_D \mathbb{E}_{\mathbf{x}^s\sim \widehat{\mathcal{S}}} \log [D(\mathbf{z}^s \otimes \widehat{\mathbf{y}}^s)] +  \mathbb{E}_{\mathbf{x}^t\sim \widehat{\mathcal{T}}} \log[1- D(\mathbf{z}^t \otimes \widehat{\mathbf{y}}^t)].
\end{equation}
Conditioning on $\widehat{\mathbf{y}}$, CDAN fully captures cross-variance between the feature representation and classifier prediction, resulting in better alignment of the joint distributions.

\cite{SimultaneousDADomainAndTasks} align the class distributions explicitly by transferring the similarity structure in classes from the source domain to the target domain. Specifically, the average output probability of data from each class is computed over the source samples as soft labels. Then the model is optimized to match the distribution over classes to the soft labels.

\paragraph{Improvements.}\label{sec:DANN_drawbacks}
DANN integrates a \textit{gradient reverse layer} (GRL) into the standard architecture to implement the minimax between the feature generator and domain classifier. However, this optimizing strategy might not work well in practice due to gradient vanishing, which is also a major problem in training GANs. For instance, when the generated target features $\mathbf{z}^t$ are very distinguishable from source features such that $D(\mathbf{z}^t)=0$, the gradient for the feature generator is small and vice versa. This makes the optimization of the feature generator difficult.
Thus, Adversarial Discriminative Domain Adaptation (ADDA) \citep{ADDA} splits the optimization of feature generator $\psi$ and domain classifier $D$ into two independent objectives, where the maximization of $D$ remains unchanged, but the objective of $\psi$ becomes
\begin{equation}\label{equ:ADDA}
	\min_{\psi} \mathbb{E}_{\mathbf{x}^t\sim \widehat{\mathcal{T}}} -\log [D(\mathbf{z}^t)].
\end{equation}
This assigns small gradients for source-like target samples and larger gradients for the other target samples. It has the same fixed-point properties as GRL while making the optimization easier for the feature generator.
Although adversarial domain adaptation enhances the feature \emph{transferability}, i.e. the ability of feature representations to bridge the discrepancy across domains, studies \citep{bsp} reveal that it is at the expense of deteriorating the \emph{discriminability}, i.e. the easiness of separating categories over the fixed feature representations of both domains. Spectral analysis shows that only the eigenvectors corresponding
to the largest singular values tend to carry transferable
knowledge, while other eigenvectors may reflect domain
variations and thus be overly penalized in domain adversarial training. However, these eigenvectors may convey crucial discriminative information, and thus the discriminability is weakened.
To tackle this transferability-discriminability dilemma, Batch Spectral Penalization (BSP) \citep{bsp} penalizes the largest singular values so that the other eigenvectors
can be relatively strengthened to boost the feature
discriminability. Domain Separation Network (DSN) \citep{DSN} introduces a private subspace for each domain, which preserves
domain-specific information, such as background and low-level image statistics. Then domain alignment is performed safely in the shared subspace, which is orthogonal to the private subspace responsible for the discriminative tasks.

\paragraph{Domain Adversarial Leanring in Real-World Scenarios.}
Domain adversarial learning has largely improved the performance on unlabeled target domain  and has been widely adopted in many applications \citep{FCN_in_wild, DAFaster}.
However, real-world scenarios are much more complex. Here we list several situations that are often encountered yet not well resolved, and review some existing solutions to them.

\begin{figure}[!t]
	\centering
	\includegraphics[width=1.\textwidth]{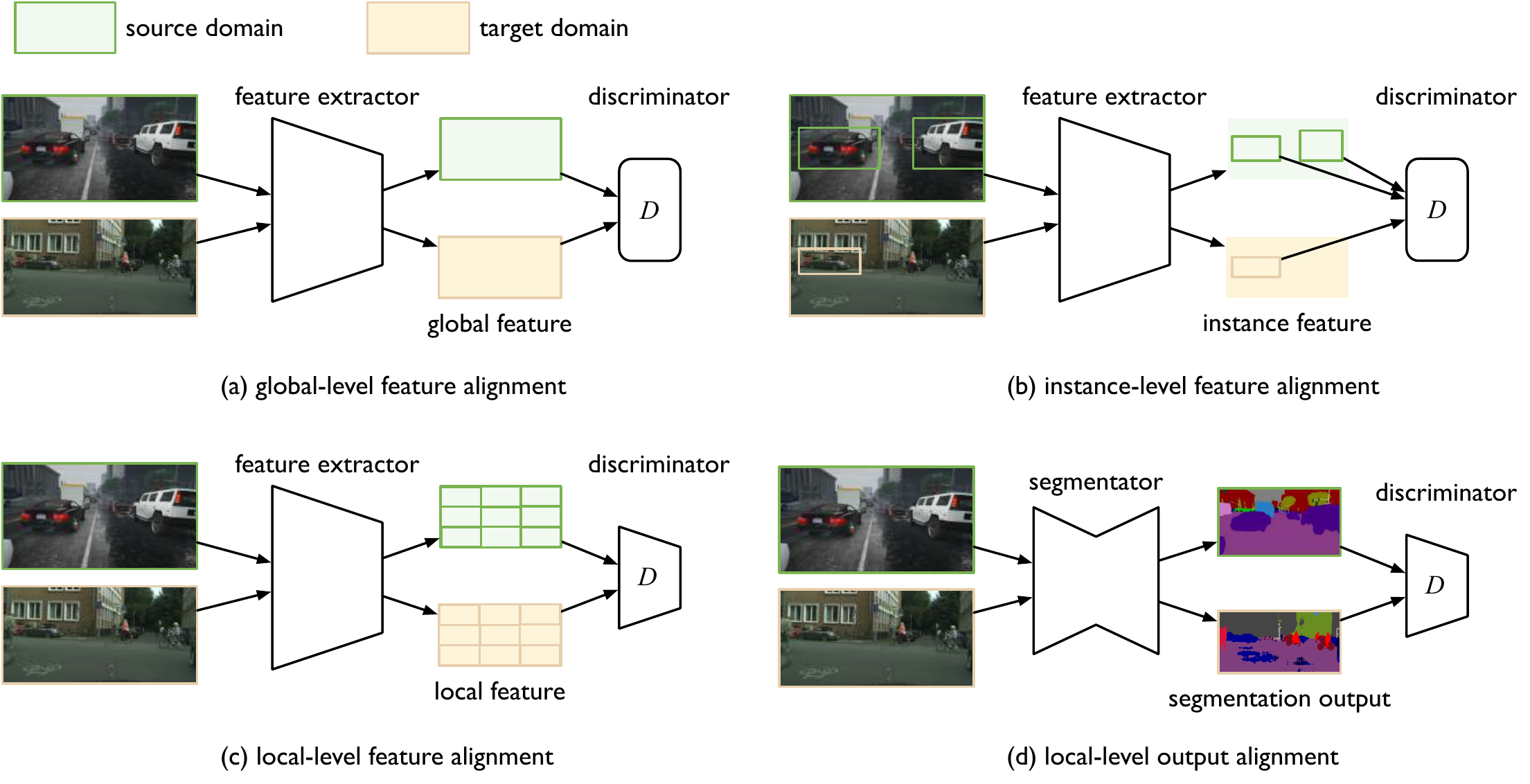}
	\caption{
		Where to adapt in different methods. (a) DANN performs alignment on global features. (b) DA-Faster performs alignment on both image-level and instance-level features. (c) SWDA performs alignment on local features. (d) Adapt-SegMap performs alignment on local outputs.
	}
	\label{fig:where_to_adapt}
	\vspace{-5pt}
\end{figure}

\textit{Which part to adapt is unknown.} In image recognition, we only need to classify the input. Yet in applications such as object detection, we need to locate the \textit{Region of Interests} (RoIs) first and then classify them.
Due to the distribution shift across domains, the localization of RoIs on the target domain is not reliable, thus which part to adapt in the adversarial training is unknown.
To alleviate this problem, as shown in Figure \ref{fig:where_to_adapt},
DA-Faster \citep{DAFaster} performs domain alignment at both image level and instance level, where the image-level alignment is believed to improve the localization of RoIs on the target domain.
SWDA \citep{SWDA} argues that alignment of the local features is more effective than that of the global features for better localization.
Although the above adversarial training methods have improved the transferability of object detectors,
the discriminability might get lost in
the adversarial adaptation process as mentioned by BSP \citep{bsp}. Since discriminability is crucial for the localization of RoIs, D-adapt \citep{jiang2021decoupled} introduces parameter-independent category adaptor and bounding box adaptor to decouple adversarial adaptation from detector training, which yields sharp improvement.

\textit{There are structural dependencies between labels of each sample.} In low-level classification tasks, such as semantic segmentation or token classification (Named Entity Recognition, Parts-of-speech tagging, etc.), adaptation on \emph{features} \citep{FCN_in_wild} may not be a good option, because the feature of each pixel or each token is high-dimensional and there exist many pixels or tokens in a single sample. Every coin has two sides. Compared to the high-level classification tasks, the \emph{output space} of these low-level tasks contains much richer information of the distribution, e.g., scene layout and context, and thus adaptation on the output space can reduce the domain shift. As shown in Figure \ref{fig:where_to_adapt}, Adapt-SegMap \citep{AdaptSegNet} trains a discriminator to distinguish whether the segmentation output is from the source or the target, while the feature generator is encouraged to generate similar segmentation outputs across domains. It explicitly aligns the output distributions of target and source domains, and implicitly adapts the feature distributions.
Further, ADVENT \citep{Advent} minimizes the distribution distance on the \textit{self-information} distributions, where the entropy of segmentation outputs is fed to the discriminator. In this way, the conditional information is neglected while more attention is paid to the structural dependencies between local semantics.

\subsubsection{Hypothesis Adversarial Learning}\label{sec:DADisparityDiscrepancy}

\begin{figure}[!t]
	\centering
	\includegraphics[width=1.\textwidth]{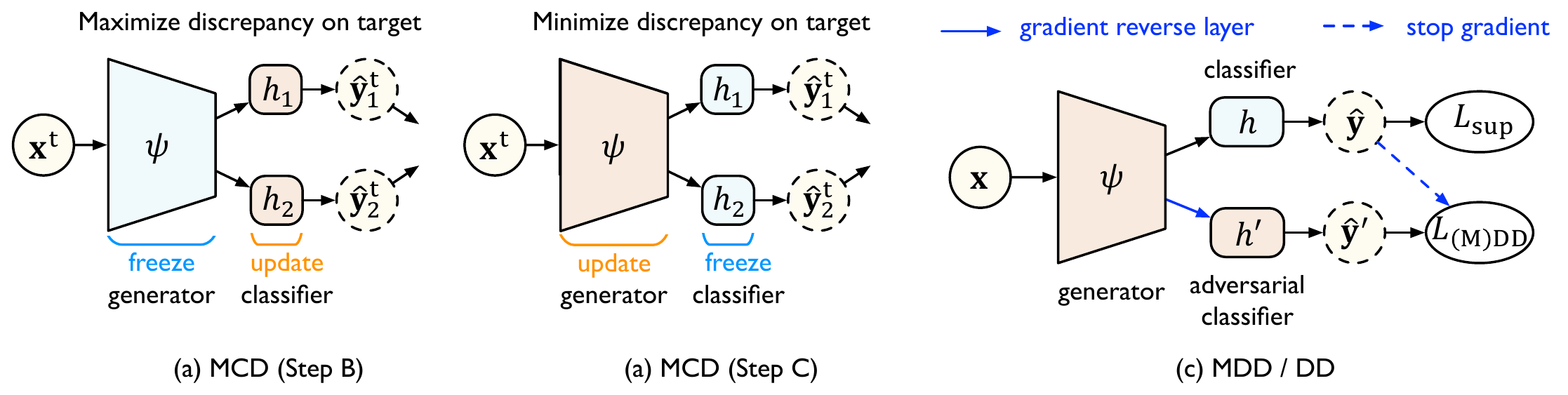}
	\caption{
		(a) The training of MCD has three steps. Step A: Both the classifiers and the feature generator are trained to classify the source samples correctly.  Step B: The classifiers $h_1$ and $h_2$ learn to maximize the discrepancy on the target samples.
		(b) Step C: The feature generator $\psi$ learns to minimize the discrepancy on the target samples.
		(c) DD and MDD introduce an adversarial classifier $h'$ to maximize the discrepancy and trains the feature generator $\psi$ to minimize the source error as well as the discrepancy.
	}
	\label{fig:MDD}
\end{figure}

Inevitably, there is still a gap between the \textit{proxy distance} used in previous methods and the hypothesis-induced discrepancies in the theories. To close this gap, Maximum Classifier Discrepancy (MCD) \citep{MCD} starts to estimate and optimize $\mathcal H\Delta \mathcal H$-Divergence in a fully parameterized way. As shown in Figure \ref{fig:MDD}, MCD maximizes the discrepancy between two classifiers' outputs to detect target samples far from the support of source distribution, i.e. estimate $\mathcal H\Delta \mathcal H$-Divergence. A feature generator then learns to generate target features near the support to minimize the discrepancy, i.e. minimize the domain discrepancy. MCD uses the $L_1$ distance to calculate the discrepancy, while Sliced Wasserstein Discrepancy (SWD) \citep{SWD} adopts the Wasserstein metric, which is the natural geometry for probability measures induced by the optimal transport theory.
In theory MCD is closer to $\mathcal H\Delta \mathcal H$-Divergence, yet in experiments it is slow in convergence and very sensitive to hyper-parameters. The reason is that there exist two classifiers $h$ and $h'$ in MCD that maximize the discrepancy, which makes the minimax optimization hard to reach equilibrium.

Disparity Discrepancy (DD) \citep{MDD} provides a tighter bound by taking supremum in hypothesis space $\mathcal H$ rather than $\mathcal H\Delta \mathcal H$. This will significantly ease the minimax optimization. As shown in Figure \ref{fig:MDD}, DD introduces an adversarial classifier $h'$ sharing the same hypothesis space with $h$. The supremum in $d_{h,\mathcal H}(\mathcal{S},\mathcal{T})$ is approximated by
\begin{equation}
	\begin{split}
		L_{\text{DD}} (h, \psi) = \max_{h'} & \
		\mathbb{E}_{\mathbf{x}^s \sim \widehat{\mathcal{S}}} L^s\left[h'(\psi(\mathbf{x}^s)), h(\psi(\mathbf{x}^s))\right] \\
		- & \ \mathbb{E}_{\mathbf{x}^t \sim \widehat{\mathcal{T}}} L^t\left[h'(\psi(\mathbf{x}^t)), h(\psi(\mathbf{x}^t))\right], \\
	\end{split}
\end{equation}
where $L^s$ and $L^t$ are specific loss functions defined on the source domain and target domain respectively.
Based on the theory~\citep{MDD}, when the adversarial classifier $h'$ is close to the supremum, minimizing the following terms will decrease the target error $\epsilon_{\mathcal{T}}$,
\begin{equation}
	\min_{\psi, h} \
	\mathbb{E}_{(\mathbf{x}^s, \mathbf{y}^s) \sim \widehat{\mathcal{S}}} L_{\text{CE}}(h(\psi(\mathbf{x}^s)), \mathbf{y}^s)
	+ \lambda L_{\text{DD}} (h, \psi),
\end{equation}
where $\lambda$ is a tradeoff hyper-parameter.
An intuitive explanation is that DD is looking for an adversarial classifier $h'$ that predicts correctly on the source domain while making different predictions from $h$ on the target domain. And then the feature generator $\psi$ is encouraged to generate features near the decision boundary to avoid such situations.

However, DD is still limited to the $01$ loss in the classification setting. Based on scoring functions and margin loss, Margin Disparity Discrepancy (MDD) \citep{MDD} goes a crucial step forward and provides a \emph{margin} theory for the multi-class classification setting. The margin $\rho$ is attained by introducing parameter $\gamma \triangleq \exp\rho$ in the disparity discrepancy,
\begin{equation}\label{equ:MDD}
	\begin{split}
		L_{\text{MDD}} (h, \psi)
		= \max_{h'} \
		\gamma & \ \mathbb{E}_{\mathbf{x}^s \sim \widehat{\mathcal{S}}}
		\log \left[\sigma_{h(\psi(\mathbf{x}^s))} (h'(\psi(\mathbf{x}^s)))\right] \\
		+ & \ \mathbb{E}_{\mathbf{x}^t \sim \widehat{\mathcal{T}}}
		\log \left[1-\sigma_{h(\psi(\mathbf{x}^t))} (h'(\psi(\mathbf{x}^t)))\right], \\
	\end{split}
\end{equation}
where $\sigma$ is the softmax function.
A proper $\gamma$ can constrain $h'$ in a hypothesis space of proper size to avoid overestimation of the generalization bound.
Note that in Equation \ref{equ:MDD}, the loss on the source domain is the standard cross-entropy, while that on the target domain is a modified cross-entropy to avoid gradient vanishing and ease the optimization of $h'$.

In principle, DD can be easily extended to regression problems by replacing the classifiers in Figure \ref{fig:MDD} with regressors and choosing $L^s$ and $L^t$ as the $L_1$ or $L_2$ loss commonly used in regression.
It has been extended to both keypoint detection \citep{RegDA} and bounding box localization task \citep{jiang2021decoupled}.
To tackle the challenge caused by the high-dimensional output space in the keypoint detection, Regressive
Domain Adaptation (RegDA) \citep{RegDA} introduces a spatial probability distribution to describe the sparse density of the output space and uses it to guide the optimization of the adversarial regressor $h'$. In an expectation sense, this reduces the size of the hypothesis space of $h'$ and avoids overestimation of the generalization bound in \cite{MDD}.

\subsubsection{Domain Translation}\label{sec:DATranslation}

Domain translation is the task of mapping \emph{raw data} of text, image, audio, and other data modality from the source distribution $\mathcal{S}$ to the target distribution $\mathcal{T}$. In domain adaptation problems, we can use translation models,
usually based on Generate Adversarial Networks (GAN)~\citep{GAN}, to obtain labeled source domain in the target style, i.e. translated into the target distribution. Training on such stylized source domain can yield better transferability than models trained on the original source domain.

\begin{figure}[!t]
	\centering
	\includegraphics[width=1\textwidth]{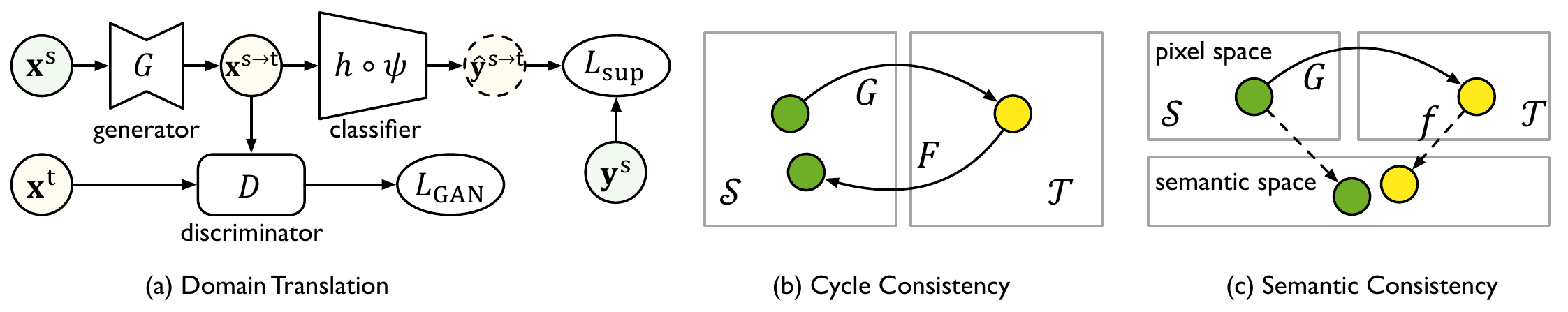}
	\caption{
	{(a)}
	The architecture for PixelDA includes a generator network  $G$, an adversarial discriminator $D$ and a task-specific classifier $h$ on feature extractor $\psi$.
	{(b)} Cycle-consistency loss: after we translate from source domain to target domain, we should recover the source data if translating back again.
	{(c)} Semantic-consistency loss: translating between domains should not change the semantic labels of the samples, where $f$ is the labeling function.
	}
	\label{fig:GAN_based}
\end{figure}

GANs reason about the marginal distribution, i.e., from a random vector, the generator network should synthesize data that resembles one that is drawn from the true distribution.
However, marginal distribution is not enough for domain adaptation, thus Coupled Generative Adversarial Networks (CoGAN) \citep{CoGAN} learns a joint distribution of multi-domain images from data, i.e., from a random vector, multiple generators should generate paired data that are from different distributions and share the same labels.
By enforcing a weight-sharing constraint between different generators, CoGAN learns a joint distribution without the existence of corresponding images in different domains. Then the shared labels of the target samples are used to train the target model.

A more common objective of domain translation is to learn a mapping $G:\mathcal{S}\rightarrow\mathcal{T}$ such that the generated sample $G(\mathbf{x})$ is indistinguishable from the training samples of the target domain. As shown in Figure \ref{fig:GAN_based}, PixelDA \citep{PixelDA} introduces an adversarial discriminator $D$ to distinguish between translated samples and target samples,
\begin{equation}
	\begin{aligned}
		L_\mathrm{GAN}(G) & = \max_{D} \mathbb{E}_{\mathbf{x}\sim \widehat{\mathcal{S}}} \log
		\left[ 1-D(G(\mathbf{x})) \right]
		+ \mathbb{E}_{\mathbf{x}\sim \widehat{\mathcal{T}}} \log
		\left[ D(\mathbf{x}) \right].  \\
	\end{aligned}
\end{equation}
The generator $G$ tries to synthesize samples $G(\mathbf{x})$ that look similar to images from the target domain by $\min_{G} L_\mathrm{GAN}(G)$.
The task-specific classifier $h$ and feature extractor $\psi$ are trained supervisedly on the target-style generated data by $\min_{\psi, h} \mathbb{E}_{(\mathbf{x}, \mathbf{y})\sim \widehat{\mathcal{S}}} L_{\text{sup}}(h\circ \psi (G (\mathbf{x})), \mathbf{y})$.

\paragraph{Cycle Consistency.} While GAN can learn a mapping between two datasets, the desired mapping may not be obtained. The source sample may be projected to an irrelevant target sample, destroying the structure or content of the original samples. Besides, multiple source samples may be mapped to the same target sample, leading to the well-known problem of \emph{mode collapse} \citep{GAN}.
Therefore, CycleGAN \citep{CycleGAN} introduces an additional mapping $F:\mathcal{T}\rightarrow\mathcal{S}$ from target to source and adds a constraint of \textit{cycle consistency} to reduce the space of possible mapping functions (Figure \ref{fig:GAN_based}). Mathematically, cycle consistency requires $F$ and $G$ to be bijections and inverse of each other. In practice, CycleGAN constrains $F(G(\mathbf{x}))  \approx \mathbf{x}$ and $G(F(\mathbf{x})) \approx \mathbf{x}$, which preserves the structure or content of samples to obtain more meaningful mappings.
CycleGAN has been widely used in domain adaptation problems, such as image classification \citep{cycada}, semantic segmentation \citep{cycada}, person re-identification \citep{PersonGAN}, robotic grasping \citep{DARoboticGrasping}, object detection \citep{diversify}, etc.
The idea goes beyond the field of image translation and is widely used in other fields, such as unsupervised machine translation \citep{unsupervised_machine_translation}, where the cycle consistency is also called back-translation. Unsupervised machine translation can further be used for cross-lingual domain adaptation tasks \citep{XNLI}, where a language is a domain.

\paragraph{Semantic Consistency.} CycleGAN is a general-purpose translation model for vision tasks and is apt at style transfer between datasets.
However, it is difficult for CycleGAN to maintain the semantic information.
It has been experimentally shown that when mapping from source to target, the problem of label flipping will easily occur \citep{PixelDA, cycada}. As a result, there will exist a lot of noisy labels in the translated dataset, hurting the performance of the target model.
Thus, ensuring semantic consistency is important for translation-based domain adaptation.
Formally, given the labeling function $f$, the labels assigned to sample $\mathbf{x}$  should be consistent with that of the translated sample, i.e., $f(\mathbf{x}) = f(G(\mathbf{x}))$ (Figure \ref{fig:GAN_based}).
Since function $f$ is not accessible, several proxy functions have been proposed to approximate the semantic consistency \citep{DTN, cycada, DARoboticGrasping}. Given a proxy function $h_{\mathrm{p}}$ and a distance measure $d$, the objective is to reduce the semantic inconsistency,
\begin{equation}
	\min_{G}L_{\mathrm{sc}}(G, h_{\mathrm{p}}) = d(h_{\mathrm{p}}(\mathbf{x}), h_{\mathrm{p}}(G(\mathbf{x}))).
\end{equation}
DTN \citep{DTN} and SimGAN \citep{SimGAN} use the feature extractor as a proxy function and the goal is to translate the low-level style while keeping the high-level features invariant.
PersonGAN \citep{PersonGAN} uses the foreground crop of the person image as the proxy function of the person's identity, which ensures that a person's identity remains the same before and after translation.
However, the constraint on feature or pixel space might be too strong, making it difficult to change the low-level style.
Therefore, Cycle-consistent Adversarial Adaptation (CyCADA) \citep{cycada} utilizes a pre-trained source model as a proxy function to encourage generating samples that have consistent predictions under the function. This proxy function effectively avoids label flipping during the translation of handwritten digit images, yet it is still not perfect.
When faced with the dataset shift of real-world problems, the predictions on the generated samples are not reliable and may provide incorrect guidance to $G$.

When the domain difference is primarily low-level, such as textures, illumination, and color, translation can effectively close the domain gap.
But when the domain is different at the high level, such as from different camera angles, translation may fail to adapt domains.
Therefore,
translation methods at the low-level and adaptation methods at the high-level mentioned in Section \ref{sec:DAMomentMatching}-\ref{sec:DADisparityDiscrepancy} are complementary and can be combined in practical applications \citep{cycada}. For example, Generate to Adapt \citep{Gen2Adapt} directly uses the generative task as an auxiliary task to align features across domains.

\subsubsection{Semi-Supervised Learning}\label{sec:DA_self_training}
\label{sec:semi_supervised}

Unsupervised Domain Adaptation (UDA) is closely related to Semi-Supervised Learning (SSL) since both of them aim at generalizing from the labeled samples to the unlabeled samples. The difference is that in SSL, both the labeled and unlabeled samples come from the same distribution while in UDA, the source and target distributions differ. Thus, SSL tasks can be considered as a special case of UDA tasks and some SSL methods can be applied in UDA tasks.
Since there is still no theoretical guarantee for SSL methods in the UDA scenario, the first question to answer is, under what assumptions can we use SSL in UDA? There are mainly three assumptions in SSL \citep{SSL}.
(1) \textit{Smoothness Assumption}: if two samples $\mathbf{x}_1$, $\mathbf{x}_2$ residing in a high-density region are close, then so should be their corresponding outputs $\mathbf{y}_1$, $\mathbf{y}_2$ .
(2) \textit{Cluster Assumption}: if points are in the same cluster, they are likely to be of the same class. It can also be interpreted as the \textit{Low-density Separation Assumption}, where the decision boundary should lie in the low-density regions.
(3) \textit{Manifold Assumption}: the high-dimensional data shall lie roughly on a low-dimensional manifold.
Both smoothness assumption and cluster assumption are helpful for classification, but not for regression problems. Thus SSL is used more commonly in classifier adaptation.
Here we review several SSL methods applied to the UDA problems.

\textbf{Consistency Regularization} encourages consistent predictions for similar data points.
Similar data points are generated by performing different data augmentations on the same data point. While many augmentation techniques are proposed for images, few are available for other data formats, such as texts and time series.
Thus this type of method is limited to certain data modalities.
Self-Ensemble \citep{SelfEnsemble} applies mean-teacher, a typical consistency regularization method, to image domain adaptation. The teacher model, which is an Exponential Moving Average (EMA) of the student model, will generate predictions to train the student model on the target domain. Due to the domain shift, the predictions are noisy, thus Mutual Mean-Teaching (MMT)~\citep{ge2020mutual} uses two collaborative networks jointly optimized under the supervision of mutual teacher models.

\textbf{Entropy Minimization} encourages the model to make confident (i.e., low-entropy) predictions on unlabeled data. It serves as an auxiliary term in many domain adaptation methods \citep{RTN, CDAN, MCD, DIRT, Advent}. The risk is that the predictions on the target domain are not reliable, and entropy minimization may hurt the performance of the model.
Thus, Minimum Class Confusion (MCC) \citep{MCC} introduces a weight for each instance, where uncertain samples have smaller weights to avoid minimizing entropy on the incorrectly classified samples.
MCC further minimizes the instance-weighted confusion
between different classes, which is simple yet frustratingly effective.
Source Hypothesis Transfer \citep{liang2020shot}
adopts an information maximization loss with a fair diversity-promoting objective, which circumvents the trivial
solutions in entropy minimization that all unlabeled data have the same one-hot
encoding.

\textbf{Pseudo-Labeling} produces proxy labels on unlabeled data and uses these noisy labels together with the labeled data to train the model.
In self-training, a confidence threshold is used to filter out unreliable proxy labels, which may fail in UDA since the model is likely to be biased towards well-transferred classes while ignoring other hard classes. Thus, Class-Balanced Self-Training (CBST) \citep{CBST} uses a class-wise confidence threshold.
Still, large noise exists in the generated pseudo labels on the target domain, and the standard Cross-Entropy (CE) loss has been shown to be sensitive to label noise \citep{RethinkGeneralization}. Towards this problem, \cite{GCE} propose the Generalized Cross-Entropy (GCE) loss as an effective solution \citep{rusak2021selflearning, liu2021cycle},
\begin{equation}
	L_{\text{GCE}}(\mathbf{x}, \tilde{y})=1/q \cdot ( 1-h_{\tilde{y}}(\mathbf{x})^q ),
\end{equation}
where $q\in(0,1]$ is a hyper-parameter to trade-off between the CE loss and the MAE loss.

\subsubsection{Remarks}

Different domain adaptation methods are compared from several perspectives in Table \ref{table:DA_comparison}. First, statistics matching, domain adversarial learning, and hypothesis adversarial learning methods are derived from theory, enjoying theoretical guarantees while domain translation and semi-supervised learning methods are still in the empirical regime.
Second, the former three categories of methods work in the feature space or the output space and are highly related to specific tasks, and some are tightly integrated to specific architectures. In contrast, translation methods work in the input space and are relatively independent of specific tasks.
However, translation models and semi-supervised learning are dependent on specific data format, and are hard to scale to different modalities.
Finally, statistics matching methods are based on nonparametric distances, which are data-efficient but weak in expressiveness, thereby more suitable for low-data regimes. In contrast, domain adversarial learning and hypothesis adversarial learning methods are based on parametric distances, which can only be measured throughout learning, but are more performant when scaling up data.

\begin{table}[htbp]
	\addtolength{\tabcolsep}{1pt}
	\centering
	\scriptsize
	\caption{Comparison between different domain adaptation methods.}
	\label{table:DA_comparison}
	\begin{threeparttable}
		\begin{tabular}{lccccc}
			\toprule
			                                & \begin{tabular}[c]{@{}c@{}}Adaptation \\ Performance\tnote{1}\end{tabular} & \begin{tabular}[c]{@{}c@{}}Data \\ Efficiency\tnote{2}\end{tabular} & \begin{tabular}[c]{@{}c@{}}Modality \\ Scalability\tnote{3}\end{tabular} & \begin{tabular}[c]{@{}c@{}}Task\\  Scalability\tnote{4}\end{tabular} & \begin{tabular}[c]{@{}c@{}} Theory \\ Guarantee\tnote{5}\end{tabular} \\
			\midrule
			Statistics Matching             & $\bigstar$                     & $\bigstar\bigstar\bigstar$     & $\bigstar\bigstar\bigstar$     & $\bigstar\bigstar$             & $\bigstar\bigstar\bigstar$     \\
			\midrule
			Domain Adversarial Learning     & $\bigstar\bigstar$             & $\bigstar\bigstar$             & $\bigstar\bigstar\bigstar$     & $\bigstar\bigstar$             & $\bigstar\bigstar\bigstar$     \\
			\midrule
			Hypothesis Adversarial Learning & $\bigstar\bigstar\bigstar$     & $\bigstar\bigstar$             & $\bigstar\bigstar\bigstar$     & $\bigstar\bigstar$             & $\bigstar\bigstar\bigstar$     \\
			\midrule
			Domain Translation              & $\bigstar\bigstar$             & $\bigstar$                     & $\bigstar$                     & $\bigstar\bigstar\bigstar$     & $\bigstar$                     \\
			\midrule
			Semi-Supervised Learning        & $\bigstar\bigstar$             & $\bigstar\bigstar$             & $\bigstar\bigstar$             & $\bigstar$                     & $\bigstar$                     \\
			\bottomrule
		\end{tabular}
		\begin{tablenotes}
			\scriptsize
			\item[1] Performance: performance when there are large-scale data in source and target domains.
			\item[2] Data Efficiency: performance when there are only small-scale data in source and target domains.
			\item[3] Modality Scalability: whether can adapt the model to various modalities, such as text, time series.
			\item[4] Task Scalability: whether can adapt the model to different tasks, such as regression, detection.
			\item[5] Theory Guarantee: whether the generalization error of target domain can be bounded in adaptation.
		\end{tablenotes}
	\end{threeparttable}
\end{table}

Domain adaptation is closely related to pre-training and task adaptation.
First, pre-training can boost the \emph{transferability} in domain adaptation, since pre-training will reduce the allowed hypothesis space and decrease the generalization bound on the target domain, as mentioned in Section \ref{sec:DA_theory}. Thus pre-training on the source domain also serves as the first step in many domain adaptation methods, such as RegDA \citep{RegDA}. Pre-training also provides some new solutions for domain adaptation. When there exists a large unlabeled target domain, a feasible solution is to first perform unsupervised pre-training on the target domain, and then fine-tune with the labeled data on the source domain. This is widely adopted in cross-lingual adaptation \citep{cite:Arxiv19XLM}.

When using pre-trained models for domain adaptation, we will also encounter the problems in task adaptation, such as the \emph{catastrophic forgetting} mentioned in \ref{sec:CatastrophicForgetting}.
Thus, many domain adaptation methods babysit the learning rates to avoid catastrophic forgetting \citep{DAN, DANN}.
Compared with task adaptation, domain adaptation increases the restriction on the task space, where the task of the source domain and that of the target domain must be the same. Due to this restriction, domain adaptation has a strict theoretical guarantee.
But this restriction is sometimes hard to satisfy in practice since we cannot ensure whether the category on the unlabeled target domain is exactly the same as the source domain \citep{OpensetDA}.
Therefore, real-world adaptation is often a mix of task adaptation and domain adaptation. How to explore the transferability in such a practical \emph{open-domain} scenario is a problem to be solved.

\section{Evaluation}

Evaluation serves as a means for (1) measuring the performance of different architectures, different pre-training and adaptation methods, and
(2) understanding the strengths and limitations of different methods.
This section will elaborate on the evaluation of \emph{transferability}, which is defined by the performance on the target task or domain.
We believe that the evaluation of different methods should be performed on large-scale datasets for a practical and meaningful comparison.
Thus, in Section \ref{sec:dataset} we list some large-scale datasets that are suitable for evaluating transferability in deep learning.
Since different methods are often based on different codebases, a fair comparison between them is rather difficult. To fill this blank, in Section \ref{sec:Library} we propose an open-source library, \texttt{TLlib}, to better evaluate transferability of different methods in a unified framework. Finally, Section \ref{sec:benchmark} provides several benchmarks for evaluating both the cross-task transferability and cross-domain transferability.

\subsection{Datasets}
\label{sec:dataset}

To evaluate the transferability in deep learning, we list several datasets that are large-scale in the number of samples and categories, the richness of tasks, and the diversity of domains.

The \textit{General Language Understanding Evaluation} (GLUE) \citep{GLUE} is one of the most famous benchmarks in NLP. As shown in Table \ref{table:GLUE}, it consists of nine sentence or sentence-pair language understanding tasks, covering a diverse range of dataset sizes, text genres, and degrees of difficulty. It is widely used to evaluate the \emph{cross-task} transferability of different pre-training and task adaptation methods.

\begin{table}[htbp]
	\scriptsize
	\centering
	\caption{Descriptions and statistics of the GLUE datasets.}
	\label{table:GLUE}
	\begin{tabular}{llllll}
		\toprule
		Corpus & \#Train & \#Test & Metrics                      & Task                & Domain              \\ \midrule
		CoLA   & 8.5k    & 1k     & Matthews corr                & acceptability       & misc.               \\
		SST-2  & 67k     & 1.8k   & acc.                         & sentiment           & movie reviews       \\
		MRPC   & 3.7k    & 1.7k   & acc./F1                      & paraphrase          & news                \\
		STS-B  & 7k      & 1.4k   & Pearson/Spearman corr        & sentence similarity & misc.               \\
		QQP    & 364k    & 391k   & acc./F1                      & paraphrase          & social QA questions \\
		MNLI   & 393k    & 20k    & matched acc./mismatched acc. & NLI                 & misc.               \\
		QNLI   & 105k    & 5.4k   & acc                          & QA/NLI              & Wikipedia           \\
		RTE    & 2.5k    & 3k     & acc                          & NLI                 & news, Wikipedia     \\
		WNLI   & 634     & 146    & acc                          & coreference/NLI     & fiction books       \\ \bottomrule
	\end{tabular}
\end{table}

In contrast, there is no common benchmark to evaluate the transferability of different methods in computer vision. Table \ref{table:vision_datasets} lists some of the widely used vision datasets.
\textit{Food-101, CIFAR-10, CIFAR-100, SUN397, Stanford Cars, FGVC Aircraft, DTD, Oxford-III Pets, Caltech-101, Oxford 102 Flowers} are used to evaluate the transferability of different architectures under task discrepancy \citep{cite:CVPR19DoBetterTransfer}.
\textit{ImageNet-R(endition)} \citep{ImageNetR} and \textit{ImageNet-Sketch} \citep{ImageNetSketch} are two variants of the ImageNet, mainly used to evaluate the \emph{cross-domain} transferablity of different architectures and pre-training methods. \textit{DomainNet} \citep{domainnet} has multiple domains sharing the same category space, and is used to evaluate the \emph{cross-domain} transferablity of different domain adaptation methods under large domain shift.

\begin{table}[htbp]
	\scriptsize
	\centering
	\caption{Descriptions and statistics of typical vision datasets.}
	\label{table:vision_datasets}
	\begin{tabular}{llllll}
		\toprule
		Dataset                        & \#Train & \#Test & \#Classes & Metric                         & Domain                         \\ \midrule
		Food-101                       & 75,750  & 25,250 & 101       & top-1                          & photos and real world images   \\
		CIFAR-10                       & 50,000  & 10,000 & 10        & top-1                          & photos and real world images   \\
		CIFAR-100                      & 50,000  & 10,000 & 100       & top-1                          & photos and real world images   \\
		SUN397                         & 19,850  & 19,850 & 397       & top-1                          & photos and real world images   \\
		Stanford Cars                  & 8,144   & 8,041  & 196       & top-1                          & photos and real world images   \\
		FGVC Aircraft                  & 6,667   & 3,333  & 100       & \begin{tabular}[c]{@{}l@{}}mean \\ per-class\end{tabular} & photos and real world images   \\
		\begin{tabular}[c]{@{}l@{}}Describable \\ Textures (DTD)\end{tabular} & 3,760   & 1,880  & 47        & top-1                          & photos and real world images   \\
		Oxford-III Pets                & 3,680   & 3,369  & 37        & \begin{tabular}[c]{@{}l@{}}mean \\ per-class\end{tabular} & photos and real world images   \\
		Caltech-101                    & 3,060   & 6,084  & 102       & \begin{tabular}[c]{@{}l@{}}mean \\ per-class\end{tabular} & photos and real world images   \\
		Oxford 102 Flowers             & 2,040   & 6,149  & 102       & \begin{tabular}[c]{@{}l@{}}mean \\ per-class\end{tabular} & photos and real world images   \\ \midrule
		ImageNet-R                     & -       & 30k    & 200       & top-1                          & \begin{tabular}[c]{@{}l@{}}art, cartoons, deviantart, graffiti,\\  embroidery, graphics, origami, \\ paintings, patterns, plastic objects, \\ plush objects, sculptures, sketches,\\  tattoos, toys, and video games\end{tabular} \\
		ImageNet-Sketch                & -       & 50k    & 1000      & top-1                          & sketch                         \\ \midrule
		DomainNet-c                    & 33,525  & 14,604 & 365       & top-1                          & clipart images                 \\
		DomainNet-p                    & 50,416  & 21,850 & 365       & top-1                          & artistic paintings             \\
		DomainNet-r                    & 120,906 & 52,041 & 365       & top-1                          & photos and real world images   \\
		DomainNet-s                    & 48,212  & 20,916 & 365       & top-1                          & sketch                         \\ \bottomrule
	\end{tabular}
\end{table}

\subsection{Library}
\label{sec:Library}

To make up for the lack of a unified codebase in some areas,
we propose an open and ongoing library, \texttt{TLlib}.
This library implements many representative adaptation algorithms in a unified way, allowing quantitative, fair, reproducible comparisons between different algorithms and promoting seamless integration of different pre-training or adaptation methods.

\paragraph{Library Usage.}
First, we give a short description of how to use \texttt{TLlib} using DANN as an instance.
In the original implementation of DANN, the domain adversarial loss, domain discriminator, feature generator, and classifier are tightly coupled together in one \texttt{nn.Module}, which causes the difficulty of reuse, e.g., the entire algorithm needs re-implementation when the input data is changed from image to text. Yet in this case, the domain adversarial loss and the domain discriminator remain unchanged and shall be reused.
Therefore, in \texttt{TLlib}, models and loss functions are decoupled.
When using DANN for any case, users need only to initialize a domain discriminator and pass it to the domain adversarial loss module, and then use this module in the same way as the cross-entropy loss module defined in PyTorch (example code below).
\texttt{TLlib} provides friendly and coherent APIs for supported algorithms. Detailed usages of these algorithms can be found at the
\href{http://tllib.thuml.ai/}{documentation}.

\begin{lstlisting}
>>> # define the domain discriminator
>>> from dalib.modules.domain_discriminator import DomainDiscriminator
>>> discriminator = DomainDiscriminator(in_feature=1024, hidden_size=1024)
>>> # define the domain adversarial loss module
>>> from dalib.adptation.dann import DomainAdversarialLoss
>>> dann = DomainAdversarialLoss(discriminator, reduction='mean')
>>> # features from the source and target domain
>>> f_s, f_t = torch.randn(20, 1024), torch.randn(20, 1024)
>>> # calculate the final loss
>>> loss = dann(f_s, f_t)
\end{lstlisting}

\paragraph{Design Philosophy.}
\texttt{TLlib} is designed to be \textit{extendible} by researchers and \textit{simple} for practitioners.
Currently, there are mainly two types of algorithm implementations. One is to encapsulate each algorithm in a Trainer, whose typical representative is \texttt{PyTorch-Lighting}. Users only need to feed the training data to it and do not need to care about the specific training process. Another strategy is to encapsulate the core loss function in each algorithm, and users need to implement the complete training process by themselves. A typical representative is \texttt{PyTorch} \citep{PyTorch}.
Although the former method is easier to use, it is less extendible. Since it is often necessary to adjust the training process in different transfer learning scenarios, \texttt{TLlib} adopts the latter method for better extendibility.
We try our best to make \texttt{TLlib} easy to start with, e.g.,
we support the automatic download of most common transfer learning datasets so that users do not need to spend time on data preparation.
Our code is in \emph{PyTorch-style} and we provide training examples of different transfer algorithms in different scenarios, which allows users to quickly adapt to \texttt{TLlib} as long as they have learned PyTorch before.
For more convenient algorithm selection, we provide a comprehensive benchmark among all those libraries. For faster algorithm reproduction, we provide training scripts for all the results in the benchmark.

\texttt{TLlib} is released under the MIT License and
available at \url{https://github.com/thuml/Transfer-Learning-Library}.
Documentation and tutorials are available on its \href{http://tllib.thuml.ai/}{website}.

\subsection{Benchmark}
\label{sec:benchmark}

This section will present a benchmark of typical pre-training and adaptation methods on the large-scale datasets described in Section \ref{sec:dataset}. Since such a benchmark is missing in the literature, we produce the results using the open library \texttt{TLlib} implemented in Section \ref{sec:Library}.

\subsubsection{Pre-Training}

\paragraph{Protocols.}
The transferability of pre-training methods is evaluated on the target task, where the adaptation process and data augmentations are kept the same for fair comparison. Hyper-parameters in adaptation are selected by the performance of target validation data.

\paragraph{Results.}
For the pre-training methods, the transferability cross different tasks and across different domains should be evaluated.
Tables \ref{table:glue_pretrain} and \ref{table:cross_task_pretrain} list the performance on various downstream tasks with different architectures and pre-training tasks. It can be concluded that architectures and pre-training methods have a great impact on the \emph{cross-task} transferability of deep networks.
Table \ref{table:cross_domain_pretrain} lists the performance on ImageNet-Sketch and ImageNet-R  with different architectures and pre-training tasks. Architectures and pre-training strategies also greatly influence the \emph{cross-domain} transferability in deep learning.

\begin{table}[htbp]
	\scriptsize
	\centering
	\caption{
		Cross-task transferability benchmark. Results of different architectures and pre-training methods are reported from the \href{https://gluebenchmark.com}{GLUE leaderboard}. BiLSTM+ELMo \citep{cite:NAACL18ELMo} serves as the baseline.
		GPT \citep{cite:GPT}, $\text{BERT}_\text{Large}$ \citep{cite:NAACL19BERT}, T5 \citep{cite:JMLR20T5}, and ERNIE \citep{sun2019ernie} have different architectures. RoBERTa \citep{roberta}, XLM \citep{cite:Arxiv19XLM}, and SpanBERT \citep{joshi2019spanbert} share the same architecture as $\text{BERT}_\text{Large}$ but employ different pre-training methods.
	}
	\label{table:glue_pretrain}
	\begin{tabular}{lccccccccccc}
		\toprule
		Model                          & CoLA & SST-2 & MRPC & STS-B & QQP  & $\text{MNLI}_{m}$ & $\text{MNLI}_{mm}$ & QNLI & RTE  & Avg  \\
		\midrule
		\begin{tabular}[c]{@{}l@{}}Human \\ Baselines\end{tabular} & 66.4 & 97.8  & 86.3 & 92.7  & 80.4 & 92                & 92.8               & 91.2 & 93.6 & 88.1 \\
		\midrule
		\begin{tabular}[c]{@{}l@{}}BiLSTM\\ +ELMo\\\end{tabular} & 36.0 & 90.4  & 84.9 & 73.3  & 64.8 & 76.4              & 76.1               & 79.9 & 56.8 & 71.0 \\
		GPT                            & 45.4 & 91.3  & 82.3 & 80.0  & 70.3 & 82.1              & 81.4               & 88.1 & 56.0 & 75.2 \\
		\begin{tabular}[c]{@{}l@{}}$\text{BERT}_\text{Large}$\end{tabular} & 60.5 & 94.9  & 89.3 & 86.5  & 72.1 & 86.7              & 85.9               & 92.7 & 70.1 & 82.1 \\
		T5                             & 71.6 & 97.5  & 92.8 & 93.1  & 90.6 & 92.2              & 91.9               & 96.9 & 92.8 & 91.0 \\
		ERNIE                          & 75.5 & 97.8  & 93.9 & 93.0  & 90.9 & 92.3              & 91.7               & 97.3 & 92.6 & 91.7 \\
		\midrule
		RoBERTa                        & 67.8 & 96.7  & 92.3 & 92.2  & 90.2 & 90.8              & 90.2               & 95.4 & 88.2 & 89.3 \\
		XLM                            & 62.9 & 95.6  & 90.7 & 88.8  & 89.8 & 89.1              & 88.5               & 94.0 & 76.0 & 86.2 \\
		SpanBERT                       & 64.3 & 94.8  & 90.9 & 89.9  & 89.5 & 88.1              & 87.7               & 94.3 & 79.0 & 86.5 \\
		\bottomrule
	\end{tabular}
\end{table}

\begin{table}[htbp]
	\scriptsize
	\addtolength{\tabcolsep}{-3.9pt}
	\caption{
		Cross-task transferability benchmark. Results on image recognition using different pre-training methods, including SimCLR \citep{cite::ICML20SimClr} and BYOL \citep{NEURIPS2020_f3ada80d}.
	}
	\label{table:cross_task_pretrain}
	\centering
	\begin{tabular}{llccccccccccc}
		\toprule
		Model                     & Pre-Training                                    & Food & CIFAR10 & CIFAR100 & SUN397 & Cars & Aircraft & DTD  & Pets & Caltech101 & Flowers & Avg  \\ \midrule
		\multirow{3}{*}{ResNet50} & Random Init                                     & 86.9 & 95.9    & 80.2     & 53.6   & 91.4 & 85.9     & 64.8 & 81.5 & 72.6       & 92.0    & 80.5 \\
		                          & SimCLR                                          & 88.2 & 97.7    & 85.9     & 63.5   & 91.3 & 88.1     & 73.2 & 89.2 & 92.1       & 97.0    & 86.6 \\
		                          & BYOL                                            & 88.5 & 97.8    & 86.1     & 63.7   & 91.6 & 88.1     & 76.2 & 91.7 & 93.8       & 97.0    & 87.5 \\ \midrule
		ResNet50                  & \multirow{3}{*}{\begin{tabular}[c]{@{}l@{}}Supervised \\ Pre-Trained \\ on ImageNet\end{tabular}} & 87.8 & 96.8    & 84.5     & 64.7   & 91.7 & 86.6     & 75.2 & 92.5 & 91.8       & 97.5    & 86.9 \\
		ResNet101                 &                                                 & 87.6 & 97.7    & 87.0     & 64.8   & 91.7 & 85.6     & 75.4 & 94.0 & 93.1       & 97.9    & 87.5 \\
		ResNet152                 &                                                 & 87.6 & 97.9    & 87.6     & 66.0   & 92.0 & 85.3     & 74.9 & 94.5 & 93.2       & 97.4    & 87.6 \\ \bottomrule
	\end{tabular}
\end{table}

\begin{table}[htbp]
	\scriptsize
	\caption{
		Cross-domain transferability benchmark. Results are reported from the PyTorch-Image-Models \citep{rw2019timm} on ImageNet using different architectures and pre-training methods. SSP refers to semi-supervised pre-training on YFCC100M \citep{cite:Arxiv19BillionSemiSupervised}. WSP refers to weakly supervised pre-training on IG-1B-Targeted \citep{cite:ECCV18ExploringWeakly}.
	}
	\label{table:cross_domain_pretrain}
	\centering
	\begin{tabular}{llccccc}
		\toprule
		\multicolumn{1}{l}{\multirow{2}{*}{Model}} & \multicolumn{1}{c}{\multirow{2}{*}{Pre-Training}} & \multirow{2}{*}{Param Count} & \multicolumn{2}{c}{ImageNet-Sketch} & \multicolumn{2}{c}{ImageNetR} \\
		\multicolumn{1}{c}{}               &                                                 &                       & top-1 & top-5 & top-1 & top-5 \\ \midrule
		ResNet50                           & \multirow{3}{*}{\begin{tabular}[c]{@{}l@{}}Standard \\ Pre-Trained \\ on ImageNet\end{tabular}} & 25.6                  & 29.6  & 46.8  & 40.4  & 54.7  \\
		ResNet152d                         &                                                 & 60.2                  & 37.9  & 58.4  & 49.3  & 64.4  \\
		$\text{ViT}_\text{large, patch16}$ &                                                 & 304.3                 & 51.8  & 73.7  & 64.3  & 76.2  \\
		\midrule
		\multirow{4}{*}{ResNext101\_32x8d} & Standard                                        & \multirow{4}{*}{88.8} & 29.4  & 48.5  & 42.6  & 58.3  \\
		                                   & SSP                                             &                       & 34.1  & 55.6  & 49.2  & 65.5  \\
		                                   & WSP                                             &                       & 54.9  & 77.5  & 75.9  & 86.2  \\
		                                   & SSP+WSP                                         &                       & 56.4  & 78.9  & 75.6  & 87.1  \\ \bottomrule
	\end{tabular}
\end{table}

\subsubsection{Task Adaptation}

\paragraph{Protocols.}
We follow the common practice in the community as described in \cite{cite:CVPR19DoBetterTransfer}.
Training iterations and data augmentations are kept the same for different task adaptation methods for a fair comparison. Hyper-parameters, such as learning rate and weight decay, of each method are selected by the performance on target validation data.

\paragraph{Results.}
We mainly investigate the \emph{cross-task} transferability between different task adaptation methods. Tables \ref{table:glue_TA} and \ref{table:image_TA} compare the performance of downstream tasks with different task adaptation methods.
Note that previous methods usually only report results on individual datasets, such as Aircraft and Stanford Cars, where regularization tuning performs better than vanilla fine-tuning by a large margin.
But the average improvements brought by different task adaptation methods on a large number of datasets are still limited. Thus, we can conclude that the effectiveness of different task adaptation algorithms largely depends on the relatedness between the target task and the pre-training task.

\begin{table}[htbp]
	\scriptsize
	\addtolength{\tabcolsep}{-3.9pt}
	\centering
	\caption{Cross-task transferability benchmark. GLUE performance with different task adaptation methods, including SMART \citep{SMART}, Adapter-Tuning \citep{houlsby2019parameter} and Diff Pruning \citep{guo-etal-2021-parameter}. Results are reported from their original papers.}
	\label{table:glue_TA}
	\begin{tabular}{llccccccccccc}
		\toprule
		Model                                       & \begin{tabular}[c]{@{}l@{}}Task\\ Adaptation\end{tabular} & \begin{tabular}[c]{@{}l@{}}New Params\\ Per Task\end{tabular} & CoLA & SST-2 & MRPC & STS-B & QQP  & $\text{MNLI}_{m}$ & $\text{MNLI}_{mm}$ & QNLI & RTE  & Avg  \\
		\midrule
		\multirow{2}{*}{RoBERTa}                    & vanilla                        & 100\%                          & 67.8 & 96.7  & 92.3 & 92.2  & 90.2 & 90.8              & 90.2               & 95.4 & 88.2 & 89.3 \\
		                                            & SMART                          & 100\%                          & 65.1 & 97.5  & 93.7 & 92.9  & 90.1 & 91.0              & 90.8               & 95.4 & 87.9 & 89.4 \\
		\midrule
		\multirow{3}{*}{\text{BERT}$_\text{Large}$} & vanilla                        & 100\%                          & 60.5 & 94.9  & 89.3 & 86.5  & 72.1 & 86.7              & 85.9               & 92.7 & 70.1 & 82.1 \\
		                                            & Adapter                        & 2.10\%                         & 59.2 & 94.3  & 88.7 & 87.3  & 89.4 & 85.4              & 85.0               & 92.4 & 71.6 & 83.7 \\
		                                            & Diff Pruning                   & 0.50\%                         & 61.1 & 94.1  & 89.7 & 86.0  & -    & 86.4              & 86.0               & 93.3 & 70.6 & -    \\
		\bottomrule
	\end{tabular}
\end{table}

\begin{table}[htbp]
	\scriptsize
	\centering
	\addtolength{\tabcolsep}{-2pt}
	\caption{Cross-task transferability benchmark. Accuracy (\%) on image classification with different task adaptation methods: LWF \citep{LWF}, DELTA \citep{cite:ICLR19Delta}, BSS \citep{cite:NIPS19BSS}, Bi-Tuning \citep{cite:Arxiv20BiTuning}. Results are reproduced by \texttt{TLlib}.}
	\label{table:image_TA}
	\begin{tabular}{lccccccccccc}
		\toprule
		\begin{tabular}[c]{@{}l@{}}Task\\ Adaptation\end{tabular} & Food & CIFAR10 & CIFAR100 & SUN397 & Cars & Aircraft & DTD  & Pets & Caltech101 & Flowers & Avg  \\ \midrule
		ResNet50                       & 85.1 & 96.9    & 84.1     & 80.7   & 87.8 & 80.1     & 74.4 & 93.2 & 92.9       & 96.5    & 87.2 \\
		LWF                            & 83.9 & 96.5    & 83.6     & 79.5   & 87.4 & 82.2     & 76.3 & 94.0 & 91.7       & 97.1    & 87.2 \\
		DELTA                          & 83.8 & 95.9    & 83.7     & 73.6   & 88.1 & 82.3     & 75.6 & 94.2 & 92.5       & 97.0    & 86.7 \\
		BSS                            & 85.0 & 96.6    & 84.2     & 80.4   & 88.4 & 81.8     & 74.3 & 93.3 & 93.0       & 96.6    & 87.4 \\
		Bi-Tuning                      & 85.7 & 97.1    & 84.3     & 80.7   & 90.3 & 84.8     & 74.6 & 93.5 & 93.4       & 97.5    & 88.2 \\ \bottomrule
	\end{tabular}
\end{table}

\subsubsection{Domain Adaptation}
\label{sec:Benchmark_DA}

\paragraph{Protocols.} We follow the standard protocols for unsupervised domain adaptation \citep{DAN, DANN}.
Training iterations and data augmentations are kept the same for different methods for a fair comparison.
For each method, hyper-parameters are selected on one task and then kept the same for all other tasks, requiring the hyper-parameters of each method to transfer across tasks. This selection strategy is more executable than the importance-weighted cross-validation \citep{IWCV} and can be applied to various practical applications, thus it is widely adopted by many competitions.

\paragraph{Results.}
Tables \ref{Table:da_domainnet} and \ref{table:ImageNet_DA} give the classification performance of different domain adaptation methods on DomainNet and ImageNet. We find that many state-of-the-art methods on small datasets do not perform well on large-scale datasets. This field shall pay more attention to improving the \emph{cross-domain} transferability of deep models on large-scale datasets.

\begin{table}[htbp]
	\scriptsize
	\caption{Cross-domain transferability benchmark. Accuracy (\%) for unsupervised domain adaptation on DomainNet. Results are reproduced from \texttt{TLlib}. }
	\centering
	\label{Table:da_domainnet}
	\begin{tabular}{lccccccccccccc}
		\toprule
		DomainNet                   & c$\shortrightarrow$p & c$\shortrightarrow$r & c$\shortrightarrow$s & p$\shortrightarrow$c & p$\shortrightarrow$r & p$\shortrightarrow$s & r$\shortrightarrow$c & r$\shortrightarrow$p & r$\shortrightarrow$s & s$\shortrightarrow$c & s$\shortrightarrow$p & s$\shortrightarrow$r & Avg  \\ \midrule
		ResNet101                   & 32.7                 & 50.6                 & 39.4                 & 41.1                 & 56.8                 & 35.0                 & 48.6                 & 48.8                 & 36.1                 & 49.0                 & 34.8                 & 46.1                 & 43.3 \\
		DAN~(\citeyear{DAN})        & 38.8                 & 55.2                 & 43.9                 & 45.9                 & 59.0                 & 40.8                 & 50.8                 & 49.8                 & 38.9                 & 56.1                 & 45.9                 & 55.5                 & 48.4 \\
		DANN~(\citeyear{DANN_JMLR}) & 37.9                 & 54.3                 & 44.4                 & 41.7                 & 55.6                 & 36.8                 & 50.7                 & 50.8                 & 40.1                 & 55.0                 & 45.0                 & 54.5                 & 47.2 \\
		ADDA~(\citeyear{ADDA})      & 38.4                 & 54.1                 & 44.1                 & 43.5                 & 56.7                 & 39.2                 & 52.8                 & 51.3                 & 40.9                 & 55.0                 & 45.4                 & 54.5                 & 48.0 \\
		JAN~(\citeyear{JAN})        & 40.5                 & 56.7                 & 45.1                 & 47.2                 & 59.9                 & 43.0                 & 54.2                 & 52.6                 & 41.9                 & 56.6                 & 46.2                 & 55.5                 & 50.0 \\
		CDAN~(\citeyear{CDAN})      & 40.4                 & 56.8                 & 46.1                 & 45.1                 & 58.4                 & 40.5                 & 55.6                 & 53.6                 & 43.0                 & 57.2                 & 46.4                 & 55.7                 & 49.9 \\
		MCD~(\citeyear{MCD})        & 37.5                 & 52.9                 & 44.0                 & 44.6                 & 54.5                 & 41.6                 & 52.0                 & 51.5                 & 39.7                 & 55.5                 & 44.6                 & 52.0                 & 47.5 \\
		MDD~(\citeyear{MDD})        & 42.9                 & 59.5                 & 47.5                 & 48.6                 & 59.4                 & 42.6                 & 58.3                 & 53.7                 & 46.2                 & 58.7                 & 46.5                 & 57.7                 & 51.8 \\
		\bottomrule
	\end{tabular}
	\vspace{-5pt}
\end{table}

\begin{table}[htbp]
	\centering
	\scriptsize
	\caption{Cross-domain transferability benchmark.  Accuracy (\%) for unsupervised domain adaptation on  ImageNet-scale datasets. Results are reproduced from \texttt{TLlib}.}
	\label{table:ImageNet_DA}
	\begin{tabular}{lcc}
		\toprule
		Task                   & ImageNet$\shortrightarrow$ImageNet-R & ImageNet$\shortrightarrow$ImageNet-Sketch \\
		Model                  & ResNet50                             & ig\_resnext101\_32x8d                     \\
		\midrule
		Source Only            & 35.6                                 & 54.9                                      \\
		DAN~\citep{DAN}        & 39.8                                 & 55.7                                      \\
		DANN~\citep{DANN_JMLR} & 52.7                                 & 56.5                                      \\
		JAN~\citep{JAN}        & 41.7                                 & 55.7                                      \\
		CDAN~\citep{CDAN}      & 53.9                                 & 58.2                                      \\
		MCD~\citep{MCD}        & 46.7                                 & 55.0                                      \\
		MDD~\citep{MDD}        & 56.2                                 & 62.4                                      \\
		\bottomrule
	\end{tabular}
	\vspace{-5pt}
\end{table}

\section{Conclusion}

In this paper, we investigate how to acquire and apply transferability in the whole lifecycle of deep learning.
In the pre-training section, we focus on how to improve the transferability of the pre-trained models by designing architecture, pre-training task, and training strategy.
In the task adaptation section, we discuss how to better preserve and utilize the transferable knowledge to improve the performance of target tasks.
In the domain adaptation section, we illustrate how to bridge the domain gap to increase the transferability for real applications.
This survey connects many isolated areas with their relation to transferability and provides a unified perspective to explore transferability in deep learning. We expect this study will attract the community's attention to the fundamental role of transferability in deep learning.

\vspace{-10pt}
\acks{This work was supported by the NSFC Grants (62022050 and 62021002), the Beijing Nova Program (Z201100006820041), and the BNRist Innovation Fund (BNR2021RC01002).
}

\newpage
\bibliography{sample}

\begin{thebibliography}{212}
\providecommand{\natexlab}[1]{#1}
\providecommand{\url}[1]{\texttt{#1}}
\expandafter\ifx\csname urlstyle\endcsname\relax
  \providecommand{\doi}[1]{doi: #1}\else
  \providecommand{\doi}{doi: \begingroup \urlstyle{rm}\Url}\fi

\bibitem[Abnar et~al.(2022)Abnar, Dehghani, Neyshabur, and
  Sedghi]{abnar2021exploring}
Samira Abnar, Mostafa Dehghani, Behnam Neyshabur, and Hanie Sedghi.
\newblock Exploring the limits of large scale pre-training.
\newblock In \emph{ICLR}, 2022.

\bibitem[Aghajanyan et~al.(2021)Aghajanyan, Zettlemoyer, and
  Gupta]{aghajanyan2020intrinsic}
Armen Aghajanyan, Luke Zettlemoyer, and Sonal Gupta.
\newblock Intrinsic dimensionality explains the effectiveness of language model
  fine-tuning.
\newblock In \emph{ACL}, 2021.

\bibitem[Amodei et~al.(2016)Amodei, Ananthanarayanan, Anubhai, Bai, Battenberg,
  Case, Casper, Catanzaro, Cheng, Chen, et~al.]{cite:ICML16DeepSpeech2}
Dario Amodei, Sundaram Ananthanarayanan, Rishita Anubhai, Jingliang Bai, Eric
  Battenberg, Carl Case, Jared Casper, Bryan Catanzaro, Qiang Cheng, Guoliang
  Chen, et~al.
\newblock Deep speech 2: End-to-end speech recognition in english and mandarin.
\newblock In \emph{ICML}, 2016.

\bibitem[Arjovsky et~al.(2019)Arjovsky, Bottou, Gulrajani, and
  Lopez-Paz]{cite:Arxiv19IRM}
Martin Arjovsky, L{\'e}on Bottou, Ishaan Gulrajani, and David Lopez-Paz.
\newblock Invariant risk minimization.
\newblock \emph{arXiv preprint arXiv:1907.02893}, 2019.

\bibitem[Arora et~al.(2017)Arora, Ge, Liang, Ma, and Zhang]{EquilibriumOfGAN}
Sanjeev Arora, Rong Ge, Yingyu Liang, Tengyu Ma, and Yi~Zhang.
\newblock Generalization and equilibrium in generative adversarial nets
  ({GAN}s).
\newblock In \emph{ICML}, 2017.

\bibitem[Bartlett and Mendelson(2002)]{RademacherComplexity}
Peter~L. Bartlett and Shahar Mendelson.
\newblock Rademacher and gaussian complexities: Risk bounds and structural
  results.
\newblock In \emph{JMLR}, 2002.

\bibitem[Beltagy et~al.(2019)Beltagy, Lo, and Cohan]{Beltagy2019SciBERT}
Iz~Beltagy, Kyle Lo, and Arman Cohan.
\newblock Scibert: Pretrained language model for scientific text.
\newblock In \emph{EMNLP}, 2019.

\bibitem[Ben-David et~al.(2010{\natexlab{a}})Ben-David, Blitzer, Crammer,
  Kulesza, Pereira, and Vaughan]{DATheroy10}
S.~Ben-David, J.~Blitzer, K.~Crammer, A.~Kulesza, F.~Pereira, and J.~W.
  Vaughan.
\newblock A theory of learning from different domains.
\newblock \emph{Machine Learning, 79}, page 151–175, 2010{\natexlab{a}}.

\bibitem[Ben-David et~al.(2006)Ben-David, Blitzer, Crammer, and
  Pereira]{DATheroy07}
Shai Ben-David, John Blitzer, Koby Crammer, and Fernando Pereira.
\newblock Analysis of representations for domain adaptation.
\newblock In \emph{NeurIPS}, 2006.

\bibitem[Ben-David et~al.(2010{\natexlab{b}})Ben-David, Lu, Luu, and
  Pal]{pmlr-v9-david10a}
Shai Ben-David, Tyler Lu, Teresa Luu, and David Pal.
\newblock Impossibility theorems for domain adaptation.
\newblock In \emph{AISTATS}, pages 129--136, 2010{\natexlab{b}}.

\bibitem[Bengio(2012)]{cite:BengioUnsupervised}
Yoshua Bengio.
\newblock Deep learning of representations for unsupervised and transfer
  learning.
\newblock In \emph{ICML workshop}, 2012.

\bibitem[Bengio et~al.(2007)Bengio, Lamblin, Popovici, and
  Larochelle]{cite:NIPS07GreedyLayerWiseTraining}
Yoshua Bengio, Pascal Lamblin, Dan Popovici, and Hugo Larochelle.
\newblock Greedy layer-wise training of deep networks.
\newblock In \emph{NeurIPS}, 2007.

\bibitem[Bengio et~al.(2013)Bengio, Courville, and
  Vincent]{RepresentationLearningReview}
Yoshua Bengio, Aaron Courville, and Pascal Vincent.
\newblock Representation learning: A review and new perspectives.
\newblock \emph{TPAMI}, 35\penalty0 (8):\penalty0 1798--1828, 2013.

\bibitem[Bengio et~al.(2021)Bengio, Lecun, and
  Hinton]{cite:Turing21DeepLearningForAI}
Yoshua Bengio, Yann Lecun, and Geoffrey Hinton.
\newblock Deep learning for ai.
\newblock \emph{Communications of the ACM}, 64\penalty0 (7):\penalty0 58--65,
  2021.

\bibitem[Bousmalis et~al.(2016)Bousmalis, Trigeorgis, Silberman, Krishnan, and
  Erhan]{DSN}
Konstantinos Bousmalis, George Trigeorgis, Nathan Silberman, Dilip Krishnan,
  and Dumitru Erhan.
\newblock Domain separation networks.
\newblock In \emph{NeurIPS}, 2016.

\bibitem[Bousmalis et~al.(2017)Bousmalis, Silberman, Dohan, Erhan, and
  Krishnan]{PixelDA}
Konstantinos Bousmalis, Nathan Silberman, David Dohan, Dumitru Erhan, and Dilip
  Krishnan.
\newblock Unsupervised pixel-level domain adaptation with generative
  adversarial networks.
\newblock In \emph{CVPR}, 2017.

\bibitem[Bousmalis et~al.(2018)Bousmalis, Irpan, Wohlhart, Bai, Kelcey,
  Kalakrishnan, Downs, Ibarz, Pastor, Konolige, Levine, and
  Vanhoucke]{DARoboticGrasping}
Konstantinos Bousmalis, Alex Irpan, Paul Wohlhart, Yunfei Bai, Matthew Kelcey,
  Mrinal Kalakrishnan, Laura Downs, Julian Ibarz, Peter Pastor, Kurt Konolige,
  Sergey Levine, and Vincent Vanhoucke.
\newblock Using simulation and domain adaptation to improve efficiency of deep
  robotic grasping.
\newblock In \emph{ICRA}, 2018.

\bibitem[Brown et~al.(2020)Brown, Mann, Ryder, Subbiah, Kaplan, Dhariwal,
  Neelakantan, Shyam, Sastry, Askell, et~al.]{cite:NIPS20GPT3}
Tom~B Brown, Benjamin Mann, Nick Ryder, Melanie Subbiah, Jared Kaplan, Prafulla
  Dhariwal, Arvind Neelakantan, Pranav Shyam, Girish Sastry, Amanda Askell,
  et~al.
\newblock Language models are few-shot learners.
\newblock In \emph{NeurIPS}, 2020.

\bibitem[Busto and Gall(2017)]{OpensetDA}
Pau~Panareda Busto and Juergen Gall.
\newblock Open set domain adaptation.
\newblock In \emph{ICCV}, 2017.

\bibitem[Caruana(1997)]{Caruana97multitasklearning}
Rich Caruana.
\newblock Multitask learning.
\newblock Technical report, 1997.

\bibitem[Chapelle et~al.(2006)Chapelle, Sch\"{o}lkopf, and Zien]{SSL}
Olivier Chapelle, Bernhard Sch\"{o}lkopf, and Alexander Zien.
\newblock \emph{Semi-Supervised Learning (Adaptive Computation and Machine
  Learning)}.
\newblock The MIT Press, 2006.
\newblock ISBN 0262033585.

\bibitem[Chen et~al.(2012)Chen, Xu, Weinberger, and
  Sha]{MarginalizedDenoisingAutoencoders}
Minmin Chen, Zhixiang Xu, Kilian~Q. Weinberger, and Fei Sha.
\newblock Marginalized denoising autoencoders for domain adaptation.
\newblock In \emph{ICML}, 2012.

\bibitem[Chen et~al.(2020)Chen, Kornblith, Norouzi, and
  Hinton]{cite::ICML20SimClr}
Ting Chen, Simon Kornblith, Mohammad Norouzi, and Geoffrey Hinton.
\newblock A simple framework for contrastive learning of visual
  representations.
\newblock In \emph{ICML}, 2020.

\bibitem[Chen et~al.(2019{\natexlab{a}})Chen, Liu, Kira, Wang, and
  Huang]{cite:ICLR19ACloserLook}
Wei-Yu Chen, Yen-Cheng Liu, Zsolt Kira, Yu-Chiang~Frank Wang, and Jia-Bin
  Huang.
\newblock A closer look at few-shot classification.
\newblock In \emph{ICLR}, 2019{\natexlab{a}}.

\bibitem[Chen and He(2021)]{cite:CVPR21SiameseRepresentation}
Xinlei Chen and Kaiming He.
\newblock Exploring simple siamese representation learning.
\newblock In \emph{CVPR}, 2021.

\bibitem[Chen et~al.(2021{\natexlab{a}})Chen, Xie, and He]{chen2021mocov3}
Xinlei Chen, Saining Xie, and Kaiming He.
\newblock An empirical study of training self-supervised vision transformers.
\newblock \emph{arXiv preprint arXiv:2104.02057}, 2021{\natexlab{a}}.

\bibitem[Chen et~al.(2019{\natexlab{b}})Chen, Wang, Fu, Long, and
  Wang]{cite:NIPS19BSS}
Xinyang Chen, Sinan Wang, Bo~Fu, Mingsheng Long, and Jianmin Wang.
\newblock Catastrophic forgetting meets negative transfer: Batch spectral
  shrinkage for safe transfer learning.
\newblock In \emph{NeurIPS}, 2019{\natexlab{b}}.

\bibitem[Chen et~al.(2019{\natexlab{c}})Chen, Wang, Long, and Wang]{bsp}
Xinyang Chen, Sinan Wang, Mingsheng Long, and Jianmin Wang.
\newblock Transferability vs. discriminability: Batch spectral penalization for
  adversarial domain adaptation.
\newblock In \emph{ICML}, 2019{\natexlab{c}}.

\bibitem[Chen et~al.(2021{\natexlab{b}})Chen, Wang, Wang, and
  Long]{DAR_ICML_21}
Xinyang Chen, Sinan Wang, Jianmin Wang, and Mingsheng Long.
\newblock Representation subspace distance for domain adaptation regression.
\newblock In \emph{ICML}, 2021{\natexlab{b}}.

\bibitem[Chen et~al.(2018)Chen, Li, Sakaridis, Dai, and Gool]{DAFaster}
Yuhua Chen, Wen Li, Christos Sakaridis, Dengxin Dai, and Luc~Van Gool.
\newblock Domain adaptive faster {R-CNN} for object detection in the wild.
\newblock In \emph{CVPR}, 2018.

\bibitem[Cho et~al.(2014)Cho, van Merri{\"e}nboer, Gulcehre, Bahdanau,
  Bougares, Schwenk, and Bengio]{cite:Arxiv14GRU}
Kyunghyun Cho, Bart van Merri{\"e}nboer, Caglar Gulcehre, Dzmitry Bahdanau,
  Fethi Bougares, Holger Schwenk, and Yoshua Bengio.
\newblock Learning phrase representations using {RNN} encoder{--}decoder for
  statistical machine translation.
\newblock In \emph{EMNLP}, 2014.

\bibitem[Chronopoulou et~al.(2019)Chronopoulou, Baziotis, and
  Potamianos]{chronopoulou-etal-2019-embarrassingly}
Alexandra Chronopoulou, Christos Baziotis, and Alexandros Potamianos.
\newblock An embarrassingly simple approach for transfer learning from
  pretrained language models.
\newblock In \emph{NAACL}, 2019.

\bibitem[Conneau et~al.(2018)Conneau, Lample, Rinott, Williams, Bowman,
  Schwenk, and Stoyanov]{XNLI}
Alexis Conneau, Guillaume Lample, Ruty Rinott, Adina Williams, Samuel~R.
  Bowman, Holger Schwenk, and Veselin Stoyanov.
\newblock {XNLI:} evaluating cross-lingual sentence representations.
\newblock In \emph{EMNLP}, 2018.

\bibitem[Courty et~al.(2017)Courty, Flamary, Habrard, and Rakotomamonjy]{jdot}
Nicolas Courty, Rémi Flamary, Amaury Habrard, and Alain Rakotomamonjy.
\newblock Joint distribution optimal transportation for domain adaptation.
\newblock In \emph{NeurIPS}, 2017.

\bibitem[Cui et~al.(2018)Cui, Song, Sun, Howard, and
  Belongie]{cite:CVPR18LargeScaleFinegrained}
Yin Cui, Yang Song, Chen Sun, Andrew Howard, and Serge Belongie.
\newblock Large scale fine-grained categorization and domain-specific transfer
  learning.
\newblock In \emph{CVPR}, pages 4109--4118, 2018.

\bibitem[Damodaran et~al.(2018)Damodaran, Kellenberger, Flamary, Tuia, and
  Courty]{deepjdot}
Bharath~Bhushan Damodaran, Benjamin Kellenberger, R{\'{e}}mi Flamary, Devis
  Tuia, and Nicolas Courty.
\newblock Deepjdot: Deep joint distribution optimal transport for unsupervised
  domain adaptation.
\newblock In \emph{ECCV}, 2018.

\bibitem[Delange et~al.(2021)Delange, Aljundi, Masana, Parisot, Jia, Leonardis,
  Slabaugh, and Tuytelaars]{Delange21}
Matthias Delange, Rahaf Aljundi, Marc Masana, Sarah Parisot, Xu~Jia, Ales
  Leonardis, Greg Slabaugh, and Tinne Tuytelaars.
\newblock A continual learning survey: Defying forgetting in classification
  tasks.
\newblock \emph{TPAMI}, page 1–20, 2021.

\bibitem[Deng et~al.(2009)Deng, Dong, Socher, Li, Li, and
  Fei-Fei]{deng_imagenet:_2009}
Jia Deng, Wei Dong, Richard Socher, Li-Jia Li, Kai Li, and Li~Fei-Fei.
\newblock Imagenet: {A} large-scale hierarchical image database.
\newblock In \emph{{CVPR}}, 2009.

\bibitem[Devlin et~al.(2019)Devlin, Chang, Lee, and
  Toutanova]{cite:NAACL19BERT}
Jacob Devlin, Ming-Wei Chang, Kenton Lee, and Kristina Toutanova.
\newblock Bert: Pre-training of deep bidirectional transformers for language
  understanding.
\newblock In \emph{NAACL}, 2019.

\bibitem[Doersch et~al.(2015)Doersch, Gupta, and
  Efros]{doersch2015unsupervised}
Carl Doersch, Abhinav Gupta, and Alexei~A. Efros.
\newblock Unsupervised visual representation learning by context prediction.
\newblock In \emph{ICCV}, 2015.

\bibitem[Donahue et~al.(2014)Donahue, Jia, Vinyals, Hoffman, Zhang, Tzeng, and
  Darrell]{cite:ICML14Decaf}
Jeff Donahue, Yangqing Jia, Oriol Vinyals, Judy Hoffman, Ning Zhang, Eric
  Tzeng, and Trevor Darrell.
\newblock Decaf: A deep convolutional activation feature for generic visual
  recognition.
\newblock In \emph{ICML}, 2014.

\bibitem[Dosovitskiy et~al.(2021)Dosovitskiy, Beyer, Kolesnikov, Weissenborn,
  Zhai, Unterthiner, Dehghani, Minderer, Heigold, Gelly, Uszkoreit, and
  Houlsby]{cite:ICLR21VIT}
Alexey Dosovitskiy, Lucas Beyer, Alexander Kolesnikov, Dirk Weissenborn,
  Xiaohua Zhai, Thomas Unterthiner, Mostafa Dehghani, Matthias Minderer, Georg
  Heigold, Sylvain Gelly, Jakob Uszkoreit, and Neil Houlsby.
\newblock An {Image} is {Worth} 16x16 {Words}: {Transformers} for {Image}
  {Recognition} at {Scale}.
\newblock In \emph{ICLR}, 2021.

\bibitem[Finn et~al.(2017)Finn, Abbeel, and Levine]{cite:ICML17MAML}
Chelsea Finn, Pieter Abbeel, and Sergey Levine.
\newblock Model-agnostic meta-learning for fast adaptation of deep networks.
\newblock In \emph{ICML}, 2017.

\bibitem[French et~al.(2018)French, Mackiewicz, and Fisher]{SelfEnsemble}
Geoffrey French, Michal Mackiewicz, and Mark~H. Fisher.
\newblock Self-ensembling for domain adaptation.
\newblock In \emph{ICLR}, 2018.

\bibitem[Ganin and Lempitsky(2015)]{DANN}
Yaroslav Ganin and Victor Lempitsky.
\newblock Unsupervised domain adaptation by backpropagation.
\newblock In \emph{ICML}, 2015.

\bibitem[Ganin et~al.(2016)Ganin, Ustinova, Ajakan, Germain, Larochelle,
  Laviolette, March, and Lempitsky]{DANN_JMLR}
Yaroslav Ganin, Evgeniya Ustinova, Hana Ajakan, Pascal Germain, Hugo
  Larochelle, Fran{\c{c}}ois Laviolette, Mario March, and Victor Lempitsky.
\newblock Domain-adversarial training of neural networks.
\newblock \emph{JMLR}, 17\penalty0 (59):\penalty0 1--35, 2016.

\bibitem[Garcia and Bruna(2018)]{cite:ICLR18GNN}
Victor Garcia and Joan Bruna.
\newblock Few-shot learning with graph neural networks.
\newblock In \emph{ICLR}, 2018.

\bibitem[Ge et~al.(2020)Ge, Chen, and Li]{ge2020mutual}
Yixiao Ge, Dapeng Chen, and Hongsheng Li.
\newblock Mutual mean-teaching: Pseudo label refinery for unsupervised domain
  adaptation on person re-identification.
\newblock In \emph{ICLR}, 2020.

\bibitem[Geirhos et~al.(2019)Geirhos, Rubisch, Michaelis, Bethge, Wichmann, and
  Brendel]{cite:ICLR19ImageNetBias}
Robert Geirhos, Patricia Rubisch, Claudio Michaelis, Matthias Bethge, Felix~A
  Wichmann, and Wieland Brendel.
\newblock Imagenet-trained cnns are biased towards texture; increasing shape
  bias improves accuracy and robustness.
\newblock In \emph{ICLR}, 2019.

\bibitem[Girshick et~al.(2014)Girshick, Donahue, Darrell, and
  Malik]{cite:CVPR14RichFeature}
Ross Girshick, Jeff Donahue, Trevor Darrell, and Jitendra Malik.
\newblock Rich feature hierarchies for accurate object detection and semantic
  segmentation.
\newblock In \emph{CVPR}, 2014.

\bibitem[Glorot et~al.(2011)Glorot, Bordes, and
  Bengio]{DAForSentimentClassification}
Xavier Glorot, Antoine Bordes, and Yoshua Bengio.
\newblock Domain adaptation for large-scale sentiment classification: A deep
  learning approach.
\newblock In \emph{ICML}, 2011.

\bibitem[Gong et~al.(2012)Gong, Shi, Sha, and Grauman]{GFK}
Boqing Gong, Yuan Shi, Fei Sha, and Kristen Grauman.
\newblock Geodesic flow kernel for unsupervised domain adaptation.
\newblock In \emph{CVPR}, 2012.

\bibitem[Gong et~al.(2013)Gong, Grauman, and Sha]{cite:DAtransform1}
Boqing Gong, Kristen Grauman, and Fei Sha.
\newblock Connecting the dots with landmarks: Discriminatively learning
  domain-invariant features for unsupervised domain adaptation.
\newblock In \emph{ICML}, 2013.

\bibitem[Goodfellow et~al.(2014)Goodfellow, Pouget-Abadie, Mirza, Xu,
  Warde-Farley, Ozair, Courville, and Bengio]{GAN}
Ian~J Goodfellow, Jean Pouget-Abadie, Mehdi Mirza, Bing Xu, David Warde-Farley,
  Sherjil Ozair, Aaron Courville, and Yoshua Bengio.
\newblock Generative adversarial networks.
\newblock In \emph{NeurIPS}, 2014.

\bibitem[Goodfellow et~al.(2015)Goodfellow, Shlens, and
  Szegedy]{cite:ICLR15AdversarialExample}
Ian~J Goodfellow, Jonathon Shlens, and Christian Szegedy.
\newblock Explaining and harnessing adversarial examples.
\newblock 2015.

\bibitem[Goyal et~al.(2021)Goyal, Lamb, Hoffmann, Sodhani, Levine, Bengio, and
  Sch{\"o}lkopf]{cite:ICLR21RIM}
Anirudh Goyal, Alex Lamb, Jordan Hoffmann, Shagun Sodhani, Sergey Levine,
  Yoshua Bengio, and Bernhard Sch{\"o}lkopf.
\newblock Recurrent independent mechanisms.
\newblock In \emph{ICLR}, 2021.

\bibitem[Gretton et~al.(2012{\natexlab{a}})Gretton, Borgwardt, Rasch,
  Sch{{\"o}}lkopf, and Smola]{AKernelTwo-SampleTest}
Arthur Gretton, Karsten~M. Borgwardt, Malte~J. Rasch, Bernhard Sch{{\"o}}lkopf,
  and Alexander Smola.
\newblock A kernel two-sample test.
\newblock \emph{JMLR}, 13\penalty0 (25):\penalty0 723--773, 2012{\natexlab{a}}.

\bibitem[Gretton et~al.(2012{\natexlab{b}})Gretton, Sejdinovic, Strathmann,
  Balakrishnan, Pontil, Fukumizu, and Sriperumbudur]{MMD}
Arthur Gretton, Dino Sejdinovic, Heiko Strathmann, Sivaraman Balakrishnan,
  Massimiliano Pontil, Kenji Fukumizu, and Bharath~K Sriperumbudur.
\newblock Optimal kernel choice for large-scale two-sample tests.
\newblock In \emph{NeurIPS}, 2012{\natexlab{b}}.

\bibitem[Grill et~al.(2020)Grill, Strub, Altch\'{e}, Tallec, Richemond,
  Buchatskaya, Doersch, Avila~Pires, Guo, Gheshlaghi~Azar, Piot, kavukcuoglu,
  Munos, and Valko]{NEURIPS2020_f3ada80d}
Jean-Bastien Grill, Florian Strub, Florent Altch\'{e}, Corentin Tallec, Pierre
  Richemond, Elena Buchatskaya, Carl Doersch, Bernardo Avila~Pires, Zhaohan
  Guo, Mohammad Gheshlaghi~Azar, Bilal Piot, koray kavukcuoglu, Remi Munos, and
  Michal Valko.
\newblock Bootstrap your own latent - a new approach to self-supervised
  learning.
\newblock In \emph{NeurIPS}, 2020.

\bibitem[Gulrajani and Lopez{-}Paz(2021)]{DomainBed}
Ishaan Gulrajani and David Lopez{-}Paz.
\newblock In search of lost domain generalization.
\newblock In \emph{ICLR}, 2021.

\bibitem[Guo et~al.(2021)Guo, Rush, and Kim]{guo-etal-2021-parameter}
Demi Guo, Alexander Rush, and Yoon Kim.
\newblock Parameter-efficient transfer learning with diff pruning.
\newblock In \emph{ACL}, 2021.

\bibitem[Guo et~al.(2019)Guo, Shi, Kumar, Grauman, Rosing, and
  Feris]{cite:CVPR19SpotTune}
Yunhui Guo, Honghui Shi, Abhishek Kumar, Kristen Grauman, Tajana Rosing, and
  Rogerio Feris.
\newblock Spottune: transfer learning through adaptive fine-tuning.
\newblock In \emph{CVPR}, 2019.

\bibitem[Gururangan et~al.(2020)Gururangan, Marasović, Swayamdipta, Lo,
  Beltagy, Downey, and Smith]{dontstoppretraining2020}
Suchin Gururangan, Ana Marasović, Swabha Swayamdipta, Kyle Lo, Iz~Beltagy,
  Doug Downey, and Noah~A. Smith.
\newblock Don't stop pretraining: Adapt language models to domains and tasks.
\newblock In \emph{ACL}, 2020.

\bibitem[He et~al.(2016)He, Zhang, Ren, and Sun]{cite:CVPR16ResNet}
Kaiming He, Xiangyu Zhang, Shaoqing Ren, and Jian Sun.
\newblock Deep residual learning for image recognition.
\newblock In \emph{CVPR}, 2016.

\bibitem[He et~al.(2017)He, Gkioxari, Dollár, and Girshick]{MaskRCNN}
Kaiming He, Georgia Gkioxari, Piotr Dollár, and Ross Girshick.
\newblock Mask r-cnn.
\newblock In \emph{ICCV}, 2017.

\bibitem[He et~al.(2019)He, Girshick, and
  Doll{\'a}r]{cite:ICCV19RethinkingPretraining}
Kaiming He, Ross Girshick, and Piotr Doll{\'a}r.
\newblock Rethinking imagenet pre-training.
\newblock In \emph{ICCV}, 2019.

\bibitem[He et~al.(2020)He, Fan, Wu, Xie, and Girshick]{cite:CVPR20MoCo}
Kaiming He, Haoqi Fan, Yuxin Wu, Saining Xie, and Ross Girshick.
\newblock Momentum contrast for unsupervised visual representation learning.
\newblock In \emph{CVPR}, 2020.

\bibitem[He et~al.(2021)He, Chen, Xie, Li, Doll{\'a}r, and
  Girshick]{cite:Arxiv21MAE}
Kaiming He, Xinlei Chen, Saining Xie, Yanghao Li, Piotr Doll{\'a}r, and Ross
  Girshick.
\newblock Masked autoencoders are scalable vision learners.
\newblock \emph{arXiv preprint arXiv:2111.06377}, 2021.

\bibitem[Hendrycks et~al.(2021)Hendrycks, Basart, Mu, Kadavath, Wang, Dorundo,
  Desai, Zhu, Parajuli, Guo, Song, Steinhardt, and Gilmer]{ImageNetR}
Dan Hendrycks, Steven Basart, Norman Mu, Saurav Kadavath, Frank Wang, Evan
  Dorundo, Rahul Desai, Tyler Zhu, Samyak Parajuli, Mike Guo, Dawn Song, Jacob
  Steinhardt, and Justin Gilmer.
\newblock The many faces of robustness: A critical analysis of
  out-of-distribution generalization.
\newblock \emph{ICCV}, 2021.

\bibitem[Hjelm et~al.(2019)Hjelm, Fedorov, Lavoie-Marchildon, Grewal, Bachman,
  Trischler, and Bengio]{cite:ICLR19DeepInfoMax}
R~Devon Hjelm, Alex Fedorov, Samuel Lavoie-Marchildon, Karan Grewal, Phil
  Bachman, Adam Trischler, and Yoshua Bengio.
\newblock Learning deep representations by mutual information estimation and
  maximization.
\newblock In \emph{ICLR}, 2019.

\bibitem[Hoffman et~al.(2016)Hoffman, Wang, Yu, and Darrell]{FCN_in_wild}
Judy Hoffman, Dequan Wang, Fisher Yu, and Trevor Darrell.
\newblock Fcns in the wild: Pixel-level adversarial and constraint-based
  adaptation.
\newblock 2016.

\bibitem[Hoffman et~al.(2018)Hoffman, Tzeng, Park, Zhu, Isola, Saenko, Efros,
  and Darrell]{cycada}
Judy Hoffman, Eric Tzeng, Taesung Park, Jun-Yan Zhu, Phillip Isola, Kate
  Saenko, Alexei Efros, and Trevor Darrell.
\newblock Cycada: Cycle-consistent adversarial domain adaptation.
\newblock In \emph{ICML}, 2018.

\bibitem[Houlsby et~al.(2019)Houlsby, Giurgiu, Jastrzebski, Morrone,
  De~Laroussilhe, Gesmundo, Attariyan, and Gelly]{houlsby2019parameter}
Neil Houlsby, Andrei Giurgiu, Stanislaw Jastrzebski, Bruna Morrone, Quentin
  De~Laroussilhe, Andrea Gesmundo, Mona Attariyan, and Sylvain Gelly.
\newblock Parameter-efficient transfer learning for {NLP}.
\newblock In \emph{ICML}, 2019.

\bibitem[Howard and Ruder(2018)]{ULMFiT}
Jeremy Howard and Sebastian Ruder.
\newblock Universal language model fine-tuning for text classification.
\newblock In \emph{ACL}, 2018.

\bibitem[Hu et~al.(2020)Hu, Liu, Gomes, Zitnik, Liang, Pande, and
  Leskovec]{graph_pretrain}
Weihua Hu, Bowen Liu, Joseph Gomes, Marinka Zitnik, Percy Liang, Vijay~S.
  Pande, and Jure Leskovec.
\newblock Pre-training graph neural networks.
\newblock In \emph{ICLR}, 2020.

\bibitem[Huang et~al.(2007)Huang, Gretton, Borgwardt, Sch\"{o}lkopf, and
  Smola]{NIPS2006_a2186aa7}
Jiayuan Huang, Arthur Gretton, Karsten Borgwardt, Bernhard Sch\"{o}lkopf, and
  Alex Smola.
\newblock Correcting sample selection bias by unlabeled data.
\newblock In \emph{NeurIPS}, 2007.

\bibitem[Ioffe and Szegedy(2015)]{BatchNorm}
Sergey Ioffe and Christian Szegedy.
\newblock Batch normalization: Accelerating deep network training by reducing
  internal covariate shift.
\newblock In \emph{ICML}, 2015.

\bibitem[Jang et~al.(2019)Jang, Lee, Hwang, and
  Shin]{cite:ICML19LearningtoTransfer}
Yunhun Jang, Hankook Lee, Sung~Ju Hwang, and Jinwoo Shin.
\newblock Learning what and where to transfer.
\newblock In \emph{ICML}, 2019.

\bibitem[Jiang et~al.(2020)Jiang, He, Chen, Liu, Gao, and Zhao]{SMART}
Haoming Jiang, Pengcheng He, Weizhu Chen, Xiaodong Liu, Jianfeng Gao, and Tuo
  Zhao.
\newblock {SMART:} robust and efficient fine-tuning for pre-trained natural
  language models through principled regularized optimization.
\newblock In \emph{ACL}, 2020.

\bibitem[Jiang et~al.(2021)Jiang, Ji, Wang, Liu, Wang, and Long]{RegDA}
Junguang Jiang, Yifei Ji, Ximei Wang, Yufeng Liu, Jianmin Wang, and Mingsheng
  Long.
\newblock Regressive domain adaptation for unsupervised keypoint detection.
\newblock In \emph{CVPR}, 2021.

\bibitem[Jiang et~al.(2022)Jiang, Chen, Wang, and Long]{jiang2021decoupled}
Junguang Jiang, Baixu Chen, Jianmin Wang, and Mingsheng Long.
\newblock Decoupled adaptation for cross-domain object detection.
\newblock In \emph{ICLR}, 2022.

\bibitem[Jin et~al.(2020)Jin, Wang, Long, and Wang]{MCC}
Ying Jin, Ximei Wang, Mingsheng Long, and Jianmin Wang.
\newblock Minimum class confusion for versatile domain adaptation.
\newblock In \emph{ECCV}, 2020.

\bibitem[Joshi et~al.(2020)Joshi, Chen, Liu, Weld, Zettlemoyer, and
  Levy]{joshi2019spanbert}
Mandar Joshi, Danqi Chen, Yinhan Liu, Daniel~S. Weld, Luke Zettlemoyer, and
  Omer Levy.
\newblock {SpanBERT}: Improving pre-training by representing and predicting
  spans.
\newblock In \emph{TACL}, 2020.

\bibitem[Kang et~al.(2019)Kang, Jiang, Yang, and Hauptmann]{CAN}
Guoliang Kang, Lu~Jiang, Yi~Yang, and Alexander~G Hauptmann.
\newblock Contrastive adaptation network for unsupervised domain adaptation.
\newblock In \emph{CVPR}, 2019.

\bibitem[Kim et~al.(2019)Kim, Jeong, Kim, Choi, and Kim]{diversify}
Taekyung Kim, Minki Jeong, Seunghyeon Kim, Seokeon Choi, and Changick Kim.
\newblock Diversify and match: A domain adaptive representation learning
  paradigm for object detection.
\newblock In \emph{CVPR}, 2019.

\bibitem[Kirkpatrick et~al.(2017)Kirkpatrick, Pascanu, Rabinowitz, Veness,
  Desjardins, Rusu, Milan, Quan, Ramalho, Grabska-Barwinska, et~al.]{EWC}
James Kirkpatrick, Razvan Pascanu, Neil Rabinowitz, Joel Veness, Guillaume
  Desjardins, Andrei~A Rusu, Kieran Milan, John Quan, Tiago Ramalho, Agnieszka
  Grabska-Barwinska, et~al.
\newblock Overcoming catastrophic forgetting in neural networks.
\newblock \emph{PNAS}, 114\penalty0 (13):\penalty0 3521--3526, 2017.

\bibitem[Kolesnikov et~al.(2020)Kolesnikov, Beyer, Zhai, Puigcerver, Yung,
  Gelly, and Houlsby]{cite:ECCV20BigTransfer}
Alexander Kolesnikov, Lucas Beyer, Xiaohua Zhai, Joan Puigcerver, Jessica Yung,
  Sylvain Gelly, and Neil Houlsby.
\newblock Big transfer (bit): General visual representation learning.
\newblock In \emph{ECCV}, 2020.

\bibitem[Kornblith et~al.(2019)Kornblith, Shlens, and
  Le]{cite:CVPR19DoBetterTransfer}
Simon Kornblith, Jonathon Shlens, and Quoc~V Le.
\newblock Do better imagenet models transfer better?
\newblock In \emph{CVPR}, 2019.

\bibitem[Kou et~al.(2020)Kou, You, Long, and Wang]{cite:NIPS20StocNorm}
Zhi Kou, Kaichao You, Mingsheng Long, and Jianmin Wang.
\newblock Stochastic normalization.
\newblock In \emph{NeurIPS}, 2020.

\bibitem[Krizhevsky et~al.(2012)Krizhevsky, Sutskever, and
  Hinton]{cite:NIPS12AlexNet}
Alex Krizhevsky, Ilya Sutskever, and Geoffrey~E. Hinton.
\newblock Imagenet classification with deep convolutional neural networks.
\newblock In \emph{{NeurIPS}}, 2012.

\bibitem[Lample and Conneau(2019)]{cite:Arxiv19XLM}
Guillaume Lample and Alexis Conneau.
\newblock Cross-lingual language model pretraining.
\newblock In \emph{NeurIPS}, 2019.

\bibitem[Lample et~al.(2017)Lample, Denoyer, and
  Ranzato]{unsupervised_machine_translation}
Guillaume Lample, Ludovic Denoyer, and Marc'Aurelio Ranzato.
\newblock Unsupervised machine translation using monolingual corpora only.
\newblock In \emph{ICLR}, 2017.

\bibitem[Lan et~al.(2020)Lan, Chen, Goodman, Gimpel, Sharma, and
  Soricut]{cite:ICLR20Albert}
Zhenzhong Lan, Mingda Chen, Sebastian Goodman, Kevin Gimpel, Piyush Sharma, and
  Radu Soricut.
\newblock Albert: A lite bert for self-supervised learning of language
  representations.
\newblock In \emph{ICLR}, 2020.

\bibitem[LeCun et~al.(2015)LeCun, Bengio, and
  Hinton]{cite:Nature15DeepLearning}
Yann LeCun, Yoshua Bengio, and Geoffrey Hinton.
\newblock Deep learning.
\newblock \emph{Nature}, 521\penalty0 (7553):\penalty0 436--444, 2015.

\bibitem[Lee et~al.(2019)Lee, Batra, Baig, and Ulbricht]{SWD}
Chen{-}Yu Lee, Tanmay Batra, Mohammad~Haris Baig, and Daniel Ulbricht.
\newblock Sliced wasserstein discrepancy for unsupervised domain adaptation.
\newblock In \emph{CVPR}, 2019.

\bibitem[Lee et~al.(2020{\natexlab{a}})Lee, Cho, and Kang]{lee2020mixout}
Cheolhyoung Lee, Kyunghyun Cho, and Wanmo Kang.
\newblock Mixout: Effective regularization to finetune large-scale pretrained
  language models.
\newblock In \emph{ICLR}, 2020{\natexlab{a}}.

\bibitem[Lee et~al.(2020{\natexlab{b}})Lee, Yoon, Kim, Kim, Kim, So, and
  Kang]{lee2020biobert}
Jinhyuk Lee, Wonjin Yoon, Sungdong Kim, Donghyeon Kim, Sunkyu Kim, Chan~Ho So,
  and Jaewoo Kang.
\newblock Biobert: a pre-trained biomedical language representation model for
  biomedical text mining.
\newblock \emph{Bioinformatics}, 36\penalty0 (4):\penalty0 1234--1240,
  2020{\natexlab{b}}.

\bibitem[Lewis et~al.(2020)Lewis, Liu, Goyal, Ghazvininejad, Mohamed, Levy,
  Stoyanov, and Zettlemoyer]{lewis_bart:_2020}
Mike Lewis, Yinhan Liu, Naman Goyal, Marjan Ghazvininejad, Abdelrahman Mohamed,
  Omer Levy, Veselin Stoyanov, and Luke Zettlemoyer.
\newblock {BART}: {Denoising} {Sequence}-to-{Sequence} {Pre}-training for
  {Natural} {Language} {Generation}, {Translation}, and {Comprehension}.
\newblock In \emph{{ACL}}, 2020.

\bibitem[Li et~al.(2018)Li, Farkhoor, Liu, and Yosinski]{IntrinsicDimension}
Chunyuan Li, Heerad Farkhoor, Rosanne Liu, and Jason Yosinski.
\newblock Measuring the intrinsic dimension of objective landscapes.
\newblock In \emph{ICLR}, 2018.

\bibitem[Li and Liang(2021)]{li2021prefixtuning}
Xiang~Lisa Li and Percy Liang.
\newblock Prefix-tuning: Optimizing continuous prompts for generation.
\newblock In \emph{ACL}, 2021.

\bibitem[Li et~al.(2019)Li, Xiong, Wang, Rao, Liu, Chen, and
  Huan]{cite:ICLR19Delta}
Xingjian Li, Haoyi Xiong, Hanchao Wang, Yuxuan Rao, Liping Liu, Zeyu Chen, and
  Jun Huan.
\newblock Delta: Deep learning transfer using feature map with attention for
  convolutional networks.
\newblock In \emph{ICLR}, 2019.

\bibitem[Li et~al.(2017)Li, Wang, Shi, Liu, and Hou]{AdaBN}
Yanghao Li, Naiyan Wang, Jianping Shi, Jiaying Liu, and Xiaodi Hou.
\newblock Revisiting batch normalization for practical domain adaptation.
\newblock In \emph{ICLR Workshop}, 2017.

\bibitem[Li and Hoiem(2018)]{LWF}
Zhizhong Li and Derek Hoiem.
\newblock Learning without forgetting.
\newblock \emph{TPAMI}, 40\penalty0 (12):\penalty0 2935--2947, 2018.

\bibitem[Liang et~al.(2020)Liang, Hu, and Feng]{liang2020shot}
Jian Liang, Dapeng Hu, and Jiashi Feng.
\newblock Do we really need to access the source data? source hypothesis
  transfer for unsupervised domain adaptation.
\newblock In \emph{ICML}, 2020.

\bibitem[Liu et~al.(2021{\natexlab{a}})Liu, Wang, and Long]{liu2021cycle}
Hong Liu, Jianmin Wang, and Mingsheng Long.
\newblock Cycle self-training for domain adaptation.
\newblock In \emph{NeurIPS}, 2021{\natexlab{a}}.

\bibitem[Liu and Tuzel(2016)]{CoGAN}
Ming{-}Yu Liu and Oncel Tuzel.
\newblock Coupled generative adversarial networks.
\newblock In \emph{NeurIPS}, 2016.

\bibitem[Liu et~al.(2021{\natexlab{b}})Liu, Yuan, Fu, Jiang, Hayashi, and
  Neubig]{ppp_survey}
Pengfei Liu, Weizhe Yuan, Jinlan Fu, Zhengbao Jiang, Hiroaki Hayashi, and
  Graham Neubig.
\newblock Pre-train, prompt, and predict: A systematic survey of prompting
  methods in natural language processing, 2021{\natexlab{b}}.

\bibitem[Liu et~al.(2019{\natexlab{a}})Liu, He, Chen, and Gao]{liu2019mt-dnn}
Xiaodong Liu, Pengcheng He, Weizhu Chen, and Jianfeng Gao.
\newblock Multi-task deep neural networks for natural language understanding.
\newblock In \emph{ACL}, 2019{\natexlab{a}}.

\bibitem[Liu et~al.(2019{\natexlab{b}})Liu, Ott, Goyal, Du, Joshi, Chen, Levy,
  Lewis, Zettlemoyer, and Stoyanov]{cite:Arxiv19Roberta}
Yinhan Liu, Myle Ott, Naman Goyal, Jingfei Du, Mandar Joshi, Danqi Chen, Omer
  Levy, Mike Lewis, Luke Zettlemoyer, and Veselin Stoyanov.
\newblock Roberta: A robustly optimized bert pretraining approach.
\newblock \emph{arXiv preprint arXiv:1907.11692}, 2019{\natexlab{b}}.

\bibitem[Liu et~al.(2019{\natexlab{c}})Liu, Ott, Goyal, Du, Joshi, Chen, Levy,
  Lewis, Zettlemoyer, and Stoyanov]{roberta}
Yinhan Liu, Myle Ott, Naman Goyal, Jingfei Du, Mandar Joshi, Danqi Chen, Omer
  Levy, Mike Lewis, Luke Zettlemoyer, and Veselin Stoyanov.
\newblock Roberta: {A} robustly optimized {BERT} pretraining approach.
\newblock 2019{\natexlab{c}}.

\bibitem[Long et~al.(2013)Long, Wang, Ding, Sun, and Yu]{JDA}
Mingsheng Long, Jianmin Wang, Guiguang Ding, Jiaguang Sun, and Philip~S. Yu.
\newblock Transfer feature learning with joint distribution adaptation.
\newblock In \emph{ICCV}, 2013.

\bibitem[Long et~al.(2015)Long, Cao, Wang, and Jordan]{DAN}
Mingsheng Long, Yue Cao, Jianmin Wang, and Michael~I. Jordan.
\newblock Learning transferable features with deep adaptation networks.
\newblock In \emph{ICML}, 2015.

\bibitem[Long et~al.(2016)Long, Wang, and Jordan]{RTN}
Mingsheng Long, Jianmin Wang, and Michael~I. Jordan.
\newblock Unsupervised domain adaptation with residual transfer networks.
\newblock In \emph{NeurIPS}, 2016.

\bibitem[Long et~al.(2017)Long, Zhu, Wang, and Jordan]{JAN}
Mingsheng Long, Han Zhu, Jianmin Wang, and Michael~I Jordan.
\newblock Deep transfer learning with joint adaptation networks.
\newblock In \emph{ICML}, 2017.

\bibitem[Long et~al.(2018)Long, Cao, Wang, and Jordan]{CDAN}
Mingsheng Long, Zhangjie Cao, Jianmin Wang, and Michael~I. Jordan.
\newblock Conditional adversarial domain adaptation.
\newblock In \emph{NeurIPS}, 2018.

\bibitem[Long et~al.(2019)Long, Cao, Cao, Wang, and Jordan]{DAN19}
Mingsheng Long, Yue Cao, Zhangjie Cao, Jianmin Wang, and Michael~I. Jordan.
\newblock Transferable representation learning with deep adaptation networks.
\newblock \emph{TPAMI}, 41\penalty0 (12):\penalty0 3071--3085, 2019.

\bibitem[Louizos et~al.(2018)Louizos, Welling, and Kingma]{louizos2018learning}
Christos Louizos, Max Welling, and Diederik~P. Kingma.
\newblock Learning sparse neural networks through l\_0 regularization.
\newblock In \emph{ICLR}, 2018.

\bibitem[Mahajan et~al.(2018)Mahajan, Girshick, Ramanathan, He, Paluri, Li,
  Bharambe, and van~der Maaten]{cite:ECCV18ExploringWeakly}
Dhruv Mahajan, Ross Girshick, Vignesh Ramanathan, Kaiming He, Manohar Paluri,
  Yixuan Li, Ashwin Bharambe, and Laurens van~der Maaten.
\newblock Exploring the limits of weakly supervised pretraining.
\newblock In \emph{{ECCV}}, 2018.

\bibitem[Mallya and Lazebnik(2018)]{Piggyback}
Arun Mallya and Svetlana Lazebnik.
\newblock Piggyback: Adding multiple tasks to a single, fixed network by
  learning to mask.
\newblock In \emph{ECCV}, 2018.

\bibitem[Mansour et~al.(2009)Mansour, Mohri, and
  Rostamizadeh]{GeneralizedDATheory}
Yishay Mansour, Mehryar Mohri, and Afshin Rostamizadeh.
\newblock Domain adaptation: Learning bounds and algorithms.
\newblock In \emph{COLT}, 2009.

\bibitem[Munkhdalai and Yu(2017)]{cite:ICML17MetaNetworks}
Tsendsuren Munkhdalai and Hong Yu.
\newblock Meta networks.
\newblock In \emph{ICML}, 2017.

\bibitem[Ngiam et~al.(2018)Ngiam, Peng, Vasudevan, Kornblith, Le, and
  Pang]{cite:Arxiv18DomainAdaptiveTransfer}
Jiquan Ngiam, Daiyi Peng, Vijay Vasudevan, Simon Kornblith, Quoc~V Le, and
  Ruoming Pang.
\newblock Domain adaptive transfer learning with specialist models.
\newblock \emph{arXiv preprint arXiv:1811.07056}, 2018.

\bibitem[Nguyen et~al.(2020)Nguyen, Hassner, Seeger, and
  Archambeau]{cite:ICML20LEEP}
Cuong Nguyen, Tal Hassner, Matthias Seeger, and Cedric Archambeau.
\newblock Leep: A new measure to evaluate transferability of learned
  representations.
\newblock In \emph{ICML}, 2020.

\bibitem[Oord et~al.(2019)Oord, Li, and Vinyals]{cite:Arxiv18CPC}
Aaron van~den Oord, Yazhe Li, and Oriol Vinyals.
\newblock Representation learning with contrastive predictive coding.
\newblock \emph{NeurIPS}, 2019.

\bibitem[Pan and Yang(2010)]{transfer_survey}
Sinno~Jialin Pan and Qiang Yang.
\newblock A survey on transfer learning.
\newblock \emph{TKDE}, pages 1345--1359, 2010.

\bibitem[Pan et~al.(2011)Pan, Tsang, Kwok, and Yang]{TCA}
Sinno~Jialin Pan, Ivor~W. Tsang, James~T. Kwok, and Qiang Yang.
\newblock Domain adaptation via transfer component analysis.
\newblock \emph{TNNLS}, pages 199--210, 2011.

\bibitem[Paszke et~al.(2019)Paszke, Gross, Massa, Lerer, Bradbury, Chanan,
  Killeen, Lin, Gimelshein, Antiga, Desmaison, Kopf, Yang, DeVito, Raison,
  Tejani, Chilamkurthy, Steiner, Fang, Bai, and Chintala]{PyTorch}
Adam Paszke, Sam Gross, Francisco Massa, Adam Lerer, James Bradbury, Gregory
  Chanan, Trevor Killeen, Zeming Lin, Natalia Gimelshein, Luca Antiga, Alban
  Desmaison, Andreas Kopf, Edward Yang, Zachary DeVito, Martin Raison, Alykhan
  Tejani, Sasank Chilamkurthy, Benoit Steiner, Lu~Fang, Junjie Bai, and Soumith
  Chintala.
\newblock Pytorch: An imperative style, high-performance deep learning library.
\newblock In \emph{NeurIPS}, 2019.

\bibitem[Peng et~al.(2019)Peng, Bai, Xia, Huang, Saenko, and Wang]{domainnet}
Xingchao Peng, Qinxun Bai, Xide Xia, Zijun Huang, Kate Saenko, and Bo~Wang.
\newblock Moment matching for multi-source domain adaptation.
\newblock In \emph{ICCV}, 2019.

\bibitem[Peters et~al.(2016)Peters, B{\"u}hlmann, and
  Meinshausen]{cite:JRSS16ICP}
Jonas Peters, Peter B{\"u}hlmann, and Nicolai Meinshausen.
\newblock Causal inference by using invariant prediction: identification and
  confidence intervals.
\newblock \emph{Journal of the Royal Statistical Society. Series B (Statistical
  Methodology)}, pages 947--1012, 2016.

\bibitem[Peters et~al.(2017)Peters, Janzing, and
  Sch{\"o}lkopf]{cite:Book17ElementsofCausalInference}
Jonas Peters, Dominik Janzing, and Bernhard Sch{\"o}lkopf.
\newblock \emph{Elements of causal inference: foundations and learning
  algorithms}.
\newblock The MIT Press, 2017.

\bibitem[Peters et~al.(2018)Peters, Neumann, Iyyer, Gardner, Clark, Lee, and
  Zettlemoyer]{cite:NAACL18ELMo}
Matthew~E Peters, Mark Neumann, Mohit Iyyer, Matt Gardner, Christopher Clark,
  Kenton Lee, and Luke Zettlemoyer.
\newblock Deep contextualized word representations.
\newblock In \emph{NAACL}, 2018.

\bibitem[Pires et~al.(2019)Pires, Schlinger, and Garrette]{multilingual_bert}
Telmo Pires, Eva Schlinger, and Dan Garrette.
\newblock How multilingual is multilingual bert?
\newblock In \emph{ACL}, 2019.

\bibitem[Quionero-Candela et~al.(2009)Quionero-Candela, Sugiyama, Schwaighofer,
  and Lawrence]{cite:Book09DSS}
J.~Quionero-Candela, M.~Sugiyama, A.~Schwaighofer, and N.~D. Lawrence.
\newblock \emph{Dataset shift in machine learning}.
\newblock The MIT Press, 2009.

\bibitem[Radford et~al.(2018)Radford, Narasimhan, Salimans, and
  Sutskever]{cite:GPT}
Alec Radford, Karthik Narasimhan, Tim Salimans, and Ilya Sutskever.
\newblock Improving language understanding by generative pre-training.
\newblock \emph{Technical report, OpenAI}, 2018.

\bibitem[Radford et~al.(2021)Radford, Kim, Hallacy, Ramesh, Goh, Agarwal,
  Sastry, Askell, Mishkin, Clark, et~al.]{cite:Arxiv21CLIP}
Alec Radford, Jong~Wook Kim, Chris Hallacy, Aditya Ramesh, Gabriel Goh,
  Sandhini Agarwal, Girish Sastry, Amanda Askell, Pamela Mishkin, Jack Clark,
  et~al.
\newblock Learning transferable visual models from natural language
  supervision.
\newblock In \emph{ICML}, 2021.

\bibitem[Raffel et~al.(2020)Raffel, Shazeer, Roberts, Lee, Narang, Matena,
  Zhou, Li, and Liu]{cite:JMLR20T5}
Colin Raffel, Noam Shazeer, Adam Roberts, Katherine Lee, Sharan Narang, Michael
  Matena, Yanqi Zhou, Wei Li, and Peter~J Liu.
\newblock Exploring the limits of transfer learning with a unified text-to-text
  transformer.
\newblock \emph{JMLR}, 21\penalty0 (140):\penalty0 1--67, 2020.

\bibitem[Raghu et~al.(2020)Raghu, Raghu, Bengio, and Vinyals]{cite:ICLR20ANIL}
Aniruddh Raghu, Maithra Raghu, Samy Bengio, and Oriol Vinyals.
\newblock Rapid learning or feature reuse? towards understanding the
  effectiveness of maml.
\newblock In \emph{ICLR}, 2020.

\bibitem[Raghu et~al.(2019)Raghu, Zhang, Kleinberg, and
  Bengio]{cite:NIPS19Transfusion}
Maithra Raghu, Chiyuan Zhang, Jon Kleinberg, and Samy Bengio.
\newblock Transfusion: Understanding transfer learning for medical imaging.
\newblock In \emph{NeurIPS}, 2019.

\bibitem[Rebuffi et~al.(2017)Rebuffi, Bilen, and Vedaldi]{Rebuffi17}
S-A Rebuffi, H.~Bilen, and A.~Vedaldi.
\newblock Learning multiple visual domains with residual adapters.
\newblock In \emph{NeurIPS}, 2017.

\bibitem[Redko et~al.(2020)Redko, Morvant, Habrard, Sebban, and
  Bennani]{redko2020survey}
Ievgen Redko, Emilie Morvant, Amaury Habrard, Marc Sebban, and Younès Bennani.
\newblock A survey on domain adaptation theory: learning bounds and theoretical
  guarantees, 2020.

\bibitem[Ren et~al.(2015)Ren, He, Girshick, and Sun]{Faster}
Shaoqing Ren, Kaiming He, Ross Girshick, and Jian Sun.
\newblock Faster r-cnn: Towards real-time object detection with region proposal
  networks.
\newblock In \emph{NeurIPS}, 2015.

\bibitem[Rosenstein(2005)]{Rosenstein2005ToTO}
Michael~T. Rosenstein.
\newblock To transfer or not to transfer.
\newblock In \emph{NeurIPS}, 2005.

\bibitem[Rusak et~al.(2021)Rusak, Schneider, Gehler, Bringmann, Brendel, and
  Bethge]{rusak2021selflearning}
Evgenia. Rusak, Steffen Schneider, Peter Gehler, Oliver Bringmann, Wieland
  Brendel, and Matthias Bethge.
\newblock Adapting imagenet-scale models to complex distribution shifts with
  self-learning.
\newblock \emph{arXiv preprint arXiv:2104.12928}, 2021.

\bibitem[Russakovsky et~al.(2015)Russakovsky, Deng, Su, Krause, Satheesh, Ma,
  Huang, Karpathy, Khosla, Bernstein, Berg, and Fei-Fei]{ImageNet}
Olga Russakovsky, Jia Deng, Hao Su, Jonathan Krause, Sanjeev Satheesh, Sean Ma,
  Zhiheng Huang, Andrej Karpathy, Aditya Khosla, Michael Bernstein,
  Alexander~C. Berg, and Li~Fei-Fei.
\newblock {ImageNet Large Scale Visual Recognition Challenge}.
\newblock \emph{IJCV}, 115\penalty0 (3):\penalty0 211--252, 2015.

\bibitem[Rusu et~al.(2019)Rusu, Rao, Sygnowski, Vinyals, Pascanu, Osindero, and
  Hadsell]{cite:ICLR19LEO}
Andrei~A Rusu, Dushyant Rao, Jakub Sygnowski, Oriol Vinyals, Razvan Pascanu,
  Simon Osindero, and Raia Hadsell.
\newblock Meta-learning with latent embedding optimization.
\newblock In \emph{ICLR}, 2019.

\bibitem[Saito et~al.(2018)Saito, Watanabe, Ushiku, and Harada]{MCD}
Kuniaki Saito, Kohei Watanabe, Yoshitaka Ushiku, and Tatsuya Harada.
\newblock Maximum classifier discrepancy for unsupervised domain adaptation.
\newblock In \emph{CVPR}, 2018.

\bibitem[Saito et~al.(2019)Saito, Ushiku, Harada, and Saenko]{SWDA}
Kuniaki Saito, Yoshitaka Ushiku, Tatsuya Harada, and Kate Saenko.
\newblock Strong-weak distribution alignment for adaptive object detection.
\newblock In \emph{CVPR}, 2019.

\bibitem[Salman et~al.(2020)Salman, Ilyas, Engstrom, Kapoor, and
  Madry]{cite:NIPS20AdversariallyRoubstTransfer}
Hadi Salman, Andrew Ilyas, Logan Engstrom, Ashish Kapoor, and Aleksander Madry.
\newblock Do adversarially robust imagenet models transfer better?
\newblock In \emph{NeurIPS}, 2020.

\bibitem[Sankaranarayanan et~al.(2018)Sankaranarayanan, Balaji, Castillo, and
  Chellappa]{Gen2Adapt}
Swami Sankaranarayanan, Yogesh Balaji, Carlos~D. Castillo, and Rama Chellappa.
\newblock Generate to adapt: Aligning domains using generative adversarial
  networks.
\newblock In \emph{CVPR}, 2018.

\bibitem[Santoro et~al.(2016)Santoro, Bartunov, Botvinick, Wierstra, and
  Lillicrap]{cite:ICML16MetaLearningMemory}
Adam Santoro, Sergey Bartunov, Matthew Botvinick, Daan Wierstra, and Timothy
  Lillicrap.
\newblock Meta-learning with memory-augmented neural networks.
\newblock In \emph{ICML}, 2016.

\bibitem[Schick and Schütze(2020)]{schick2020exploiting}
Timo Schick and Hinrich Schütze.
\newblock Exploiting cloze questions for few-shot text classification and
  natural language inference.
\newblock In \emph{EACL}, 2020.

\bibitem[Schmidhuber(1987)]{cite:1987EvolutionaryPrinciples}
J{\"u}rgen Schmidhuber.
\newblock \emph{Evolutionary principles in self-referential learning}.
\newblock PhD thesis, Technische Universit{\"a}t M{\"u}nchen, 1987.

\bibitem[Sch{\"o}lkopf et~al.(2012)Sch{\"o}lkopf, Janzing, Peters, Sgouritsa,
  Zhang, and Mooij]{cite:ICML12OnCausalandAntiCausal}
Bernhard Sch{\"o}lkopf, Dominik Janzing, Jonas Peters, Eleni Sgouritsa, Kun
  Zhang, and Joris Mooij.
\newblock On causal and anticausal learning.
\newblock In \emph{{ICML}}, 2012.

\bibitem[Sch{\"o}lkopf et~al.(2021)Sch{\"o}lkopf, Locatello, Bauer, Ke,
  Kalchbrenner, Goyal, and Bengio]{cite:ProceedingIEEE21Causal}
Bernhard Sch{\"o}lkopf, Francesco Locatello, Stefan Bauer, Nan~Rosemary Ke, Nal
  Kalchbrenner, Anirudh Goyal, and Yoshua Bengio.
\newblock Toward causal representation learning.
\newblock \emph{Proceedings of the IEEE}, 109\penalty0 (5):\penalty0 612--634,
  2021.

\bibitem[Senior et~al.(2020)Senior, Evans, Jumper, Kirkpatrick, Sifre, Green,
  Qin, {\v{Z}}{\'\i}dek, Nelson, Bridgland, et~al.]{cite:Nature20AlphaFold}
Andrew~W Senior, Richard Evans, John Jumper, James Kirkpatrick, Laurent Sifre,
  Tim Green, Chongli Qin, Augustin {\v{Z}}{\'\i}dek, Alexander~WR Nelson, Alex
  Bridgland, et~al.
\newblock Improved protein structure prediction using potentials from deep
  learning.
\newblock \emph{Nature}, 577\penalty0 (7792):\penalty0 706--710, 2020.

\bibitem[Sermanet et~al.(2013)Sermanet, Eigen, Zhang, Mathieu, Fergus, and
  LeCun]{cite:Arxiv13Overfeat}
Pierre Sermanet, David Eigen, Xiang Zhang, Micha{\"e}l Mathieu, Rob Fergus, and
  Yann LeCun.
\newblock Overfeat: Integrated recognition, localization and detection using
  convolutional networks.
\newblock \emph{arXiv preprint arXiv:1312.6229}, 2013.

\bibitem[Shrivastava et~al.(2017)Shrivastava, Pfister, Tuzel, Susskind, Wang,
  and Webb]{SimGAN}
Ashish Shrivastava, Tomas Pfister, Oncel Tuzel, Josh Susskind, Wenda Wang, and
  Russell Webb.
\newblock Learning from simulated and unsupervised images through adversarial
  training.
\newblock In \emph{CVPR}, 2017.

\bibitem[Shu et~al.(2018)Shu, Bui, Narui, and Ermon]{DIRT}
Rui Shu, Hung~H. Bui, Hirokazu Narui, and Stefano Ermon.
\newblock A dirt-t approach to unsupervised domain adaptation.
\newblock In \emph{ICLR}, 2018.

\bibitem[Shu et~al.(2021{\natexlab{a}})Shu, Cao, Gao, Wang, and
  Long]{cite:Arxiv21OmniTraining}
Yang Shu, Zhangjie Cao, Jinghan Gao, Jianmin Wang, and Mingsheng Long.
\newblock Omni-training for data-efficient deep learning.
\newblock \emph{arXiv preprint arXiv:2110.07510}, 2021{\natexlab{a}}.

\bibitem[Shu et~al.(2021{\natexlab{b}})Shu, Kou, Cao, Wang, and
  Long]{cite:ICML21ZooTuning}
Yang Shu, Zhi Kou, Zhangjie Cao, Jianmin Wang, and Mingsheng Long.
\newblock Zoo-tuning: Adaptive transfer from a zoo of models.
\newblock In \emph{ICML}, 2021{\natexlab{b}}.

\bibitem[Silver et~al.(2016)Silver, Huang, Maddison, Guez, Sifre, Van
  Den~Driessche, Schrittwieser, Antonoglou, Panneershelvam, Lanctot,
  et~al.]{cite:Nature16AlphaGO}
David Silver, Aja Huang, Chris~J Maddison, Arthur Guez, Laurent Sifre, George
  Van Den~Driessche, Julian Schrittwieser, Ioannis Antonoglou, Veda
  Panneershelvam, Marc Lanctot, et~al.
\newblock Mastering the game of go with deep neural networks and tree search.
\newblock \emph{Nature}, 529\penalty0 (7587):\penalty0 484--489, 2016.

\bibitem[Snell et~al.(2017)Snell, Swersky, and Zemel]{PrototypicalNetworks}
Jake Snell, Kevin Swersky, and Richard~S. Zemel.
\newblock Prototypical networks for few-shot learning.
\newblock In \emph{NeurIPS}, 2017.

\bibitem[Sriperumbudur et~al.(2010)Sriperumbudur, Gretton, Fukumizu,
  Schölkopf, and Lanckriet]{HilbertSpaceOnProbability}
Bharath~K. Sriperumbudur, Arthur Gretton, Kenji Fukumizu, Bernhard Schölkopf,
  and Gert R.~G. Lanckriet.
\newblock Hilbert space embeddings and metrics on probability measures.
\newblock \emph{JMLR}, 2010.

\bibitem[Sugiyama et~al.(2007)Sugiyama, Krauledat, and M{{\"u}}ller]{IWCV}
Masashi Sugiyama, Matthias Krauledat, and Klaus-Robert M{{\"u}}ller.
\newblock Covariate shift adaptation by importance weighted cross validation.
\newblock \emph{JMLR}, 8\penalty0 (35):\penalty0 985--1005, 2007.

\bibitem[Sugiyama et~al.(2008)Sugiyama, Nakajima, Kashima, Buenau, and
  Kawanabe]{NIPS2007_be83ab3e}
Masashi Sugiyama, Shinichi Nakajima, Hisashi Kashima, Paul Buenau, and Motoaki
  Kawanabe.
\newblock Direct importance estimation with model selection and its application
  to covariate shift adaptation.
\newblock In \emph{NeurIPS}, 2008.

\bibitem[Sun and Saenko(2016)]{coral}
Baochen Sun and Kate Saenko.
\newblock Deep coral: Correlation alignment for deep domain adaptation.
\newblock In \emph{ECCV}, 2016.

\bibitem[Sun et~al.(2019{\natexlab{a}})Sun, Liu, Chua, and
  Schiele]{cite:CVPR19MeteTransferLearning}
Qianru Sun, Yaoyao Liu, Tat-Seng Chua, and Bernt Schiele.
\newblock Meta-transfer learning for few-shot learning.
\newblock In \emph{CVPR}, 2019{\natexlab{a}}.

\bibitem[Sun et~al.(2019{\natexlab{b}})Sun, Wang, Li, Feng, Chen, Zhang, Tian,
  Zhu, Tian, and Wu]{sun2019ernie}
Yu~Sun, Shuohuan Wang, Yukun Li, Shikun Feng, Xuyi Chen, Han Zhang, Xin Tian,
  Danxiang Zhu, Hao Tian, and Hua Wu.
\newblock Ernie: Enhanced representation through knowledge integration.
\newblock \emph{arXiv preprint arXiv:1904.09223}, 2019{\natexlab{b}}.

\bibitem[Taigman et~al.(2017)Taigman, Polyak, and Wolf]{DTN}
Yaniv Taigman, Adam Polyak, and Lior Wolf.
\newblock Unsupervised cross-domain image generation.
\newblock In \emph{ICLR}, 2017.

\bibitem[Thrun and Pratt(1998)]{cite:LearningtoLearnBook}
Sebastian Thrun and Lorien Pratt.
\newblock \emph{Learning to learn}.
\newblock Springer Science \& Business Media, 1998.

\bibitem[Tian et~al.(2020)Tian, Krishnan, and Isola]{cite:ECCV20CMC}
Yonglong Tian, Dilip Krishnan, and Phillip Isola.
\newblock Contrastive multiview coding.
\newblock In \emph{ECCV}, 2020.

\bibitem[Torrey and Shavlik(2010)]{transfer_learning}
Lisa Torrey and Jude Shavlik.
\newblock Transfer learning.
\newblock 2010.

\bibitem[Tsai et~al.(2018)Tsai, Hung, Schulter, Sohn, Yang, and
  Chandraker]{AdaptSegNet}
Yi{-}Hsuan Tsai, Wei{-}Chih Hung, Samuel Schulter, Kihyuk Sohn, Ming{-}Hsuan
  Yang, and Manmohan Chandraker.
\newblock Learning to adapt structured output space for semantic segmentation.
\newblock In \emph{CVPR}, 2018.

\bibitem[Tzeng et~al.(2014)Tzeng, Hoffman, Zhang, Saenko, and Darrell]{DDC}
Eric Tzeng, Judy Hoffman, Ning Zhang, Kate Saenko, and Trevor Darrell.
\newblock Deep domain confusion: Maximizing for domain invariance.
\newblock 2014.

\bibitem[Tzeng et~al.(2015)Tzeng, Hoffman, Darrell, and
  Saenko]{SimultaneousDADomainAndTasks}
Eric Tzeng, Judy Hoffman, Trevor Darrell, and Kate Saenko.
\newblock Simultaneous deep transfer across domains and tasks.
\newblock In \emph{ICCV}, pages 4068--4076, 2015.

\bibitem[Tzeng et~al.(2017)Tzeng, Hoffman, Saenko, and Darrell]{ADDA}
Eric Tzeng, Judy Hoffman, Kate Saenko, and Trevor Darrell.
\newblock Adversarial discriminative domain adaptation.
\newblock In \emph{CVPR}, 2017.

\bibitem[Vaswani et~al.(2017)Vaswani, Shazeer, Parmar, Uszkoreit, Jones, Gomez,
  Kaiser, and Polosukhin]{cite:NIPS17Transformer}
Ashish Vaswani, Noam Shazeer, Niki Parmar, Jakob Uszkoreit, Llion Jones,
  Aidan~N Gomez, {\L}ukasz Kaiser, and Illia Polosukhin.
\newblock Attention is all you need.
\newblock In \emph{NeurIPS}, 2017.

\bibitem[Veli{\v{c}}kovi{\'c} et~al.(2019)Veli{\v{c}}kovi{\'c}, Fedus,
  Hamilton, Li{\`o}, Bengio, and Hjelm]{cite:ICLR19GraphInfoMax}
Petar Veli{\v{c}}kovi{\'c}, William Fedus, William~L Hamilton, Pietro Li{\`o},
  Yoshua Bengio, and R~Devon Hjelm.
\newblock Deep graph infomax.
\newblock In \emph{ICLR}, 2019.

\bibitem[Vincent et~al.(2008)Vincent, Larochelle, Bengio, and
  Manzagol]{cite:ICML08DenoisingAE}
Pascal Vincent, Hugo Larochelle, Yoshua Bengio, and Pierre-Antoine Manzagol.
\newblock Extracting and composing robust features with denoising autoencoders.
\newblock In \emph{ICML}, 2008.

\bibitem[Vinyals et~al.(2016)Vinyals, Blundell, Lillicrap, Wierstra,
  et~al.]{cite:NIPS16MatchingNet}
Oriol Vinyals, Charles Blundell, Timothy Lillicrap, Daan Wierstra, et~al.
\newblock Matching networks for one shot learning.
\newblock In \emph{NeurIPS}, 2016.

\bibitem[Vinyals et~al.(2019)Vinyals, Babuschkin, Czarnecki, Mathieu, Dudzik,
  Chung, Choi, Powell, Ewalds, Georgiev, et~al.]{cite:Nature19AlphaStar}
Oriol Vinyals, Igor Babuschkin, Wojciech~M Czarnecki, Micha{\"e}l Mathieu,
  Andrew Dudzik, Junyoung Chung, David~H Choi, Richard Powell, Timo Ewalds,
  Petko Georgiev, et~al.
\newblock Grandmaster level in starcraft ii using multi-agent reinforcement
  learning.
\newblock \emph{Nature}, 575\penalty0 (7782):\penalty0 350--354, 2019.

\bibitem[Vu et~al.(2019)Vu, Jain, Bucher, Cord, and Perez]{Advent}
Tuan-Hung Vu, Himalaya Jain, Maxime Bucher, Matthieu Cord, and Patrick Perez.
\newblock Advent: Adversarial entropy minimization for domain adaptation in
  semantic segmentation.
\newblock In \emph{CVPR}, 2019.

\bibitem[Wang et~al.(2019{\natexlab{a}})Wang, Singh, Michael, Hill, Levy, and
  Bowman]{GLUE}
Alex Wang, Amanpreet Singh, Julian Michael, Felix Hill, Omer Levy, and
  Samuel~R. Bowman.
\newblock {GLUE}: A multi-task benchmark and analysis platform for natural
  language understanding.
\newblock In \emph{ICLR}, 2019{\natexlab{a}}.

\bibitem[Wang et~al.(2019{\natexlab{b}})Wang, Ge, Lipton, and
  Xing]{ImageNetSketch}
Haohan Wang, Songwei Ge, Zachary Lipton, and Eric~P Xing.
\newblock Learning robust global representations by penalizing local predictive
  power.
\newblock In \emph{NeurIPS}, 2019{\natexlab{b}}.

\bibitem[Wang et~al.(2019{\natexlab{c}})Wang, Jin, Long, Wang, and
  Jordan]{TransNorm}
Ximei Wang, Ying Jin, Mingsheng Long, Jianmin Wang, and Michael~I Jordan.
\newblock Transferable normalization: Towards improving transferability of deep
  neural networks.
\newblock In \emph{NeurIPS}, 2019{\natexlab{c}}.

\bibitem[Wang et~al.(2021)Wang, Gao, Long, and Wang]{cite:ICML21SelfTuning}
Ximei Wang, Jinghan Gao, Mingsheng Long, and Jianmin Wang.
\newblock Self-tuning for data-efficient deep learning.
\newblock In \emph{ICML}, 2021.

\bibitem[Wang et~al.(2019{\natexlab{d}})Wang, Dai, P{\'{o}}czos, and
  Carbonell]{CharacterizingNegativeTransfer}
Zirui Wang, Zihang Dai, Barnab{\'{a}}s P{\'{o}}czos, and Jaime~G. Carbonell.
\newblock Characterizing and avoiding negative transfer.
\newblock In \emph{CVPR}, 2019{\natexlab{d}}.

\bibitem[Wei et~al.(2022)Wei, Bosma, Zhao, Guu, Yu, Lester, Du, Dai, and
  Le]{wei2021finetuned}
Jason Wei, Maarten Bosma, Vincent~Y. Zhao, Kelvin Guu, Adams~Wei Yu, Brian
  Lester, Nan Du, Andrew~M. Dai, and Quoc~V. Le.
\newblock Finetuned language models are zero-shot learners.
\newblock In \emph{ICLR}, 2022.

\bibitem[Wei et~al.(2018)Wei, Zhang, Gao, and Tian]{PersonGAN}
Longhui Wei, Shiliang Zhang, Wen Gao, and Qi~Tian.
\newblock Person transfer gan to bridge domain gap for person
  re-identification.
\newblock In \emph{CVPR}, 2018.

\bibitem[Wightman(2019)]{rw2019timm}
Ross Wightman.
\newblock Pytorch image models.
\newblock \url{https://github.com/rwightman/pytorch-image-models}, 2019.

\bibitem[Wu and He(2018)]{GroupNorm}
Yuxin Wu and Kaiming He.
\newblock Group normalization.
\newblock In \emph{ECCV}, 2018.

\bibitem[Wu et~al.(2018)Wu, Xiong, Stella, and Lin]{wu2018unsupervised}
Zhirong Wu, Yuanjun Xiong, X~Yu Stella, and Dahua Lin.
\newblock Unsupervised feature learning via non-parametric instance
  discrimination.
\newblock In \emph{CVPR}, 2018.

\bibitem[Xu et~al.(2021)Xu, Luo, Zhang, Tan, Chang, Huang, and
  Huang]{xu2021raise}
Runxin Xu, Fuli Luo, Zhiyuan Zhang, Chuanqi Tan, Baobao Chang, Songfang Huang,
  and Fei Huang.
\newblock Raise a child in large language model: Towards effective and
  generalizable fine-tuning.
\newblock In \emph{EMNLP}, 2021.

\bibitem[Yalniz et~al.(2019)Yalniz, J{\'e}gou, Chen, Paluri, and
  Mahajan]{cite:Arxiv19BillionSemiSupervised}
I~Zeki Yalniz, Herv{\'e} J{\'e}gou, Kan Chen, Manohar Paluri, and Dhruv
  Mahajan.
\newblock Billion-scale semi-supervised learning for image classification.
\newblock \emph{arXiv preprint arXiv:1905.00546}, 2019.

\bibitem[Yang et~al.(2019)Yang, Dai, Yang, Carbonell, Salakhutdinov, and
  Le]{cite:NIPS19XLNet}
Zhilin Yang, Zihang Dai, Yiming Yang, Jaime Carbonell, Russ~R Salakhutdinov,
  and Quoc~V Le.
\newblock Xlnet: Generalized autoregressive pretraining for language
  understanding.
\newblock In \emph{NeurIPS}, 2019.

\bibitem[Yao et~al.(2019)Yao, Wei, Huang, and
  Li]{cite:ICML19HierarchicallyMetaLearning}
Huaxiu Yao, Ying Wei, Junzhou Huang, and Zhenhui Li.
\newblock Hierarchically structured meta-learning.
\newblock In \emph{ICML}, 2019.

\bibitem[Yosinski et~al.(2014)Yosinski, Clune, Bengio, and
  Lipson]{cite:NIPS14HowTransferable}
Jason Yosinski, Jeff Clune, Yoshua Bengio, and Hod Lipson.
\newblock How transferable are features in deep neural networks?
\newblock In \emph{NeurIPS}, 2014.

\bibitem[You et~al.(2021)You, Liu, Wang, and Long]{cite:ICML21LogME}
Kaichao You, Yong Liu, Jianmin Wang, and Mingsheng Long.
\newblock Logme: Practical assessment of pre-trained models for transfer
  learning.
\newblock In \emph{ICML}, 2021.

\bibitem[Zamir et~al.(2018)Zamir, Sax, Shen, Guibas, Malik, and
  Savarese]{taskonomy}
Amir~Roshan Zamir, Alexander Sax, William~B. Shen, Leonidas~J. Guibas, Jitendra
  Malik, and Silvio Savarese.
\newblock Taskonomy: Disentangling task transfer learning.
\newblock In \emph{CVPR}, 2018.

\bibitem[Zellinger et~al.(2017)Zellinger, Grubinger, Lughofer, Natschläger,
  and Saminger-Platz]{CMD}
Werner Zellinger, Thomas Grubinger, Edwin Lughofer, Thomas Natschläger, and
  Susanne Saminger-Platz.
\newblock Central moment discrepancy (cmd) for domain-invariant representation
  learning.
\newblock In \emph{ICLR}, 2017.

\bibitem[Zhang et~al.(2017)Zhang, Bengio, Hardt, Recht, and
  Vinyals]{RethinkGeneralization}
Chiyuan Zhang, Samy Bengio, Moritz Hardt, Benjamin Recht, and Oriol Vinyals.
\newblock Understanding deep learning requires rethinking generalization.
\newblock In \emph{ICLR}, 2017.

\bibitem[Zhang et~al.(2019{\natexlab{a}})Zhang, Yu, Jiao, Xing, Ghaoui, and
  Jordan]{robust_accuracy_trade_off}
Hongyang Zhang, Yaodong Yu, Jiantao Jiao, Eric~P. Xing, Laurent~El Ghaoui, and
  Michael~I. Jordan.
\newblock Theoretically principled trade-off between robustness and accuracy.
\newblock In \emph{ICML}, 2019{\natexlab{a}}.

\bibitem[Zhang et~al.(2019{\natexlab{b}})Zhang, Sax, Zamir, Guibas, and
  Malik]{sidetuning2019}
Jeffrey~O. Zhang, Alexander Sax, Amir Zamir, Leonidas~J. Guibas, and Jitendra
  Malik.
\newblock Side-tuning: Network adaptation via additive side networks.
\newblock 2019{\natexlab{b}}.

\bibitem[Zhang et~al.(2019{\natexlab{c}})Zhang, Liu, Long, and Jordan]{MDD}
Yuchen Zhang, Tianle Liu, Mingsheng Long, and Michael Jordan.
\newblock Bridging theory and algorithm for domain adaptation.
\newblock In \emph{ICML}, 2019{\natexlab{c}}.

\bibitem[Zhang and Sabuncu(2018)]{GCE}
Zhilu Zhang and Mert~R. Sabuncu.
\newblock Generalized cross entropy loss for training deep neural networks with
  noisy labels.
\newblock In \emph{NeurIPS}, 2018.

\bibitem[Zhao et~al.(2021)Zhao, Wu, Lau, and Lin]{cite:ICLR21InstanceTransfer}
Nanxuan Zhao, Zhirong Wu, Rynson W.~H. Lau, and Stephen Lin.
\newblock What makes instance discrimination good for transfer learning?
\newblock In \emph{ICLR}, 2021.

\bibitem[Zheng et~al.(2021)Zheng, Guha, Anderson, Henderson, and
  Ho]{zhengguha2021}
Lucia Zheng, Neel Guha, Brandon~R. Anderson, Peter Henderson, and Daniel~E. Ho.
\newblock When does pretraining help? assessing self-supervised learning for
  law and the casehold dataset.
\newblock In \emph{ICAIL}, 2021.

\bibitem[Zhong et~al.(2020)Zhong, Wang, Kou, Wang, and
  Long]{cite:Arxiv20BiTuning}
Jincheng Zhong, Ximei Wang, Zhi Kou, Jianmin Wang, and Mingsheng Long.
\newblock Bi-tuning of pre-trained representations.
\newblock \emph{arXiv preprint arXiv:2011.06182}, 2020.

\bibitem[Zhu et~al.(2017)Zhu, Park, Isola, and Efros]{CycleGAN}
Jun-Yan Zhu, Taesung Park, Phillip Isola, and Alexei~A Efros.
\newblock Unpaired image-to-image translation using cycle-consistent
  adversarial networks.
\newblock In \emph{ICCV}, 2017.

\bibitem[Zhu et~al.(2015)Zhu, Kiros, Zemel, Salakhutdinov, Urtasun, Torralba,
  and Fidler]{cite:ICCV15BookCorpus}
Yukun Zhu, Ryan Kiros, Rich Zemel, Ruslan Salakhutdinov, Raquel Urtasun,
  Antonio Torralba, and Sanja Fidler.
\newblock Aligning books and movies: Towards story-like visual explanations by
  watching movies and reading books.
\newblock In \emph{ICCV}, 2015.

\bibitem[Zhuang et~al.(2021)Zhuang, Qi, Duan, Xi, Zhu, Zhu, Xiong, and
  He]{Zhuang21}
Fuzhen Zhuang, Zhiyuan Qi, Keyu Duan, Dongbo Xi, Yongchun Zhu, Hengshu Zhu, Hui
  Xiong, and Qing He.
\newblock A comprehensive survey on transfer learning.
\newblock \emph{Proceedings of the IEEE}, 109\penalty0 (1):\penalty0 43--76,
  2021.

\bibitem[Zou et~al.(2018)Zou, Yu, Kumar, and Wang]{CBST}
Yang Zou, Zhiding Yu, B.~V. K.~Vijaya Kumar, and Jinsong Wang.
\newblock Unsupervised domain adaptation for semantic segmentation via
  class-balanced self-training.
\newblock In \emph{ECCV}, 2018.

\end{thebibliography}

\end{document}